\documentclass{article}

\PassOptionsToPackage{numbers, compress}{natbib}
\usepackage[preprint]{neurips_2025}

\usepackage[utf8]{inputenc} 
\usepackage[T1]{fontenc}    
\usepackage{hyperref}       
\usepackage{url}            
\usepackage{booktabs}       
\usepackage{amsfonts}       
\usepackage{nicefrac}       
\usepackage{microtype}      
\usepackage{xcolor}         

\usepackage[accsupp]{axessibility} 
\usepackage{multirow}
\usepackage{graphicx}
\usepackage{bm}
\usepackage{bbm}
\usepackage{makecell}
\usepackage{pbox}
\usepackage{float}
\usepackage[export]{adjustbox}

\usepackage{colortbl}
\usepackage{subcaption}
\usepackage{wrapfig}
\usepackage{array}
\usepackage{caption}
\usepackage{enumitem}

\definecolor{myorange}{RGB}{245,156,74}
\definecolor{myred}{RGB}{197,41,114}
\hypersetup{
  colorlinks=true,
  urlcolor=red,
  linkcolor=red,
  citecolor=-myorange,
  anchorcolor=blue
}

\RequirePackage{xspace}
\makeatletter
\DeclareRobustCommand\onedot{\futurelet\@let@token\@onedot}
\def\@onedot{\ifx\@let@token.\else.\null\fi\xspace}

\def\eg{\emph{e.g}\onedot} 

\def\ie{\emph{i.e}\onedot}

\makeatother

\title{UniRestorer: Universal Image Restoration via Adaptively Estimating Image Degradation at Proper Granularity}

\author{%
  Jingbo Lin$^{1}$,\quad Zhilu Zhang$^{1}$,\quad Wenbo Li$^{2}$,\quad Renjing Pei$^{2}$,\quad Hang Xu$^{2}$, \\ \vspace{2.2mm} \textbf{Hongzhi Zhang}$^{1}$,\quad \textbf{Wangmeng Zuo}$^{1}$ 
  \\
    \footnotesize $^{1}$Harbin Institute of Technology\quad 
    \footnotesize $^{2}$Huawei Noah’s Ark Lab\\
}

\begin{document}

\maketitle

\begin{abstract}
Recently, considerable progress has been made in all-in-one image restoration.
Generally, existing methods can be degradation-agnostic or degradation-aware.
However, the former are limited in leveraging degradation-specific restoration, and the latter suffer from the inevitable error in degradation estimation. 
Consequently, the performance of existing methods has a large gap compared to specific single-task models.
In this work, we make a step forward in this topic, and present our UniRestorer with improved restoration performance.
Specifically, we perform hierarchical clustering on degradation space, and train a multi-granularity mixture-of-experts (MoE) restoration model. 
Then, UniRestorer adopts both degradation and granularity estimation to adaptively select an appropriate expert for image restoration.
In contrast to existing degradation-agnostic and -aware methods, UniRestorer can leverage degradation estimation to benefit degradation-specific restoration, and use granularity estimation to make the model robust to degradation estimation error. 
Experimental results show that our UniRestorer outperforms state-of-the-art all-in-one methods by a large margin, and is promising in closing the performance gap to specific single-task models.
The code and pre-trained models will be publicly released.
\end{abstract}
\section{Introduction}
\label{sec:introduction}

\begin{figure*}[t]
	\centering
	\setlength{\abovecaptionskip}{0.1cm}
	\setlength{\belowcaptionskip}{-0.3cm}
	\includegraphics[width=0.85\columnwidth]{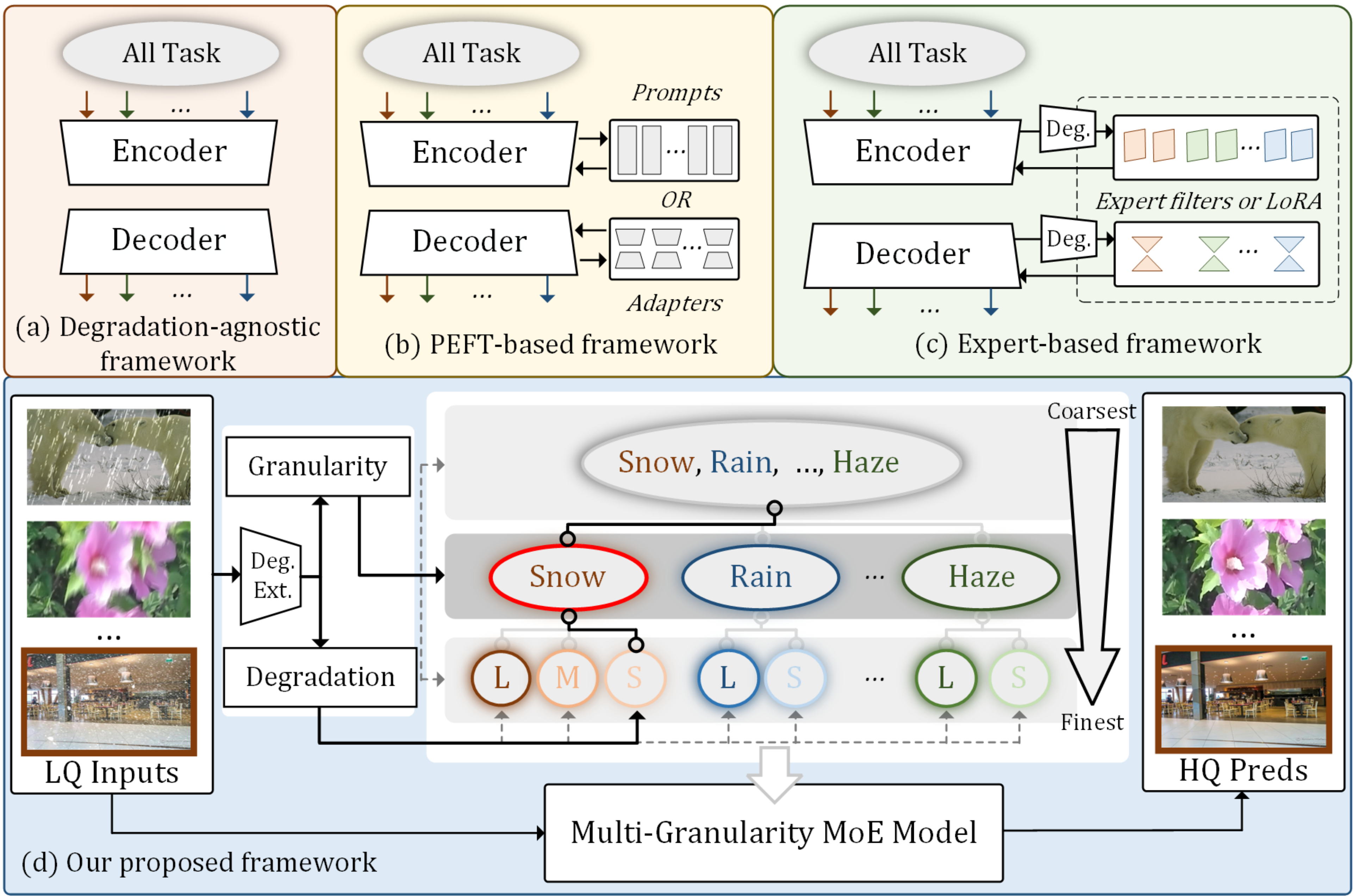}
	\caption{\small \small \textbf{Illustration of representative all-in-one image restoration frameworks}. 
 \textbf{(a)} Degradation-agnostic methods~\cite{wang2021realesrgan, BSRGAN} train a shared backbone using data from all tasks and are limited in leveraging degradation-specific restoration.
 \textbf{(b)} PEFT-based methods~\cite{AirNet, IDR, PromptIR, Transweather, OneRestore, mioir, MPerceiver} apply learnable prompts or adapters to the backbone to adapt various tasks.
 \textbf{(c)} Expert-based methods~\cite{HQ_50K, RestoreAgent, grids, tgsr, D2CSR, ADMS, HAIR, instructir, DASR, U_WADN, WGWSNet} train specific MoE, LoRA, or filters for specific tasks. 
 However, both (b) and (c) suffer from the inevitable error in degradation estimation.
 \textbf{(d)} We construct a multi-granularity degradation set and a multi-granularity MoE restoration model.
 By adaptively estimating image degradation at proper granularity, our UniRestorer can be effective in leveraging degradation-specific restoration while being robust to degradation estimation error. 
 }
\label{fig:introduction_type_of_framework}
\vspace{-3.5mm}
\end{figure*}
Image restoration is a fundamental task in computer vision, it aims to restore high-quality (HQ) images from corresponding low-quality (LQ) counterparts, including denoising~\cite{zhang2021learning, sidd}, deblurring~\cite{zhang2022self, GoPro, realblur}, de-weathering~\cite{liu2024learning, Snow100K, test1200, test2800}, low-light enhancement\cite{zhang2024bracketing, LOL, lolv2}, etc. 
In the past decade, advanced deep neural architectures~\cite{ResNet, Transformer, liu2021Swin, zhang2023controlvideo, CLIP} have driven image restoration methods to evolve from task-specific~\cite{DnCNN, SRCNN, DeblurGANv2, PReNet, Dehazeformer, desnownet} to task-agnostic~\cite{MPRNet, SwinIR, NAFNet, Restormer,lin2024improving} backbones.
Recently, influenced by the success of foundation models in natural language processing and high-level vision, there has been increasing interest in addressing multiple image restoration tasks within a single framework, known as all-in-one image restoration~\cite{jiang2024survey}.
A straightforward approach is to utilize data from all tasks to train a degradation-agnostic model (see Fig.~\ref{fig:introduction_type_of_framework} (a)), but it easily yields unsatisfactory results due to the conflicts between certain tasks (\eg, denoising and deblurring).
Inspired by parameter-efficient fine-tuning (PEFT) methods~\cite{prompt, adapter}, 
learnable prompts~\cite{AirNet, IDR, PromptIR, Transweather, OneRestore, mioir} and adapters~\cite{MPerceiver} are integrated into the restoration model to modulate intermediate features (see Fig.~\ref{fig:introduction_type_of_framework} (b)), allowing the model to process different tasks more specifically.
Furthermore, some methods introduce mixture-of-experts (MoE)~\cite{HQ_50K, RestoreAgent, grids, tgsr, D2CSR} (MoE-like using filters~\cite{ADMS, HAIR, instructir, DASR, U_WADN, WGWSNet} and low-rank adaptation~\cite{InstructIPT, lorair}) modules, routing the current feature to specific experts based on deep features and estimated degradation (see Fig.~\ref{fig:introduction_type_of_framework} (c)).

Despite rapid progress, these methods show limited performance and fall significantly behind specific single-task models. 
Degradation-agnostic models cannot leverage the benefits from degradation estimation-based priors, limiting their effectiveness in all-in-one image restoration.
PEFT-based and MoE-based methods provide degradation-aware ways for processing different corruptions by conditioning on degradation estimation. 
However, their effectiveness heavily relies on accurate degradation estimation, which is inherently challenging due to the vast and diverse degradation space of all-in-one image restoration.
Consequently, degradation-aware models are inevitably vulnerable to estimation errors, which can lead to inappropriate prompt or expert routing, ultimately degrading restoration quality.
From the above analysis, one plausible solution to improve all-in-one restoration is to estimate and leverage image degradation while being robust to degradation estimation error. 
To this end, we propose to represent image degradation at multiple levels of granularities rather than a single-level, and construct a multi-granularity degradation representation (DR) set by hierarchizing the degradation space.
We then build a corresponding multi-granularity MoE restoration model, where fine-grained experts specialize in specific degradations to enhance restoration accuracy, while coarse-grained experts generalize across broader degradation spaces.
To enable more robust expert routing for a degraded image, beyond vanilla degradation estimation, we introduce granularity estimation to indicate the degree of degradation estimation error.
For example, in Fig.~\ref{fig:introduction_type_of_framework} (d), given an image suffering from large snow degradation, if the degradation estimation is ambiguous, the granularity estimation will assign a coarser-grained expert, \eg, a model for general snow, to alleviate the effect of estimation error. Conversely, when the degradation estimation is confident, granularity estimation will assign a finer-grained expert to maximize the restoration quality, \eg, a model for large snow. 
By jointly leveraging both degradation and granularity estimation, our approach can mitigate the adverse effects of inaccurate degradation estimation, achieving high restoration and generalization performance simultaneously.

Specifically, we take the following steps to implement our UniRestorer.
First, we optimize a degradation extractor which is aware of finer-grained image degradation.
Second, all low-quality images in the training set are fed into the degradation extractor, and their features are then hierarchically clustered into multiple DR groups at different granularities, respectively, which forms the multi-granularity degradation set.
Third, we train a multi-granularity MoE restoration model based on the constructed multi-granularity degradation set.
Fourth, given an image with unknown degradation, we learn routers based on both degradation and granularity estimation results. The router identifies which DR group the degradation and granularity best match in the degradation set, and then allocates the corresponding expert for inference.

\begin{figure*}[t]
	\centering
	\setlength{\abovecaptionskip}{0.1cm}
	\setlength{\belowcaptionskip}{-0.3cm}
	\includegraphics[width=0.98\textwidth]{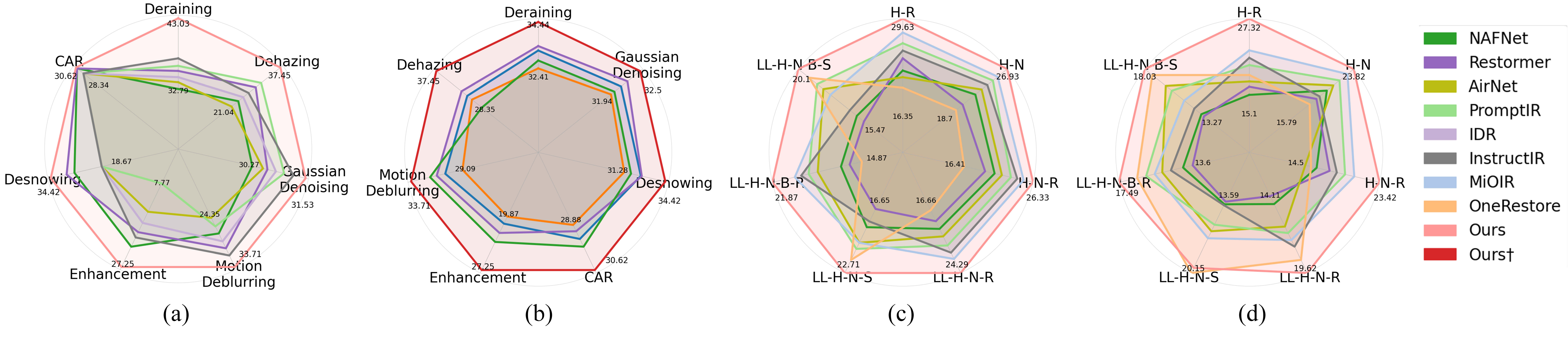}
	\caption{\small \small \textbf{Comparisons with task-agnostic methods, all-in-one methods, and single-task models.}
    Ours and Ours\dag denote \textit{auto} and \textit{instruction} modes, which are respectively used for a fair comparison with all-in-one and specific single-task models.
    \textbf{(a)} Comparisons on single-degradation all-in-one setting.
    \textbf{(b)} Comparisons with specific single-task models.
    \textbf{(c)} Comparisons on mixed-degradation all-in-one (in-of-distribution) setting.
    \textbf{(d)} Comparisons on mixed-degradation all-in-one (out-of-distribution) setting.}
	\label{fig:comparisons_radar}
\vspace{-3.5mm}
\end{figure*}

To demonstrate the effectiveness of the proposed method, we conduct experiments on both single-degradation and mixed-degradation scenarios.
For the single-degradation scenario, we unify seven widely studied image restoration tasks. For the mixed-degradation scenario, we introduce six types of degradations.
In addition, we conduct a comprehensive comparative evaluation.
The results indicate that our method not only significantly outperforms existing all-in-one methods but also achieves performance that is competitive with, or even superior to, specific single-task models (see Fig.~\ref{fig:comparisons_radar}).

The main contributions can be summarized as follows:
\begin{itemize}

    \item In this paper, we propose UniRestorer, a novel all-in-one image restoration framework that can estimate and leverage robust degradation representation to adaptively handle diverse image corruptions.
    
    \item We propose to train a multi-granularity MoE restoration model, where finer-experts specialize in specific degradations to enhance restoration accuracy, while coarser-experts generalize on broader degradation spaces to ensure generalization ability.

    \item To alleviate the impact of degradation estimation error on expert routing, we propose to estimate a proper granularity for image degradation. By jointly leveraging degradation and granularity estimation, our method adaptively allocates the most appropriate expert for different corruptions.

    \item Extensive experiments are conducted on single-degradation and mixed-degradation scenarios, showing that our method significantly outperforms existing all-in-one methods and effectively closes the performance gap with specific single-task models.
\end{itemize}


\section{Related Work} 
\label{sec:related_work}
\textbf{Image Restoration for Specific Tasks}\,\,
Pioneer works propose tailored frameworks for specific degradations, \eg, DnCNN~\cite{DnCNN} for denoising, SRCNN~\cite{SRCNN} for superresolution, DeblurGANv2~\cite{DeblurGANv2} for deblurring, PReNet~\cite{PReNet} for draining, Dehazeformer~\cite{Dehazeformer} for dehazing, and DesnowNet~\cite{desnownet} for desnowing.
Benefits from advanced deep neural networks, some works tend to develop task-agnostic models, such as MPRNet~\cite{MPRNet}, SwinIR~\cite{SwinIR}, NAFNet~\cite{NAFNet}, and Restormer~\cite{Restormer}, which become strong baselines in various image restoration tasks.

\noindent\textbf{PEFT-based All-in-One Image Restoration}\,\,
To adapt to various corruptions, pioneer all-in-one image restoration methods~\cite{PromptIR, IDR, Transweather, AirNet, mioir, MPerceiver} integrate PEFT-based techniques~\cite{prompt, adapter} and feature modulation~\cite{SRMD} into the restoration backbone.
For example, 
TransWeather~\cite{Transweather} and AirNet~\cite{AirNet} learn degradation-related embeddings as prompts to modulate deep features based on different corrupted inputs.
Similarly, PromptIR~\cite{PromptIR} and IDR~\cite{IDR} employ prompt learning in a more powerful backbone~\cite{Restormer} and achieve state-of-the-art performance.
MiOIR~\cite{mioir} further exploits the benefits of incorporating explicit prompt and adaptive prompt learning.
\noindent\textbf{Expert-based All-in-One Image Restoration}\,\,
Mixture-of-Experts (MoE) has proven to be a solution to multi-task learning due to its conditional processing capability~\cite{mod_squad, mthl, uniperceivermoe}.
Although prompt learning-based techniques work similarly to MoE in conditional processing, they are still limited in performance and task range.
Recent MoE-based all-in-one image restoration methods can be categorized into implicit MoE~\cite{HQ_50K, instructir, U_WADN, HAIR} and explicit MoE~\cite{WGWSNet, ADMS, DaAIR, grids, InstructIPT, RestoreAgent, lorair}.
The experts in the former methods are implicitly learned in training and do not correspond to a specific task.
Methods belonging to explicit MoE learn experts for specific tasks, unleashing the power of additionally increased parameters. 
Based on a shared backbone, WGWSNet~\cite{WGWSNet} and ADMS~\cite{ADMS} learn specific filters for specific tasks.
To be more efficient, DaAIR~\cite{DaAIR}, InstructIPT~\cite{InstructIPT}, and LoRA-IR~\cite{lorair} learn individual LoRA weights or adapters for each task. By combining knowledge of the shared backbone and task-specific modules, these methods can adapt to different tasks effectively.
GRIDS~\cite{grids} and RestoreAgent~\cite{RestoreAgent} divide one complete degradation space into subspaces and learn specific expert full models to alleviate potential conflicts between different degradations.

Despite progress achieved, most existing methods learn DRs at a single, coarse granularity and rely solely on degradation estimation as the routing condition, making them vulnerable to estimation errors. In contrast, our approach explores DRs across multiple granularities, jointly utilizing coarse levels (the same as task-agnostic settings) and finer levels beyond those used in existing methods.
In addition, we incorporate degradation estimation and its error into route process, enabling more robust decision-making.

%

%

\section{Proposed Method}
\label{sec:proposed_method}

\begin{figure*}[t]
	\centering
	\setlength{\abovecaptionskip}{0.1cm}
	\setlength{\belowcaptionskip}{-0.3cm}
	\includegraphics[width=0.95\textwidth]{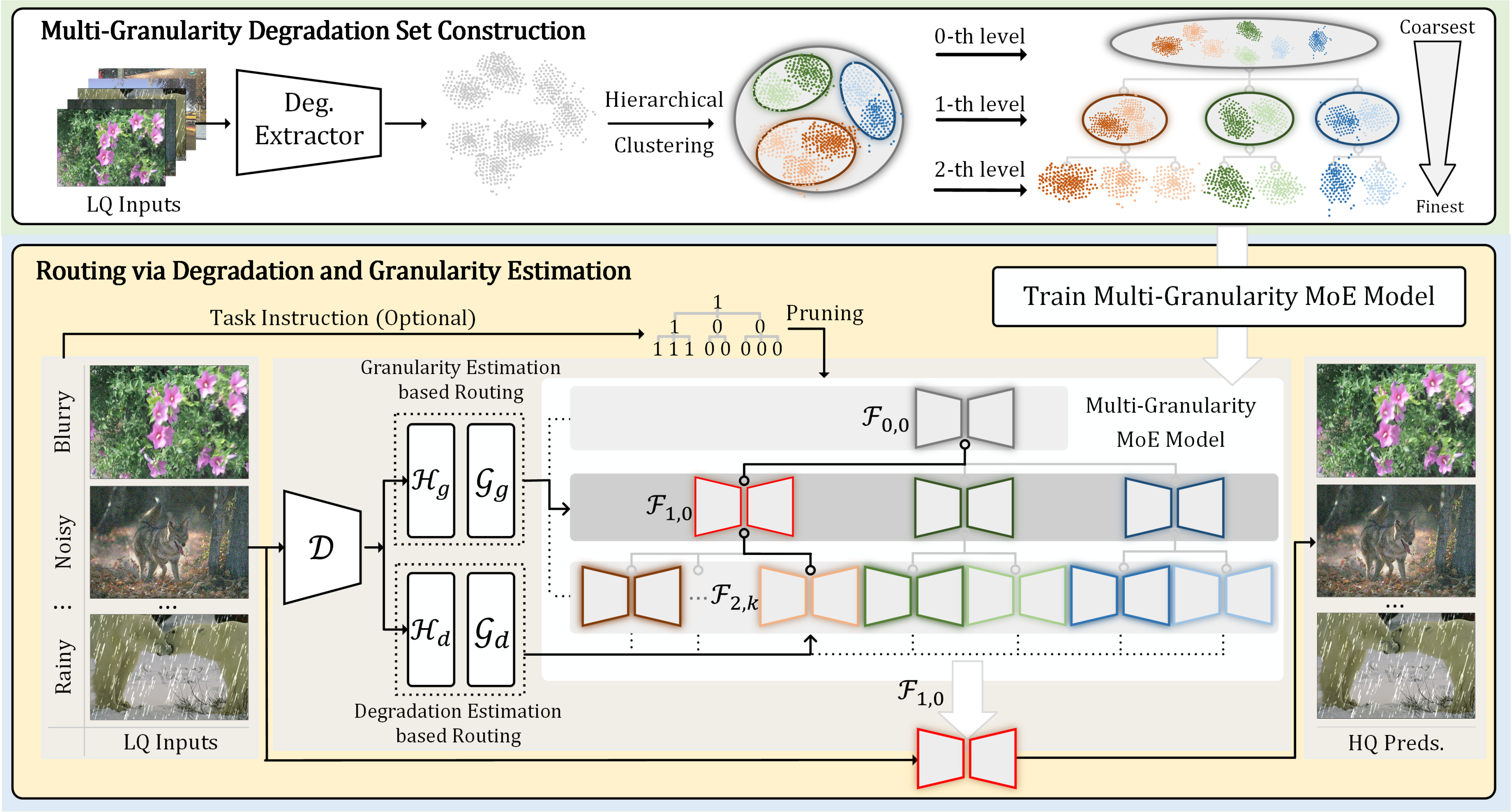}
	\caption{\small \small \textbf{Illustration of our proposed UniRestorer}. We develop a multi-granularity degradation set by hierarchical clustering on extracted DRs at different granularities. 
    Based on the multi-granularity degradation set, we train a multi-granularity MoE restoration model.
    Besides vanilla degradation estimation, we introduce granularity estimation to indicate the degree of degradation estimation error. Adopting both degradation and granularity estimation, we train routers to adaptively allocate an expert to unknown corrupted input.
    }
    \label{fig:framework}
\vspace{-3mm}
\end{figure*}

In this section, we introduce our all-in-one image restoration framework, UniRestorer.
We first optimize one degradation representation (DR) extractor that learns DRs at a fine-grained level.
Based on DRs extracted from all low-quality training data, we conduct hierarchical data clustering that separates DRs into multiple DR groups at different granularities (see Fig.~\ref{fig:framework}), leading to a multi-granularity degradation set and MoE restoration model.
To alleviate the inevitable error caused by degradation estimation in routing, we adopt both degradation and granularity estimation to allocate the most appropriate expert to given unknown corrupted inputs.
\subsection{Preliminaries}
\label{sec:preliminaries}
Mixture-of-Experts~\cite{sparse_gated_moe, switch_transformer, gshard, st_moe, soft_moe, deepspeed_moe} is a promising way for scaling up and deploying large or gigantic models due to its competitive computational bounds and latency compared to the vanilla single model. 
Expert network set $\{\mathcal{F}\}$ and routing network $\mathcal{G}(\cdot)$ are two basic components of MoE architecture. The expert network can be a sub-layer or a sub-network (\eg, convolution layer or feed-forward layer), and it also can be an independently trained network.
Given input $\mathbf{X}$, the routing network $\mathcal{G}(\cdot)$ is used to calculate the contribution of each expert to the final output,
\begin{equation}
\setlength{\abovedisplayskip}{1pt}
\setlength{\belowdisplayskip}{1pt}
    y = \sum_{i=0}^N \mathcal{G}(h) \cdot \mathcal{F}_i(\mathbf{X}),
\label{eq:moe}
\end{equation}
where $h$ is commonly a type of representation of input $\mathbf{X}$ which the routing network can assign expert conditioning on.
The routing network is typically a noisy top-k routing network parameterized with $w_{g}$ and $w_{n}$, which models $P(\mathcal{F}_k|h)$ as the probability of using the $k$-th expert $\mathcal{F}_k$ and selects experts with top-k probability to contribute to the final output. The noisy routing process can be described as follows,
\begin{equation}
\setlength{\abovedisplayskip}{1pt}
\setlength{\belowdisplayskip}{1pt}
\begin{small}
    \mathcal{G}(h)
    =
    \operatorname{Softmax}(h \cdot w_g) + \mathcal{N}(0, 1) \operatorname{Softplus}(h \cdot w_n),
\end{small}
\label{eq:moe-rout}
\end{equation}
where $\operatorname{Softplus}$ is the smooth approximation to the function $\operatorname{ReLU}$. Then, the $\operatorname{TopK}(\cdot, k)$ function is used to select the experts with the largest $k$ values according to $\mathcal{G}(h)$.

\subsection{Multi-Granularity Degradation and Expert}
\label{sec:multi_granularity_degradtion_representation}
\noindent\textbf{Fine-Grained DR Extractor}\,\,
Considering the convenience of using text to represent the type of degradation on corrupted images,
we adopt a CLIP-based method, DA-CLIP~\cite{daclip}, as our basic backbone.
Using contrastive learning, DA-CLIP trains an image controller to separate different corrupted inputs.
However, aiming to exploit finer-grained DRs, it is essential to enable the DR extractor to be aware of finer-grained representation of degradation.
Motivated by the above, we train a fine-grained DR extractor to fit our goal.
We construct a dataset with fine-grained level degradations.
In our experiments, we found that the estimation of degradation degree is sensitive to distortions, \eg, noise can be heavily alleviated by transformations in the pre-processing of CLIP.
To alleviate the effect of distortion on the original degradation, 
In training, we first crop high-resolution clean images into $224 \times 224$ patches and then add synthetic degradations with different types and degrees to obtain low-quality inputs.
During inference, we center-crop the main body of given low-quality inputs to alleviate the influence of the resolution gap between training and inference.
Following~\cite{daclip}, we adopt the same training setups to train our DR extractor.
The fine-grained DR $e$ can be learned by the DR extractor $\mathcal{D}$, \ie,
\begin{equation}
\setlength{\abovedisplayskip}{1pt}
\setlength{\belowdisplayskip}{1pt}
    e = \mathcal{D}(\mathbf{X}).
\end{equation}
When training is finished, we employ $\mathcal{D}$ to obtain the DR set $\{e\}$ from all training samples.
\noindent\textbf{Hierarchical Degradation Clustering}\,\,
To construct the proposed multi-granularity degradation set, we utilize an unsupervised data clustering approach~\cite{mitchell1997machine}, which is simple yet effective. 
The constructed degradation set is hierarchical. 
The $i$-th level of granularity consists of non-overlapped DR groups, each of which is related to a data distribution, and samples drawn from one distribution are with similar DRs.
In our implementation, we construct a hierarchical degradation set with $n$ levels of granularities ($n$$=$$3$). 
The coarsest level is $0$-th level and the finest level is $($$n$$-$$1$$)$-th level.
We hierarchically cluster training samples from top to down.
For example, we first conduct data clustering at the coarsest-grained level by,
\begin{equation}
\setlength{\abovedisplayskip}{1pt}
\setlength{\belowdisplayskip}{1pt}
  \{u\} \leftarrow \operatorname{K-means} (\{e\}),
  \label{equ:kmeans}
\end{equation}
where $\{u\}$ is the set of cluster centers of finer-grained level. Based on $\{u\}$, we use function $\operatorname{argmin}_{u_{j} \in \{u\}}(||e-u_{j}||_2^2)$ to decide which cluster the samples belong to, accordingly we separate data and $\{e\}$ into the same number of groups as cluster centers.
We hierarchically conduct the same process as Eq.~(\ref{equ:kmeans}) based on newly obtained DR groups until the finest-grained level.
As a result, we construct a multi-granularity degradation set, and we can separate training samples into non-overlapped groups from the coarsest-grained level to the finest-grained level. Referring to the total number of degradation types and degrees in training DR extractor, we empirically set $1$, $7$, and $19$ clusters for the $0$-th level to the $($$n$$-$$1$$)$-th level, respectively.

\noindent\textbf{Multi-Granularity MoE Restoration Model}\,\,
We train a multi-granularity MoE restoration model based on the obtained multi-granularity degradation set.
Expert networks are trained with $\ell_1$ loss as the reconstruction loss between predicted results $\hat{\mathbf{Y}}$ and target $\mathbf{Y}$, \ie,
\begin{equation}
\setlength{\abovedisplayskip}{1pt}
\setlength{\belowdisplayskip}{1pt}
    \ell_1 = ||\hat{\mathbf{Y}} - \mathbf{Y} ||_1.
\end{equation}
Experts trained at coarser-grained levels can generalize to a large range of degradations, and experts trained in finer-grained levels can highly improve the restoration performance on degradations that they are good at.
Therefore, given corrupted inputs, we can estimate their DRs and assign them to the most appropriate expert.
Note that our main contribution lies in the design of estimating and leveraging robust DRs to improved all-in-one image restoration performance. To achieve this, we adopt MoE framework, where we follow prior works such as GRIDS~\cite{grids} and RestoreAgent~\cite{RestoreAgent}, employing SOTA image restoration networks as experts. Though straightforward, this strategy is simple and effective in practice. Moreover, our approach is orthogonal to the specific design of the MoE structure and can be readily combined with lightweight expert architectures or efficient fine-tuning techniques, such as LoRA~\cite{lora,adapter}.
More discussion can be found in Sec.~\textcolor{red}{C} of \textit{Suppl.}
\subsection{Routing via Degradation and Granularity Estimation}
\label{sec:confidence_aware_router}
It makes sense to use estimated DRs as the routing condition $h$ in Eq.~(\ref{eq:moe}).
However, degradation estimation is error-prone, which may lead to inaccurate routing and thus impede restoration performance.
To this end, based on the constructed multi-granularity degradation set, we estimate image degradation at the proper granularity level, and perform routing based on both degradation and granularity estimation, which makes our method robust to degradation estimation error.
We train two separate branches ($\mathcal{H}_d$ and $\mathcal{H}_g$) upon $\mathcal{D}$ for degradation estimation vector $e_{deg}$ at the finest-grained level and granularity estimation vector $e_{gran}$.
Then, we further train two routers $\mathcal{G}_{d}(\cdot)$ and $\mathcal{G}_{g}(\cdot)$ as Eq.~(\ref{eq:moe-rout}) based on degradation and granularity estimation, respectively.
The routing process can be described as follows,
\begin{equation}
\setlength{\abovedisplayskip}{1pt}
\setlength{\belowdisplayskip}{1pt}
\begin{aligned}
    \mathcal{F}_{n-1,k}&=\operatorname{TopK}(\mathcal{G}_{d}(e_{deg})|\{\mathcal{F}\}_{n-1}) \\
    \mathcal{F}_{i,j}&=\operatorname{TopK}(\mathcal{G}_{g}(e_{gran})|\{\mathcal{F}_{0,0},\cdots,\mathcal{F}_{n-1,k}\}).
\label{equ:routing}
\end{aligned}
\end{equation}
$\mathcal{G}_{d}$ is conducted in the finest-grained level. Based on its routing result, it further selects a set of experts that can solve degradations belonging to DR group $\{e\}_{n-1,k}$ and are trained in different granularity levels from $0$ to $n$$-$$1$.
Furthermore, $\mathcal{G}_{g}$ is responsible for selecting an expert training in the proper granularity according to granularity estimation.
Inspired by data uncertainty learning~\cite{Chang_2020_CVPR}, we learn $e_{deg}$ and $e_{gran}$ by introducing loss as,
\begin{equation}
\setlength{\abovedisplayskip}{1pt}
\setlength{\belowdisplayskip}{1pt}
    \mathcal{L}_{dg} = \frac{1}{2e_{gran}}(u_{y_i} - e_{deg})^2 + \frac{1}{2}\operatorname{ln}e_{gran},
    \label{equ:uncertainty_loss}
\end{equation}
where $u_{y_i}$ is the weight center in the finest-grained level obtained by Eq.~(\ref{equ:kmeans}). 
We use the first term to build the relationships of $e_{deg}$ and $e_{gran}$, when the distance between current degradation estimation and corresponding weight center is high, a larger $e_{gran}$ is used to ensure that $\mathcal{L}_{dg}$ is not too high. Therefore, the granularity can be seen as the degree of the current degradation estimation error. 
If estimated with a large error, granularity estimation will allocate a coarser-grained expert; Conversely, if degradation estimation is precise, granularity estimation will allocate a finer-grained expert.
Thus, conditioning on both $e_{deg}$ and $e_{gran}$, we can dynamically allocate given corrupted inputs with the most appropriate expert.
For stable training, and to avoid the result of $\mathcal{G}_{g}$ collapse to the finest-grained level, we further adopt load-balance loss as used in~\cite{sparse_gated_moe},
\begin{equation}
\setlength{\abovedisplayskip}{1pt}
\setlength{\belowdisplayskip}{1pt}
    \mathcal{L}_{load} = {\sigma}/{\mu},
    \label{equ:load_balance}
\end{equation}
where $\mu$ and $\sigma$ are mean and standard deviation of experts load in granularity level, respectively.
At last, we train our routers by jointly adopting loss as,
\begin{equation}
\setlength{\abovedisplayskip}{1pt}
\setlength{\belowdisplayskip}{1pt}
    \mathcal{L}_{total} = \ell_1 + \alpha \mathcal{L}_{dg} + \beta \mathcal{L}_{load},
\label{equ:total_loss}
\end{equation}
where $\alpha$ and $\beta$ are two hyper-parameters to trade-off the effect of $\mathcal{L}_{dg}$ and $\mathcal{L}_{load}$.

\textbf{Instruction Mode}\,\,
Typically, our framework automatically conducts the processing pipeline of extraction-recognition-routing-restoration (\textit{auto} mode). When users are convinced about the type of degradation, we provide options that users can provide the type of degradation to instruct the routing process (\textit{instruction} mode), as shown in Fig.~\ref{fig:framework}. The instruction plays the role of pruning, it utilizes a set of masks to prune the corresponding DR groups from the multi-granularity degradation set, leading to a more efficient and accurate routing.

\begin{table*}[!t]
\caption{\small \small \textbf{All-in-One image restoration (single-degradation) results}. `$n$T' means the average performance of $n$ tasks (3T: Derain, Dehaze, Denoise; 5T: Derain, Dehaze, Denoise, Deblur, Lowlight; 7T: all).}
\vspace{-1mm}
\label{table:single_in_one_image_restoration_results}
\centering
\resizebox{\textwidth}{!}{%
\setlength{\tabcolsep}{3pt}
\begin{tabular}{ l  c c  c c  c c  c c  c c  c c  c c  c c c}
\toprule[0.15em]

    & \multicolumn{2}{c}{\textbf{Derain}}          
    & \multicolumn{2}{c}{\textbf{Dehaze}}        
    & \multicolumn{2}{c}{\textbf{Denoise}}      
    & \multicolumn{2}{c}{\textbf{Deblur}} 
    & \multicolumn{2}{c}{\textbf{Lowlight}}  
    & \multicolumn{2}{c}{\textbf{Desnow}}           
    & \multicolumn{2}{c}{\textbf{CAR}}  
    & \multicolumn{3}{c}{\textbf{Average}}
    \\
    
    & \multicolumn{2}{c}{Rain100L\cite{Rain200H}} 
    & \multicolumn{2}{c}{SOTS\cite{reside}} 
    & \multicolumn{2}{c}{BSD68\cite{bsd68}} 
    & \multicolumn{2}{c}{GoPro\cite{GoPro}} 
    & \multicolumn{2}{c}{LOLv1\cite{LOL}} 
    & \multicolumn{2}{c}{Snow100K\cite{Snow100K}}
    & \multicolumn{2}{c}{LIVE1\cite{live1}} 
    & 3T & 5T & 7T
    \\ 

        \textbf{Methods} 
            & PSNR & SSIM
            & PSNR & SSIM
            & PSNR & SSIM
            & PSNR & SSIM
            & PSNR & SSIM
            & PSNR & SSIM
            & PSNR & SSIM
            & \multicolumn{3}{c}{PSNR}
            \\
            
\midrule[0.05em]
    MPRNet~\cite{MPRNet}             & 25.56 & 0.903 & 16.37 & 0.806 & 23.92 & 0.702 & 25.40 & 0.858 & 7.75  & 0.124 & 19.33 & 0.740 & 27.77 & 0.889 & 18.51 & 22.68 & 19.66 \\
    SwinIR~\cite{SwinIR}             & 21.20 & 0.837 & 15.94 & 0.775 & 24.12 & 0.780 & 16.86 & 0.735 & 13.80 & 0.409 & 18.60 & 0.718 & 15.61 & 0.746 & 17.55 & 17.09 & 18.45 \\
    NAFNet~\cite{NAFNet}             & 31.32 & 0.967 & 22.43 & 0.907 & 30.27 & 0.918 & 26.53 & 0.877 & \underline{23.09} & 0.796 & 26.33 & 0.889 & \textbf{31.26} & \textbf{0.946} & 24.55 & 25.76 & 26.28 \\
    Restormer~\cite{Restormer}       & 34.25 & 0.981 & 29.64 & 0.971 & 31.00 & \underline{0.929} & 27.50 & 0.896 & 22.80 & 0.821 & \underline{28.24} & \underline{0.914} & \underline{31.21} & \underline{0.946} & 30.46 & 28.56 & \underline{28.27} \\
\midrule[0.05em]
    DL-3T~\cite{DL}                     & 32.62 & 0.931 & 26.92 & 0.391 & 30.12 & 0.838 & - & - & - & - & - & - & - & - & 28.09 & - & - \\
    AirNet-3T~\cite{AirNet}             & 34.90 & 0.967 & 27.94 & 0.962 & 31.06 & 0.872 & - & - & - & - & - & - & - & - & 29.29 & - & - \\
    PromptIR-3T~\cite{PromptIR}         & 36.37 & 0.972 & 30.58 & 0.974 & 31.12 & 0.873 & - & - & - & - & - & - & - & - & 31.50 & - & - \\
    InstructIR-3T~\cite{instructir}  & 37.98 & 0.978 & 30.22 & 0.959 & 31.32 & 0.875 & - & - & - & - & - & - & - & - & 31.49 & - & - \\
    DaAIR-3T~\cite{DaAIR} & 37.10 & 0.978 & \underline{32.30} & \underline{0.981} & 31.06 & 0.868 & - & - & - & - & - & - & - & - & 32.51 & - & - \\
    PromptIR-TUR-3T~\cite{tur} & \underline{38.57} & \underline{0.984} & 31.17 & 0.978 & 31.19 & 0.872 & - & - & - & - & - & - & - & - & 32.67 & - & - \\
    MoCEIR-3T~\cite{moce} & \underline{38.57} & \underline{0.984} & 31.34 & 0.979 & 31.24 & 0.873 & - & - & - & - & - & - & - & - & \underline{32.73}  & - & - \\
\midrule[0.05em]
    DL-5T~\cite{DL}                     & 21.96 & 0.762 & 20.54 & 0.826 & 23.09 & 0.745 & 19.86 & 0.672 & 19.83 & 0.712 & - & - & - & - & 21.02 & 20.28 & -\\
    Transweather-5T~\cite{Transweather} & 29.43 & 0.905 & 21.32 & 0.885 & 29.00 & 0.841 & 25.12 & 0.757 & 21.21 & 0.792 & - & - & - & - & 23.31 & 24.41 & -\\
    TAPE-5T~\cite{TAPE}                 & 29.67 & 0.904 & 22.16 & 0.861 & 30.18 & 0.855 & 24.47 & 0.763 & 18.97 & 0.621 & - & - & - & - & 24.10 & 24.28 & -\\
    AirNet-5T~\cite{AirNet}             & 32.98 & 0.951 & 21.04 & 0.884 & 30.91 & 0.882 & 24.35 & 0.781 & 18.18 & 0.735 & - & - & - & - & 23.82 & 24.10 & -\\
    IDR-5T~\cite{IDR}                   & 35.65 & 0.965 & 25.20 & 0.938 & 31.09 & 0.883 & 26.65 & 0.810 & 20.70 & 0.820 & - & - & - & - & 27.36 & 26.86 & -\\
    InstructIR-5T~\cite{instructir}  & 36.84 & 0.973 & 27.10 & 0.956 & \underline{31.40} & 0.887 & 29.40 & 0.886 & 23.00 & 0.836 & - & - & - & - & 28.99 & 29.19 & -\\
    DaAIR-5T~\cite{DaAIR} & 36.28 & 0.975 & 31.97 & 0.980 & 31.07 & 0.878 & 29.51 & 0.890 & 22.38 & 0.825  & - & - & - & - & 32.20 & 30.24 & - \\ 
    Transweather-TUR-5T~\cite{tur} & 33.09 & 0.952 & 29.68 & 0.966 & 30.40 & 0.869 & 26.63 & 0.815 & 23.02 & 0.838 & - & - & - & - & 29.96 & 28.56 & - \\ 
    DCPT-PromptIR-5T~\cite{dcpt} & 37.32 & 0.978 & 30.72 & 0.977 & 31.32 & 0.885 & 28.84 & 0.877 & 23.35 & 0.840   & - & - & - & - & 31.46 & 30.31 & - \\
    MoCEIR-5T~\cite{moce} & 38.04 & 0.982 & 30.48 & 0.974 & 31.34 & 0.887 & \underline{30.05} & \underline{0.899} & 23.00 & \underline{0.852} & - & - & - & - & 31.39 & \underline{30.58} & - \\
\midrule[0.05em]
    Ours    & \textbf{41.68} & \textbf{0.996} & \textbf{36.44} & \textbf{0.984} & \textbf{31.43} & \textbf{0.936} & \textbf{31.48} & \textbf{0.948} & \textbf{26.67} & \textbf{0.863} & \textbf{31.12} & \textbf{0.940} & 30.57 & 0.926 & \textbf{36.71} & \textbf{33.38} & \textbf{31.34} \\
    
    
\bottomrule[0.15em]
\end{tabular}
}
\vspace{-5mm}
\end{table*}

\begin{table*}[!t]
\caption{\small \small \textbf{All-in-One image restoration (mixed-degradation) results}. The performance is evaluated on DIV2K-valid with \textit{in-dist.} and \textit{out-dist.} degradation parameters. `$n$T' means the average performance of $n$ tasks (3T: H-R, H-N, H-N-R; 5T: H-R, H-N, H-N-R, LL-H-N-R, LL-H-N-B-R; 7T: all).}
\label{table:multiple_in_one_image_restoration_results}
\vspace{-1mm}
\centering
\resizebox{\textwidth}{!}{%
\setlength{\tabcolsep}{3pt}
\begin{tabular}{l  cc  cc  cc  cc  cc  cc  cc  ccc}
\toprule[0.15em]

    & \multicolumn{2}{c}{\textbf{H-R}}
    & \multicolumn{2}{c}{\textbf{H-N}}
    & \multicolumn{2}{c}{\textbf{H-N-R}}
    & \multicolumn{2}{c}{\textbf{LL-H-N-R}}
    & \multicolumn{2}{c}{\textbf{LL-H-N-S}}
    & \multicolumn{2}{c}{\textbf{LL-H-N-B-R}}
    & \multicolumn{2}{c}{\textbf{LL-H-N-B-S}}
    & \multicolumn{3}{c}{\textbf{Average}} 
    \\

    & &
    & &
    & &
    & &
    & &
    & &
    & &
    & 3T & 5T & 7T
    \\
    
                \textbf{Methods}  
            & PSNR & SSIM
            & PSNR & SSIM
            & PSNR & SSIM
            & PSNR & SSIM
            & PSNR & SSIM
            & PSNR & SSIM
            & PSNR & SSIM
            & \multicolumn{3}{c}{PSNR}
            \\ 
            
\midrule[0.05em]

    \multicolumn{18}{c}{\textit{In-distribution}}  \\
            
\midrule[0.05em]
    MPRNet~\cite{MPRNet}  & 17.22 & 0.691 & 19.02 & 0.717 & 17.18 & 0.639 & 17.25 & 0.549 & 17.18 & 0.583 & 16.99 & 0.466 & \underline{16.80} & 0.506 & 17.80 & 17.57 & 17.37  \\
    SwinIR~\cite{SwinIR}  & 16.35 & 0.620 & 18.74 & 0.693 & 16.48 & 0.585 & 16.78 & 0.513 & 16.73 & 0.555 & 16.54 & 0.434 & 16.10 & 0.482 & 17.19 & 17.01 & 16.81 \\
    NAFNet~\cite{NAFNet}  & 19.14 & 0.775 & 22.71 & 0.843 & 18.56 & 0.723 & 17.41 & 0.603 & 17.09 & 0.631 & 16.37 & 0.505 & 15.83 & \underline{0.530} & 20.13 & 18.98 & 18.15 \\
    Restormer~\cite{Restormer} & 19.85 & 0.795 & 20.44 & 0.820 & 17.64 & 0.719 & 16.86 & 0.588 & 16.56 & 0.623 & 15.92 & 0.483 & 15.47 & 0.520 & 19.31 & 18.27 & 17.53 \\
    
\midrule[0.05em]
    AirNet~\cite{AirNet}          & 18.67 & 0.752 & 23.34 & 0.855 & 19.03 & 0.726 & - & - & - & - & - & - & - & - & 20.34 & - & -\\
    PromptIR~\cite{Transweather}  & 22.20 & 0.825 & 24.23 & \textbf{0.886} & 20.86 & 0.772 & - & - & - & - & - & - & - & - & 22.43 & - & - \\
    MiOIR~\cite{mioir}            & \underline{24.58} & \underline{0.892} & \underline{24.46} & 0.879 & \underline{22.42} & \underline{0.818} & \underline{19.46} & 0.661 & - & - & \underline{18.12} & 0.561 & - & - & \underline{23.82} & \underline{21.80} & - \\
    InstructIR~\cite{instructir}  & 21.57 & 0.827 & 23.80 & 0.838 & 21.14 & 0.775 & 19.21 & \underline{0.666} & - & - & 18.06 & \underline{0.564} & - & - & 22.17 & 20.75 & - \\
    OneRestore~\cite{OneRestore}  & 17.35 & 0.711 & 18.70 & 0.669 & 16.41 & 0.617 & 16.66 & 0.621 & 20.24 & \underline{0.769} & - & - & - & - & 17.48 & - & - \\
    
\midrule[0.05em]
    Ours     & \textbf{29.63} & \textbf{0.950}& \textbf{26.93} & \underline{0.880} & \textbf{26.33} & \textbf{0.875} & \textbf{24.29} & \textbf{0.830} & \textbf{22.71} & \textbf{0.790} & \textbf{21.87} & \textbf{0.735} & \textbf{20.10} & \textbf{0.729} & \textbf{27.63} & \textbf{25.81} & \textbf{24.55} \\
    

\midrule[0.05em]

    \multicolumn{18}{c}{\textit{Out-of-distribution}}  \\

\midrule[0.05em]
    MPRNet~\cite{MPRNet}  & 15.71 & 0.636 & 16.38 & 0.598 & 15.57 & 0.560 &  14.84 & 0.345 & 14.59 & 0.355 & 14.58 & 0.286 & \underline{14.33} & 0.299 & 15.88 & 15.41 & 15.14 \\
    SwinIR~\cite{SwinIR}  & 15.10 & 0.556 & 15.79 & 0.553 & 14.50 & 0.486 & 14.72 & 0.341 & 14.32 & 0.353 & 14.29 & 0.282 & 13.95 & 0.298 & 15.13 & 14.88 & 14.66\\
    NAFNet~\cite{NAFNet}  & 16.77 & 0.713 & 19.68 & 0.751 & 15.11 & 0.595 & 14.40 & 0.385 & 14.05 & 0.390 & 13.89 & 0.317 & 13.27 & \underline{0.324} & 17.18 & 16.00 & 15.31\\
    Restormer~\cite{Restormer} & 17.44 & 0.723 & 17.31 & 0.738 & 17.44 & \underline{0.723} & 14.11 & 0.378 & 13.59 & 0.388 & 13.60 & 0.314 & 12.83 & 0.318 & 17.39 & 15.97 & 15.18\\
    
\midrule[0.05em]
    AirNet~\cite{AirNet}         & 17.50 & 0.732 & 20.16 & 0.777 & 15.07 & 0.380 & - & - & - & - & - & - & - & - & 17.57 & - & - \\
    PromptIR~\cite{Transweather} & 19.84 & 0.797 & 20.57 & 0.795 & 18.22 & 0.672 & - & - & - & - & - & - & - & - & 19.54 & - & - \\
    MiOIR~\cite{mioir}           & \underline{20.85} & \underline{0.850} & \underline{22.39} & \underline{0.821} & \underline{18.87} & 0.700 & 15.27 & 0.406 & - & - & \underline{14.70} & \underline{0.332} & - & - & \underline{20.70} & \underline{18.41} & -\\
    InstructIR~\cite{instructir}  & 20.33 & 0.825 & 19.02 & 0.636 & 17.74 & 0.597 & 15.34 & 0.386 & - & - & 14.52 & 0.296 & - & - & 19.03 & 17.39 & - \\
    OneRestore~\cite{OneRestore}  & 17.81 & 0.732 & 16.77 & 0.580 & 15.10 & 0.595 & \underline{17.49} & \underline{0.566} & \textbf{21.04} & \textbf{0.682} & - & - & - & - & 16.56 & - & - \\
    
\midrule[0.05em]
    Ours          & \textbf{27.32} & \textbf{0.937} & \textbf{23.82} & \textbf{0.811} & \textbf{23.42} & \textbf{0.801} & \textbf{20.21} & \textbf{0.661} & \underline{20.15} & \underline{0.631} & \textbf{17.49} & \textbf{0.529} & \textbf{18.03} & \textbf{0.543} & \textbf{24.85} & \textbf{22.45} & \textbf{21.49}\\
    
\bottomrule[0.15em]
\end{tabular}
}
\vspace{-7mm}
\end{table*}


\vspace{-4mm}

\section{Experiments}
\label{sec:experiments}
\subsection{Experimental Setup}
\label{sec:experimental_setup}

\textbf{Task and Dataset}\,\,
We train and evaluate our framework on single-degradation and mixed-degradation scenarios. 
In single-degradation scenario, we train a unified model to handle seven distinct image restoration tasks. We collect and integrate data from seven widely studied tasks. We adopt Rain100L~\cite{Rain200H} for deraining (`R'), RESIDE~\cite{reside} for dehazing (`H'), DFBW~\cite{div2k,flickr2k,bsd68,wed} for Gaussian color denoising (`N') and compression artifacts removal (`CAR'), GoPro~\cite{GoPro} for single-image motion deblurring (`B'), LOLv1~\cite{LOL} for low-light enhancement (`LL'), and Snow100K~\cite{Snow100K} for desnowing (`S').    
In mixed-degradation scenario, we develop a more complex degradation pipeline based on~\cite{BSRGAN}, which includes six types of degradation. We design seven challenging mixed degradation scenarios for evaluation. The training data is synthesized from clean images in DF2K~\cite{div2k,flickr2k}.
To demonstrate the generalization ability, we conduct comparisons on \textit{real-world} and \textit{unseen} degradations, including LHP~\cite{lhp} for real-world deraining, LOLv2~\cite{lolv2} for real-world low-light enhancement, RealSnow~\cite{Snow100K} for real-world desnowing, RainDrop~\cite{raindrop} for raindrop removal, TOLED~\cite{toled} for under-display camera image restoration, UIEB~\cite{uieb} for underwater image restoration, and out-of-distribution (\textit{out-dist.}) degradation parameters for mixed-degradation.
%
%
%
%
%
%



\begin{figure*}[t]
	\centering
	\setlength{\abovecaptionskip}{0.1cm}
	\setlength{\belowcaptionskip}{-0.3cm}
	\includegraphics[width=0.95\textwidth]{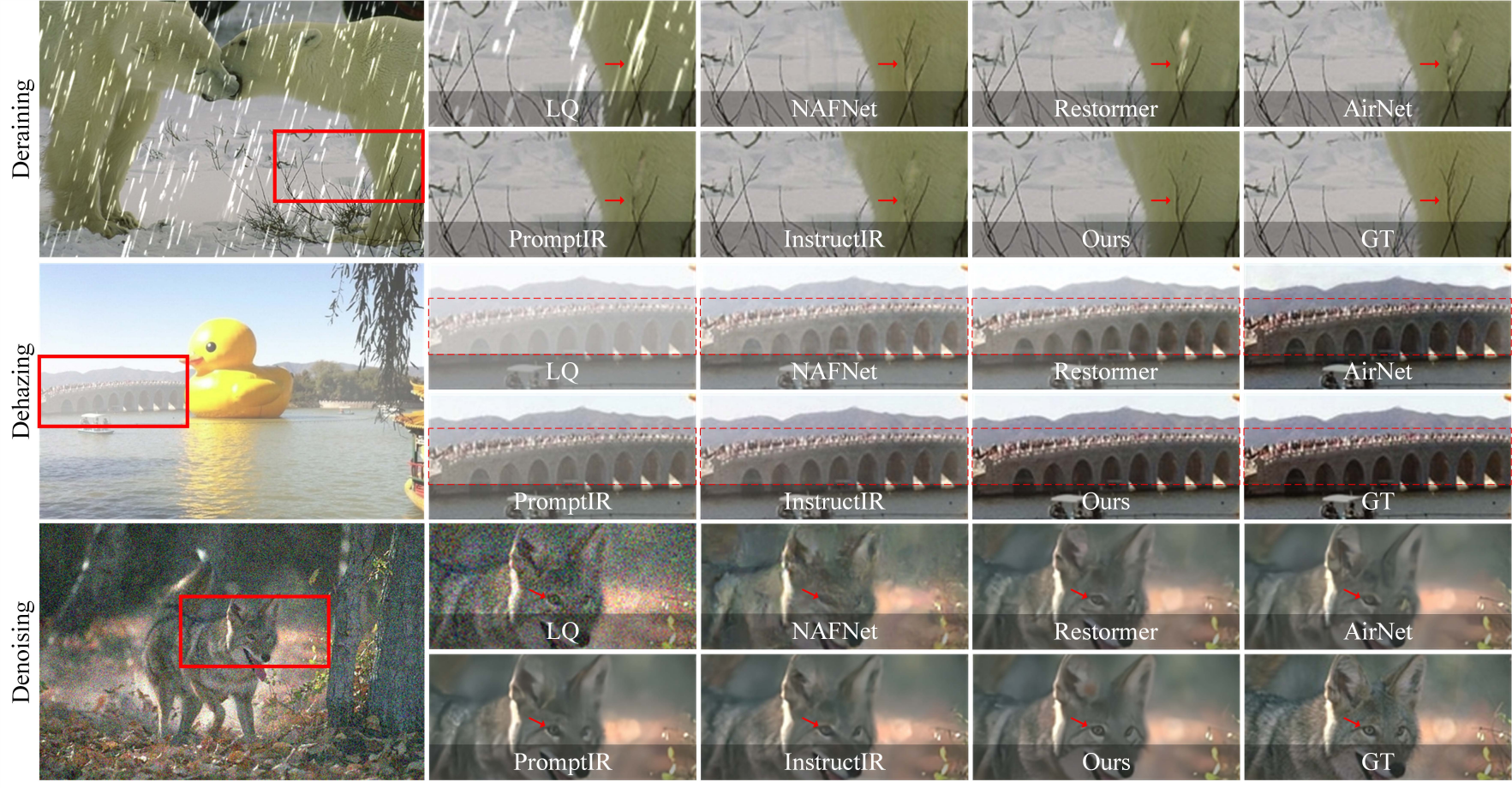}
	\caption{\small \small \textbf{Visual comparison of results on All-in-One image restoration (single-degradation)}. Our method can effectively remove the degradation pattern (\ie, rainstreak, haze, and noise), restore clearer texture (\ie, `branches' and `eyes' pointed by red arrows) and closer color (\ie, `bridge'). More results can be seen in \textit{Suppl.}}
	\label{fig:visual_comparisons_all_in_one_sio}
\end{figure*}   
\begin{table*}[!t]
\caption{\small \small \textbf{Compare to single-task image restoration models}. The task-agnostic methods are trained from scratch on each task.  \dag ~ denotes \textit{instruction} mode. PSNR scores are reported.}
\label{table:specific_image_restoration_results}
\vspace{-1mm}
\centering
\resizebox{1\textwidth}{!}{%
\setlength{\tabcolsep}{3pt}
\begin{tabular}{ l | c c c c | c c c | c c c | c c c c | c | c | c }
\toprule[0.15em]

     \textbf{Methods} 
    & \multicolumn{4}{c|}{\textbf{Derain}}          
    & \multicolumn{3}{c|}{\textbf{Denoise}}        
    & \multicolumn{3}{c|}{\textbf{Desnow}}      
    & \multicolumn{4}{c|}{\textbf{CAR.}} 
    & \textbf{Lowlight}
    & \textbf{Deblur}         
    & \textbf{Dehaze}  
    \\
    & 200H
    & 200L
    & DID
    & DDN

    & \multicolumn{3}{c|}{BSD68\cite{bsd68}}
    & \multicolumn{3}{c|}{Snow100K\cite{Snow100K}}
    & \multicolumn{4}{c|}{LIVE1\cite{live1}}
    
    & LOLv1 
    & GoPro 
    & SOTS 
    \\

    & \cite{Rain200H}
    & \cite{Rain200H}
    & \cite{test1200}
    & \cite{test2800}

    & $\sigma$=$15$
    & $\sigma$=$25$
    & $\sigma$=$50$

    & S
    & M
    & L

    & $q$=$10$
    & $q$=$20$
    & $q$=$30$
    & $q$=$40$
    
    & \cite{LOL}
    & \cite{GoPro}
    & \cite{reside}
    \\
    \midrule[0.05em]
    MPRNet~\cite{MPRNet} & \underline{31.53} & 35.92 & \underline{35.34} & \underline{34.19} & 34.91 & 32.67 & 29.62 & 36.22 & 34.49& 30.63 & 27.87 & 30.23 & \underline{31.57} & \underline{32.50} & 23.51 & 32.66 & 31.07  \\
    SwinIR~\cite{SwinIR} & 29.69 & 38.35 & 34.51 & 33.26 & 34.81 & 32.51 & 29.29 & 33.45 & 32.00 & 28.40 & 26.16 & 28.62 & 29.93 & 30.81 & 19.87 & 29.09 & 28.35 \\
    Restormer~\cite{Restormer} & 30.98 & \underline{40.33} & 35.26 & 34.16 & \underline{34.92} & \underline{32.70} & \underline{29.70} & \underline{36.50} & \underline{34.77} & \underline{30.86} & 27.89 & 30.21 & 31.56 & 32.48 & 23.66 & 32.92 & \underline{33.18} \\
    NAFNet~\cite{NAFNet} & 29.08 & 38.93 & 35.21 & 32.84 & 34.80 & 32.58 & 29.50 & 35.94 & 33.95 & 29.50 & \textbf{27.95} & \underline{30.24} & 31.56 & 32.47 & \underline{23.75} & \underline{33.69} & 22.70 \\
    
    \midrule[0.05em]
   
    Ours\dag & \textbf{32.01} & \textbf{41.61} & \textbf{35.50} & \textbf{34.40} & \textbf{35.00} & \textbf{32.75} & \textbf{29.75} & \textbf{36.94} & \textbf{35.18} & \textbf{31.14} & \underline{27.94} & \textbf{30.31} & \textbf{31.66} & \textbf{32.58} & \textbf{27.25} & \textbf{33.71} & \textbf{37.45} \\
    
\bottomrule[0.15em]
\end{tabular}
}
\vspace{-8mm}
\end{table*}
\begin{table*}[h]
\begin{minipage}[t]{0.49\textwidth}
\caption{\small \small \textbf{Generalization performance on real-world datasets}. All methods are evaluated without additional training. PSNR scores are reported.}
\vspace{-1mm}
\label{table:generalization_image_restoration_results_real-world}
\centering
\resizebox{\textwidth}{!}{%
\setlength{\tabcolsep}{3pt}
\begin{tabular}{ l  c   c  c  c }
\toprule[0.15em]


    & \textbf{Deraining}
    & \textbf{Low-light}
    & \textbf{Desnowing}
    & \multirow{2}{*}{\textbf{Average}}
    
    \\



    \textbf{Methods}  
    & LHP~\cite{lhp}
    & LOLv2~\cite{lolv2}
    & RealSnow~\cite{WGWSNet}
    & 
    
     \\



    
\midrule[0.05em]
          

    MPRNet~\cite{MPRNet}       & 29.16  & 9.67  & 25.82 & 27.10\\
    SwinIR~\cite{SwinIR}       & 21.74   & 18.67 & 18.45  & 20.91 \\
    NAFNet~\cite{NAFNet}       & 26.04   & 26.46  & \underline{27.51}  & 26.33 \\
    Restormer~\cite{Restormer} & 27.23  & 28.92  & 26.73  & 27.26  \\
          
    PromptIR~\cite{PromptIR}        & 25.68 & -  & -  & - \\
    MiOIR~\cite{mioir}              & 28.69 & 10.62  & - & -\\
    InstructIR~\cite{instructir}    & 28.93  & \underline{30.31}  & -  & - \\
    OneRestore~\cite{OneRestore}    & 24.55  & 17.76  & 20.12  & 23.24 \\
    
\midrule[0.05em]

        Ours      & \textbf{30.04}  & \textbf{30.40}  & \textbf{28.04}  & \textbf{29.70} \\
\bottomrule[0.15em]
\end{tabular}
}
\end{minipage}
\hfill
\begin{minipage}[t]{0.49\textwidth}
\centering
\caption{\small \small \textbf{Generalization performance on unseen corruptions}. All methods are directly evaluated without additional training on corresponding datasets. PSNR scores are reported.}
\label{table:generalization_image_restoration_results_unseen}
\vspace{-1mm}
\resizebox{1\textwidth}{!}{%
\setlength{\tabcolsep}{3pt}
\begin{tabular}{ l  c c c }
\toprule[0.15em]

    & \textbf{Raindrop}
    & \textbf{UDC}
    & \textbf{Underwater}
    \\
    
      \textbf{Methods}  
    & Raindrop~\cite{raindrop}
    & TOLED~\cite{toled}
    & UIEB~\cite{uieb}
    \\
    
\midrule[0.05em]
    NAFNet~\cite{NAFNet}       & 22.89 & 26.89 & 15.54 \\
    Restormer~\cite{Restormer}    & 23.34 & 28.44 & 15.80\\
    PromptIR~\cite{PromptIR}     & 22.98 & 25.02 & 16.12 \\
    MiOIR~\cite{mioir}        & 23.59 & 20.11 & 16.26 \\
    OneRestore~\cite{OneRestore}   & 22.02 & 15.16 & 15.87 \\
\midrule[0.05em]
    Ours         & \textbf{24.91} & \textbf{29.64} & \textbf{18.07} \\
\bottomrule[0.15em]
\end{tabular}
}
\end{minipage}
\vspace{-3.5mm}
\end{table*}

%
%
%

%
%
\textbf{Baseline and Metric}\,\,
We adopt task-agnostic methods~\cite{MPRNet,SwinIR,Restormer,NAFNet} and recent all-in-one methods~\cite{DL,Transweather,TAPE,AirNet,PromptIR,IDR,instructir,OneRestore, moce, dcpt, tur, DaAIR} as baselines.
In comparison, the performance of task-agnostic methods and ours is reported using models trained on seven tasks and the developed mixed degradation space in single-degradation and mixed-degradation setups, respectively. 
For other methods, we follow the evaluation setting of~\cite{RestoreAgent} and only evaluate on tasks they support.
We adopt PSNR and SSIM as metrics. In comparison of single-task models, ablation study, and \textit{Suppl}., we only report PSNR.
Following~\cite{PromptIR}, metrics are calculated in RGB-channels, except that metrics are calculated on Y channel of YCbCr color space in single deraining task comparison~\cite{Restormer}.

%
%
%
%

\begin{table*}
\begin{minipage}[t]{0.34\textwidth}
\caption{\small \small Effect of DR extractors.}  
\label{table:degradation_extractor}
\vspace{-1mm}
\centering
\resizebox{1\textwidth}{!}{%
\setlength{\tabcolsep}{3pt}
\begin{tabular}{l c c c c}
\toprule[0.15em]
         & Rain & Rain & Test  & Test \\
 Methods & 200H & 200L & 1200  & 2800 \\
 
\midrule[0.1em]
VGG~\cite{vgg}            & 31.47 & 40.59 & 35.26 & 34.18 \\
DDR~\cite{ddr}            & 30.97 & 40.37 & 35.25 & 34.17 \\
DA-CLIP~\cite{daclip}     & 31.65 & 40.67 & 35.24 & 34.12 \\
Manual               & 31.88 & 41.43 & 35.32 & 34.23 \\
Ours                      & \textbf{32.01} & \textbf{41.61} & \textbf{35.50} & \textbf{34.40} \\
\bottomrule[0.15em]
\end{tabular}
}
\vspace{1mm}
\end{minipage}
\hfill
\begin{minipage}[t]{0.3\textwidth}
\centering
\caption{\small \small Effect of fineness.} 
\label{table:finer_granularity}
\vspace{-1mm}
\resizebox{0.68\textwidth}{!}{%
\setlength{\tabcolsep}{3pt}
\begin{tabular}{ c c  c }
\toprule[0.15em]
        \multirow{2}{*}{\# DRs}       & \multicolumn{2}{c}{MiO (Average)} \\
                                     &  \textit{In-dist}.& \textit{Out-dist}.    \\
\midrule[0.1em]
                           1  & 22.06 & 17.23  \\
                           2  & 22.42 & 17.51  \\ 
                           4  & 23.75 & \textbf{18.49}  \\
                           8  & 24.22 & 18.35  \\
                           16 &  \textbf{24.30}   & 18.44  \\
\bottomrule[0.15em]
\end{tabular}
}
\end{minipage}
\hfill
\begin{minipage}[t]{0.33\textwidth}
\caption{\small \small Effect of \#granularities.} 
\label{table:granularity_estimation}
\centering
\vspace{-1mm}
\resizebox{1\textwidth}{!}{%
\setlength{\tabcolsep}{3pt}
\begin{tabular}{ c  c  c  c }
\toprule[0.15em]
      
      \multirow{2}{*}{\# Levels}
      &  \multirow{2}{*}{$\{$ \# DRs  $\}$} 
      &  \multicolumn{2}{c}{MiO (Average)}    
      \\

      &
      & \textit{In-dist}. 
      & \textit{Out-dist}.   
      \\
      
\midrule[0.1em]
 1 & $\{$ 8 $\}$          & 24.22 & 18.35  \\
 2 & $\{$ 1, 8 $\}$       & 24.27 & 18.86  \\
 2 & $\{$ 4, 8 $\}$       & 24.44 & 19.06  \\
 3 & $\{$ 1, 4, 8 $\}$    & \textbf{24.46} & 19.45  \\
 4 & $\{$ 1, 4, 8, 16$\}$ & 24.41 & \textbf{19.47}  \\
\bottomrule[0.15em]
\end{tabular}
}
\end{minipage}
\vspace{-7.5mm}
\end{table*}


\noindent\textbf{Training Details}\,\,
For the training of experts, we generally use the AdamW optimizer with batch-size of 8 and patch-size of 128, we take the CosineAnnealing learning rate scheduler with the initial learning rate of $3e^{-4}$. After the training, the weights of experts are frozen, and we train degradation and granularity estimation based routers. For the training of routers, we adopt Adam optimizer with batch-size of 8 and patch-size of 256, we adopt a fixed learning rate of $1e^{-3}$, and we set $\alpha$ to $0.1$ and $\beta$ to $0.01$ in Eq.~(\ref{equ:total_loss}).
All experiments are trained on 8 NVIDIA A6000 GPUs. 
More details of the experimental setups and training details can be seen in \textit{Suppl}.

\subsection{Comparison with State-of-The-Art Methods}
\label{sec:comparison_with_sota_method}

%
\noindent\textbf{Comparison with All-in-One Models}\,\,
As shown in Tab.~\ref{table:single_in_one_image_restoration_results} and Fig.~\ref{fig:visual_comparisons_all_in_one_sio}, our method can significantly outperform task-agnostic and all-in-one methods by a large margin in single-degradation.
In mixed-degradation, our method continuously achieves higher performance than task-agnostic and all-in-one methods from simple to complex mixed degradation (Tab.~\ref{table:multiple_in_one_image_restoration_results}).
In addition, comparisons on \textit{out-dist.}, \textit{real-world} (Tab.~\ref{table:generalization_image_restoration_results_real-world}), and \textit{unseen} (Tab.~\ref{table:generalization_image_restoration_results_unseen}) image restoration datasets reveal that our method also guarantees better generalization performance than others.
We found that it becomes challenging for all-in-one models to achieve significant performance gains over SOTA task-agnostic methods as the number of restoration tasks increases, suggesting that the use of tailored prompts or expert modules may have limited impact during training. This can be attributed to the used trivial degradation representations, which may fail to effectively guide methods to learn different restoration process.

\noindent\textbf{Comparison with Single-Task Models}\,\,
For a fair comparison, we evaluate the performance of our method in \textit{instruction} mode (Ours\dag).
The results in Tab.~\ref{table:specific_image_restoration_results} show that our method achieved competitive or better performance compared to single-task models across various image restoration tasks.
The main reason lies in the fact that when single-task models are required to handle a broader range of degradation patterns (\eg, all levels of snow), their performance on specific degradation (\eg, large snow) tends to be not guaranteed. Our method can learn image degradations in finer-grained levels, allowing each expert to focus on specific degradation (\eg, different levels of snow). As a result, our approach can achieve competitive or better performance compared to single-task models.
%

%
%

%

%

%
%
%
%
%

\subsection{Ablation Study}
\label{sec:discussion}
%

%
\noindent\textbf{Effect of DR Extractors}\,\,
In Tab.~\ref{table:degradation_extractor}, we compare the performance of experts trained on data clustered by different DR extractors on the deraining task, `Manual' refers to data split by the original dataset source. Since our degradation extractor is aware of finer-grained DRs, the clustered training data can be more consistent in DR, thus our method can achieve better performance.

\noindent\textbf{Effect of DR Fineness}\,\,
In Tab.~\ref{table:finer_granularity}, we investigate the impact of \#DRs by varying the granularity of clustering. As grouped into finer-grained DR groups, the performance on \textit{in-dist.} consistently improves. However, due to the increased difficulty of routing, the performance gain plateaus when \#DRs exceeds 8.
For generalization performance, clustering DR groups in proper coarse-grained level performs better than finer-grained levels, suggesting that it is essential to combine multiple granularities to keep high performance on both \textit{in-dist.} and \textit{out-dist}.

\noindent\textbf{Effect of Multiple Granularities}\,\,
%
As shown in Tab.~\ref{table:granularity_estimation}, incorporating both fine-grained levels and coarse-grained levels can maintain much better restoration performance and generalization ability than employing single granularity. Taking performance, training cost, and learning difficulty into account, we finally adopt $\{$1, 4, 8$\}$ in our experiments.

\noindent\textbf{Others}\,\,
More discussions are in \textit{Suppl}. Analysis of routing method is in Sec.~\textcolor{red}{D}, including expert usage statistics (Tab.~\textcolor{red}{C}), loss functions (Tab.~\textcolor{red}{D}) and routers (Tab.~\textcolor{red}{E}). Overhead comparison is in Tab.~\textcolor{red}{F} and Tab.~\textcolor{red}{A}. More visual results are in Sec.~\textcolor{red}{H}.

\vspace{-1mm}
\section{Conclusion}
In this paper, we introduced our UniRestorer, a novel universal image restoration framework designed to leverage degradation priors for improved restoration performance while alleviating the inevitable error in degradation estimation.
To this end, we learn all-in-one image restoration in multiple granularities and propose to estimate image degradation at proper granularity.
Specifically, we perform hierarchical clustering on degradation space and develop a multi-granularity degradation set as well as a MoE restoration model.
Besides vanilla degradation estimation, granularity estimation is further introduced to indicate the degree of degradation estimation error.
By jointly utilizing degradation and granularity estimation, we train routers to adaptively allocate unknown corrupted inputs to the most appropriate expert.
Experimental results demonstrate our superior performance and promising potential in closing the performance gap to specific single-task models.
%



\renewcommand{\thesection}{\Alph{section}}
\renewcommand{\thetable}{\Alph{table}}
\renewcommand{\thefigure}{\Alph{figure}}
\renewcommand{\theequation}{\Alph{equation}}

\setcounter{section}{0}
\setcounter{figure}{0}
\setcounter{table}{0}
\setcounter{equation}{0}

The content of the supplementary material involves: 
\begin{itemize}
\vspace{2mm}
\item Datasets, degradation pipeline, and degradation parameters are in Sec.~\ref{sec:suppl_dataset}. 
\vspace{2mm}
\item Implementation and training details are in Sec.~\ref{sec:suppl_training_details}. 
\vspace{2mm}
\item Discussion of the effect of expert design is in Sec.~\ref{sec:expert_choice}. 
\vspace{2mm}
\item Discussion of the effect of the proposed routing method is in Sec.~\ref{sec:suppl_routing_analysis}. 
\vspace{2mm}
\item Overhead comparison is in Sec.~\ref{sec:suppl_overhead}. 
\vspace{2mm}
\item Limitation and future works are in Sec.~\ref{sec:limitations}. 
\vspace{2mm}
\item Broader impacts are in Sec.~\ref{sec:impacts}. 
\vspace{2mm}
\item More visual comparisons are in Sec.~\ref{sec:suppl_more_visual_comparisons}.
\end{itemize}

\section{Data}
\label{sec:suppl_dataset}
\subsection{Dataset}
%
We detail the usage of datasets in training our models on single-degradation and mixed-degradation:
(1) Single-degradation:
For training, we adopt Rain100L~\cite{Rain200H} dataset for deraining, RESIDE~\cite{reside} dataset for dehazing, DFBW~(DIV2K~\cite{div2k}, Flickr2K~\cite{flickr2k}, BSD~\cite{bsd68}, and WED~\cite{wed}) datasets for Gaussian color image denoising and compression artifacts removal, GoPro~\cite{GoPro} dataset for motion deblurring, LOLv1~\cite{LOL} dataset for low-light enhancement, Snow100K~\cite{Snow100K} dataset for desnowing.
Following~\cite{PromptIR, IDR}, we evaluate our method on Rain100L~\cite{Rain200H} for deraining, SOTS~\cite{reside} for dehazing, BSD68~\cite{bsd68} for Gaussian color image denoising, GoPro~\cite{GoPro} for motion deblurring, LOLv1~\cite{LOL} for low-light enhancement, Snow100K-L~\cite{Snow100K} for desnowing, and LIVE1~\cite{live1} for compression artifacts removal.
(2) Mixed-degradation:
In mixed-degradation, we synthesize mixed degradation datasets from clean images in DF2K (DIV2K~\cite{div2k} and Flickr2K~\cite{flickr2k}) datasets for training and evaluation, respectively. The degradation pipeline is illustrated in Fig.~\ref{fig:suppl_degradtion_pipeline}.
(3) Single-task comparison:
Due to the scale limitation of Rain100L, we collect Rain200H~\cite{Rain200H}, Rain200L~\cite{Rain200H}, DID~\cite{test1200}, and DDN~\cite{test2800} as training data to enlarge the degradation space and the number of training data for deraining task. Based on the clustered results, we retrain experts for degradation set related to rain degradation and replace original deraining experts used in single-degradation with newly trained ones. For other tasks, we use the same datasets as those used in the single-degradation.
(4) Ablation study and experiments in supplementary material:
Since the effectiveness of our method can be revealed in both performance on in-distribution and out-of-distribution samples, we conduct ablation studies on mixed-degradation, in which we can adjust degradation parameters to control data distribution. To simplify the evaluation process, we adopt Urban100~\cite{urban100} as high-quality clean data and the same degradation pipeline as mixed-degradation.

%
%
%
\subsection{Degradation pipeline and Parameters}
\begin{figure*}[t]
	\centering
	\setlength{\abovecaptionskip}{0.1cm}
	\setlength{\belowcaptionskip}{-0.3cm}
	\includegraphics[width=0.95\textwidth]{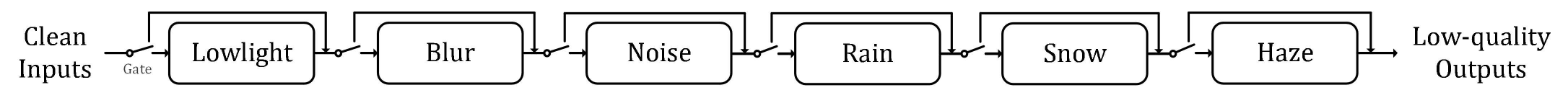}
	\caption{\small \small Degradation Pipeline.}
	\label{fig:suppl_degradtion_pipeline}
   \vspace{1mm}
\end{figure*}

%
We follow the degradation pipeline of~\cite{Zhang_2022_CVPR} to construct synthetic low-quality datasets for the training of the degradation extractor and all-in-one image restoration (mixed-degradation setup). As shown in Fig.~\ref{fig:suppl_degradtion_pipeline}, given high-quality clean inputs $y$, we orderly add synthetic lowlight, blur, noise, rain, snow, and haze to synthesize corresponding low-quality outputs $x$. We adopt gates between different degradations to gap one of them probably, the probability of all gates is generally set to $0.5$. 
The way of each degradation is introduced as follows:

\noindent\textbf{Blur}\,\,
Following~\cite{SRMD}, we synthesize blurring degradation as,
\begin{equation}
    x = y \otimes k,
\end{equation}
where $k$ is the blur kernel. In our implementation, the type of blur kernel includes isotropic and anisotropic, each of which is employed with the probability of $0.5$. The kernel size is sampled in the range from $7$ to $23$. The standard deviation of the Gaussian kernel is set from $1$ to $3$ for \textit{in-dist.} and $3$ to $4$ for \textit{out-dist.}

\noindent\textbf{Noise}\,\,
Gaussian noise is synthesized as,
\begin{equation}
    x = y + n,
\end{equation}
where $n$ is Gaussian noise with variance $\sigma^2$. $\sigma$ is set from $15$ to $35$ for \textit{in-dist.} and $35$ to $50$ for \textit{out-dist.}

\noindent\textbf{Rain}\,\,
Raining degradation is synthesized as,
\begin{equation}
    x = y + r,
\end{equation} 
where $r$ is rain streaks. Following~\cite{mioir}, $r$ is synthesized with random noise and Gaussian blur, where noise is related to the strength of the rain, and blur kernel is related to the rain direction, length, and width. 
We adopt the number of rain streaks ranges from $50$ to $100$ in \textit{in-dist.} and $100$ to $150$ for \textit{out-dist.} Rain direction is generally adopted from $-45^{\circ}$ to $45^{\circ}$, rain length is set from $20$ to $40$, and rain width is uniformly sampled from $\{3, 5, 7, 9, 11\}$.



\noindent\textbf{Haze}\,\,
Following~\cite{reside}, hazing degradation is synthesized as,
\begin{equation}
\begin{aligned}
    &x=T(y) y + a(1 - T(y)), \\
    &T(y)=e^{-\beta D(y)},
\end{aligned}
\end{equation}
where $a$ is the global atmospheric light, $T(y)$ is the transition matrix, $D(y)$ is the depth map, and $\beta$ is the scattering coefficient of the atmosphere. In haze degradation synthesizing, the value of $\beta$ largely decides the thickness of haze. We adopt one publicly released project $\operatorname{MegaDepth}$\footnote{\url{https://github.com/zhengqili/MegaDepth}} to estimate the depth image. In our experiments, the global atmospheric light is set from $0.8$ to $1$, the $\beta$ is set from $0.5$ to $1.5$ for \textit{in-dist.} and $1.5$ to $2$ for \textit{out-dist.}

\noindent\textbf{Snow}\,\,
Snowing degradation is synthesized as,
\begin{equation}
    x = y \times (1 - m) + m,
\end{equation}
where $m$ is the snow mask randomly sampled from Snow100K~\cite{Snow100K} dataset.

\noindent\textbf{Lowlight}\,\,
Following~\cite{OneRestore}, lowlight degradation is synthesized as,
\begin{equation}
    x = \frac{y}{I(y)} I(y)^\gamma
\end{equation}
where $I(y)$ is the illusion map obtained from RetinexFormer~\cite{retinexformer} and $\gamma$ is the darkening coefficient. Since we explicitly include noise degradation in our degradation pipeline, we do not additionally consider noise in this step. We set $\gamma$ from $1$ to $2$ for \textit{in-dist.} and $2$ to $2.5$ for \textit{out-dist.}

Besides, \textbf{JPEG compression} is synthesized by employing the publicly released project $\operatorname{DiffJPEG}$\footnote{\url{https://github.com/mlomnitz/DiffJPEG}}.

\section{Implementation and Training Details}
\label{sec:suppl_training_details}
\begin{figure*}[h]
	\centering
	\setlength{\abovecaptionskip}{0.1cm}
	\setlength{\belowcaptionskip}{-0.3cm}
	\includegraphics[width=\columnwidth]{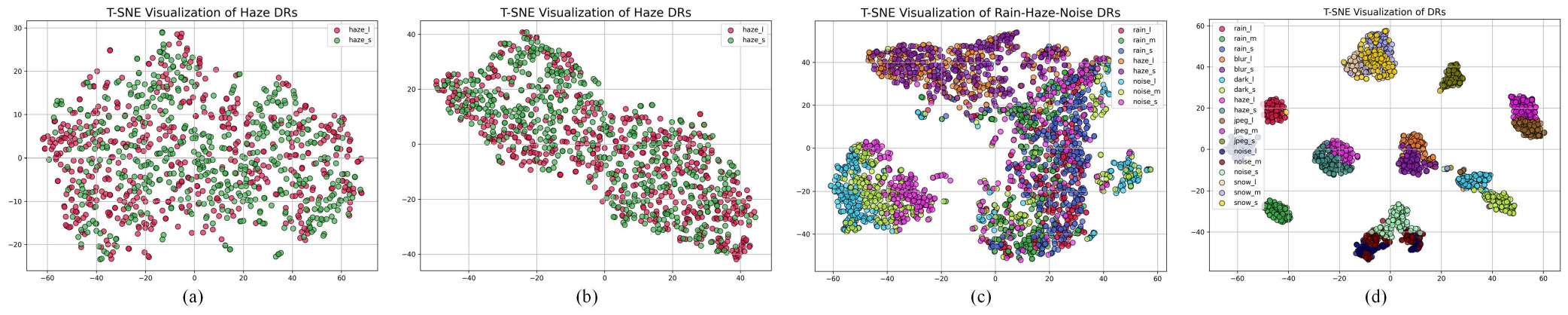}
	\caption{
    T-SNE visualization of deep features (Restormer~\cite{Restormer}, PromptIR~\cite{PromptIR}) and degradation representations (Ours).
    (a) and (b): Restormer and PromptIR are not aware of fine-grained degradation representations.
    (c) While PromptIR demonstrates a certain capability to distinguish between different types of degradations, it still exhibits misclassification across various cases.
    (d) Our method distinguish degradation representations well in both coarse-grained and fine-grained levels.}
\label{fig:tsne}
\end{figure*}

\textbf{Degradation Extractor}\,\,
Our degradation extractor is based on previous work, DA-CLIP~\cite{daclip}. 
DA-CLIP uses a lightweight image controller, a copy of the CLIP image encoder augmented with zero-initialized residual connections, to guide the image encoder output toward producing both content-related features and degradation-related embeddings. Using contrastive learning, the content embedding is aligned with clean text descriptions, and degradation embedding indicates the type of degradation, aligned with degradation-related text prompts.
For more details, please refer to the original paper.
To demonstrate the effectiveness of our degradation extractor, we visualize the T-SNE results in Fig.~\ref{fig:tsne} (a).
For example, in the single dehazing task, both Restormer and PromptIR fail to capture fine-grained DRs, as shown in Fig.~\ref{fig:tsne} (a) and Fig.~\ref{fig:tsne} (b). In contrast, our method learns DRs at finer levels of granularity and allows restoration on more specific degradations, thus achieving competitive or even better performance to single-task models.
While PromptIR can separate different types of degradations, it still exhibits misidentification across various cases.
Such cases may render the dedicated prompts less effective, which are originally as guidance for specific tasks. 
As a result, PromptIR may offer limited performance improvements over task-agnostic methods in some of the restoration tasks.

\noindent\textbf{Multi-Granularity MoE Restoration Model}\,\,
After training the degradation extractor, we construct the DR set for all training samples.
Then, we hierarchically cluster the obtained DR set into multiple groups in different granularities, constructing multi-granularity degradation sets.
Considering the task-agnostic capability, we generally adopt Restormer~\cite{Restormer} as the expert model across most tasks.
However, in single-degradation, we found Restormer cannot perform well in low-light enhancement.
We thus observe clusters predominantly composed of low-light degradations and select RetinexFormer~\cite{retinexformer} as the expert model for those clusters.
As a result, our MoE restoration model has three granularities. In the coarsest-grained level, we use single expert model to solve all observed degradations; in the second level, we use $7$ experts to solve degradation sets with coarser DRs; and in the finest-grained level, we have $19$ experts, and each expert is responsible for degradation sets with more fine-grained DRs.
In mixed-degradation, we have $1$, $4$, and $8$ experts for the first, the second, and the third granularity level, respectively.

\noindent\textbf{Structure of $\mathcal{H}$ and $\mathcal{G}$}\,\,
Without dedicated design, we adopt 5-layer MLP networks for both $\mathcal{H}_{d}$ and $\mathcal{H}_g$, in which the number of input and output channels is $512$ (the same as the dimension of DR $e$). The structures of $\mathcal{G}_d$ and $\mathcal{G}_g$ are one-layer MLP networks with input channels of $512$ and the number of output channels equal to the number of experts.

\noindent\textbf{Instruction Mode, Ours\dag}\,\,
The \textit{instruction} mode of our method allows inference based on given task instructions, \ie, the task name. In our implementation, the task instructions can convert to specific binary masks, which represent pruning or not pruning on the corresponding expert model in our MoE restoration model. According to the masks, we first prune out the corresponding expert models, then we conduct degradation- and granularity-based routing to find the most appropriate expert model for corrupted inputs. In this way, the error in degradation estimation is alleviated, the routing accuracy is highly improved, and thus the restoration performance is further improved.

\noindent\textbf{Ablation Study on Degradation Extractor}\,\,
We adopt pre-trained vgg16, pre-trained ESRGAN on bicubic 4$\times$ super-resolution, and pre-trained DACLIP as DR extractors.
We construct multi-granularity degradation sets by leveraging DRs extracted from different DR extractors. 
Then we re-train expert networks based on each obtained degradation set.
The experiments are conducted on deraining task. 
Since our degradation extractor is aware of finer-grained DRs, the clustered training data can be more consistent in DR, thus our method can achieve better performance.

\noindent\textbf{Training of Degradation Extractor}\,\,
To train our DR extractor, we adopt the ways in Sec.~\ref{sec:suppl_dataset} to synthesize various types of low-quality data. 
According to the parameter that decides the degradation degree, we separate synthetic data into multiple groups.
For deraining degradation, strength from $0$ to $50$ is small rain, $50$ to $100$ is medium rain, and $100$ to $150$ is large rain;
In denoising, $\sigma$ of Gaussian noise from $0$ to $15$ is small noise, $15$ to $35$ is medium noise, and $35$ to $50$ is large noise;
In dehazing, scattering coefficient $\beta$  larger than $0.1$ is thick haze, $\beta$ smaller than $0.1$ is normal haze;
In deblurring, kernel size of motion blur kernel larger than $15$ is large blur, kernel size smaller than $15$ is small blur;
In desnowing, we use snow masks from Snow100K dataset and adopt the official separation.
In compression artifacts removal, image quality from $0$ to $15$ is large compression, $15$ to $35$ is medium compression, $35$ to $50$ is small compression;
In lowlight enhancement, the darkening coefficient from $1$ to $1.5$ is light corruption and $1.5$ to $3$ is heavy corruption.

\noindent\textbf{Training of Multi-Granularity MoE Restoration Model}\,\,
In the training of experts, we first conduct hierarchical clustering based on collected datasets, after that, all degraded training images are divided into multiple groups in different granularities. In each granularity, we record for each group the data samples that is allocated to it. Then, we use the recorded data to train expert restoration network for each group.
After training of experts, the weights of experts are frozen, and we train routers based on degradation- and granularity- estimation.
Most works in high-level vision and NLP tasks train experts and routers jointly. To well deal with the conflicts between restoring different degradations, MoE system in image restoration works or all-in-one image restoration works prefer separate training~\cite{WGWSNet, ADMS, grids, InstructIPT, RestoreAgent, lorair}, \ie, sequentially training experts and routers.
Similarly to the mentioned works, our experts are explicitly determined to solve degradations in each degradation set, experts should be learned in advance and are not affected by the routers to guarantee the stability of the whole model. 
Thus, we learn experts and routers sequentially rather than simultaneously.

\section{Effect of Expert Design}
\label{sec:expert_choice}
\noindent\textbf{Motivation of Expert Design}\,\,
MoE is an effective solution for training large models in high-level vision and NLP tasks. One common approach is to deploy sub-networks as experts based on one shared backbone weight. For example, in all-in-one image restoration, some works~\cite{WGWSNet, ADMS} adopt this strategy to develop their MoE system for alleviating conflict among different restoration tasks.
Another way is to use a full-network as an expert directly. For examples, early work~\cite{tgsr} in real-world super-resolution adopted this way, and recent all-in-one image restoration works~\cite{grids, RestoreAgent} also followed this paradigm.
In our work, using full-network and sub-network as expert have similar computational cost.
However, using full-network has several advantages. First, it is easier to build and expand. Second, it can assign its full parameters for specific degradation, which has a more powerful fitting ability. Third, we can benefit much from the previous advanced architecture design towards specific tasks, unleashing their full power in solving all-in-one image restoration topic.

\noindent\textbf{Lightweight Experts}\,\,
We have designed more lightweight experts (by scaling down Restormer~\cite{Restormer} and NAFNet~\cite{NAFNet}) and provided complexity analysis in Tab.~\ref{table:suppl_expert_choice_comparison}. It shows that it achieves obvious performance gains with a much smaller inference cost than PromptIR~\cite{PromptIR}.

\begin{table}[t]
\centering
\caption{All-in-One image restoration results (three tasks single-degradation). We replace original experts with lightweight NAFNet (Lite-NAFNet) and Restormer (Lite-Restormer) as experts. `Inf.' means during inference. The reported Params. mainly includes restoration networks.}
\label{table:suppl_expert_choice_comparison}
\vspace{1mm}
\resizebox{0.8\textwidth}{!}{%
\setlength{\tabcolsep}{3pt}
\begin{tabular}{ l c c c | c c c c c}
\toprule[0.15em]

    & \textbf{Derain}
    & \textbf{Dehaze}
    & \textbf{Denoise}
    & \#FLOPs
    & Latency
    & Params.(M)
    & Mem. 
    & Time(h)
    \\
    
    \textbf{Methods}  
    & Rain100L
    & SOTS
    & BSD68
    & (G)
    & (s)
    & Inf./Train
    & (M) Inf.
    & Train
    \\
    
\midrule[0.05em]
    PromptIR~\cite{PromptIR}      & 36.37 & 30.58 & 31.12 & 1266.2 & 0.355 &39.6/39.6  & 3552 & 201 \\
    \midrule[0.05em]
    Ours          & 41.68 & 36.44 & 31.43 & 1155.8 & 0.484 &26.1/339.3 & 4258 & 578 \\
    Ours (Lite-NAFNet)   & 38.98 & 31.69 & 31.28 & 154.7 & 0.156 &17.1/222.4 & 1818 & 169 \\
    Ours (Lite-Restormer)   & 41.41 & 35.38 & 31.35 & 672.7 & 0.395 &14.8/192.5 & 3526 & 442 \\
    
\bottomrule[0.15em]
\end{tabular}
}
\end{table}


\begin{table}[h]
\centering
\caption{All-in-One image restoration result (mixed-degradation). `Inf.' means during inference. `-Large' indicates the version of large scale. The reported Params. mainly includes restoration networks. PromptIR-L is trained in FP16.}
\label{table:suppl_vanilla_scaling_up_comparison}
\vspace{1mm}
\resizebox{0.7\textwidth}{!}{%
\setlength{\tabcolsep}{3pt}
\begin{tabular}{l c  | c c c c c}
\toprule[0.15em]
                  & MiO (Avg.)      & \#FLOPs & Latency &  Params.(M) & Mem.     & Time (h)\\
\textbf{Methods}  & \textit{In/out-dist}. & (G)   &  (s)    &  Inf./Train  & (M) Inf. & Train\\
\midrule[0.1em]
Restormer          & 22.06 /17.23   & 1128.9  & 0.368   & 26.1/26.1    & 3244  & 48   \\
Restormer-Large    & 23.98 /18.65   & 17360.1 & 1.375   & 441.8/441.8  & 13408 & 3792 \\
PromptIR-Large     & 24.13 /18.76   & 18139.6 & 1.283   & 439.3/439.3  & 12962 & -    \\
\midrule[0.05em]
Ours         & \textbf{24.46}/\textbf{19.45}  & 1155.8 & 0.484  & 26.1/339.3 & 4258 & 578 \\

\bottomrule[0.15em]
\end{tabular}
}
\end{table}

\noindent\textbf{Compare to Vanilla Scaling Up}\,\,
According to the degradation- and granularity- estimation, the routers select the most appropriate expert for each input low-quality image, leading to a sparse inference strategy (see Tab.~\ref{table:suppl_expert_choice_comparison} and Tab.~\ref{table:overhead_comparison}).
Though our method has comparable inference \#parameters, FLOPs, and latency to methods with similar backbones, one may know the effect of training parameters.
To mitigate the gap of training \#parameters, we scale up the training \#parameters of Restormer and PromptIR to match those of ours, and we conduct experiments on mixed-degradation.
As shown in Tab.~\ref{table:suppl_vanilla_scaling_up_comparison}, even after scaling up training parameters, Restormer-Large and PromptIR-Large cannot have marginal performance gains, it reveals that simply increasing the model size is not the essence of improving the performance, which is different from the scaling law in high-level vision and NLP tasks. 
From this perspective, our method provides an effective paradigm for scaling up low-level vision image restoration models, enabling both improved performance and efficient computation.

\begin{table*}
\begin{minipage}[t]{0.32\textwidth}
\centering
\caption{Expert usage statistics.}
\label{table:routing_analysis}
\vspace{1mm}
\resizebox{1\textwidth}{!}{%
\setlength{\tabcolsep}{3pt}
\begin{tabular}{ l c c}
\toprule[0.15em]
    
    & \textit{in-dist.}
    & \textit{out-dist.}
    \\
    Level
    & 0 / 1 / 2
    & 0 / 1 / 2
    \\
\midrule[0.05em]
    Expected  & 7 / 13 / 80 & 29 / 46 / 25 \\
    Practical & 5 / 17 / 78 & 34 / 50 / 16 \\
    
\bottomrule[0.15em]
\end{tabular}
}
\vspace{1mm}
\end{minipage}
\hfill
\begin{minipage}[t]{0.32\textwidth}
\centering
\caption{Effect of loss functions.}
\label{table:suppl_loss_functions}
\vspace{1mm}
\resizebox{1\textwidth}{!}{%
\setlength{\tabcolsep}{3pt}
\begin{tabular}{c c c c c }
\toprule[0.15em]
                 \multirow{2}{*}{$\ell_1$} & \multirow{2}{*}{$\mathcal{L}_{dg}$} & \multirow{2}{*}{$\mathcal{L}_{load}$} & \multicolumn{2}{c}{MiO (Average)}     \\
           &                      &                      & \textit{In-dist}. & \textit{Out-dist}. \\
\midrule[0.1em]
 \checkmark &             &            &  24.14         & 18.37          \\
 \checkmark & \checkmark  &            &  24.35         & 18.76          \\
 \checkmark & \checkmark  & \checkmark & \textbf{24.46} & \textbf{19.45}  \\
\bottomrule[0.15em]
\end{tabular}
}
\end{minipage}
\hfill
\begin{minipage}[t]{0.32\textwidth}
\centering
\caption{Effect of the routers}
\label{table:loss_functions}
\vspace{1mm}
\resizebox{1\textwidth}{!}{%
\setlength{\tabcolsep}{3pt}
\begin{tabular}{l c c c c }
\toprule[0.15em]

\multirow{2}{*}{Method}            & \multirow{2}{*}{$\mathcal{G}_{d}$}      & \multirow{2}{*}{$\mathcal{G}_{g}$}      & \multicolumn{2}{c}{MiO (Average)}         \\
                                   &                                     &                                         & \textit{In-dist}.    & \textit{Out-dist}. \\
\midrule[0.1em]
Clustering                                  &             &            &  17.01         & 14.19           \\
\midrule[0.05em]
Only $\mathcal{G}_d$                        & \checkmark  &            &  24.34         &  18.60          \\
Ours                                        & \checkmark  & \checkmark & \textbf{24.46} & \textbf{19.45}  \\
\bottomrule[0.15em]
\end{tabular}
}
\end{minipage}
\end{table*}


\section{Effect of the Proposed Routing Method}
\label{sec:suppl_routing_analysis}
\textbf{Routing Usage Statistics}\,\,
If degradation estimation (DE) on the finest granularity level is accurate, using corresponding fine-grained experts is optimal. However, fine-grained DE is error-prone. The proposed granularity estimation measures the uncertainty of DE. When the uncertainty is large, it means that the finest-grained expert is probably not suitable for this degradation. Instead, a coarser-grained expert is more robust to estimation error, thus being selected. 
In Tab.~\ref{table:routing_analysis}, we report the practical $\#$loads and expected $\#$loads (by maximizing PSNR) of different granularity levels during inference on Urban100 dataset with mixed-degradation, and our routing accuracies are respectively 92$\%$ and 82$\%$ on \textit{in-dist.} and \textit{out-dist.}

\noindent\textbf{Effect of $\mathcal{G}_d$ and $\mathcal{G}_g$}\,\,
As shown in  Tab.~\ref{table:loss_functions}, using both $\mathcal{G}_d$ and $\mathcal{G}_g$ leads to more accurate routing. 
Without using any routers, `Clustering' cannot distinguish DRs in different granularities and cannot guarantee the right routing results, leading to lower performance. 
Using only $\mathcal{G}_d$ improves the performance on \textit{in-dist.}, while without the awareness of granularity, it does not generalize well on \textit{out-dist.}

\noindent\textbf{Effect of Loss Functions}\,\,
As shown in Tab.~\ref{table:suppl_loss_functions}, both $\mathcal{L}_{dg}$ and $\mathcal{L}_{load}$ can benefit more accurate routing. 
In particular, $\mathcal{L}_{dg}$ and $\mathcal{L}_{load}$ can avoid the routing to collapse to the finest-grained level, improving performance on \textit{out-dist.}

\section{Overhead Comparisons}
\label{sec:suppl_overhead}
We calculate FLOPs and latency with $512\times 512\times 3$ input, on a single NVIDIA A6000 GPU. As shown in Tab.~\ref{table:overhead_comparison}, thanks to mixture-of-experts, the overhead of our method is slightly higher than the used image restoration backbone, Restormer~\cite{Restormer}. In addition, the overhead of our method can be largely reduced by replacing the image restoration backbone with more lightweight ones, \eg, lightweight NAFNet (Lite-NAFNet) and lightweight Restormer (Lite-Restormer).
%

\begin{table}[h]
\begin{center}
\caption{Overhead comparisons. }
\label{table:overhead_comparison}
\vspace{1mm}
\resizebox{0.4\textwidth}{!}{%
\setlength{\tabcolsep}{3pt}
\begin{tabular}{ c c c }
\toprule[0.15em]
 \textbf{Methods}    &  \#FLOPs (G) & Latency (s) \\
\midrule[0.1em]
Restormer~\cite{Restormer}       & 1128.9 & 0.368 \\ 
NAFNet~\cite{NAFNet}             & 505.5  & 0.186 \\
PromptIR~\cite{PromptIR}         & 1266.2  & 0.355 \\
OneRestore~\cite{OneRestore}    & 94.5 & 0.238 \\

Ours (Lite-NAFNet)                     & 154.7 & 0.156 \\
Ours (Lite-Restormer)                     & 672.7 & 0.395 \\ 
Ours                            & 1155.8 & 0.484 \\
\bottomrule[0.15em]
\end{tabular}
}
\end{center}
\end{table}

\section{Limitations and Future Works}
\label{sec:limitations}
On the one hand, the training data mainly comes from publicly released datasets, as required for fair comparison with baseline methods. However, the scale of training data is essential, particularly in developing large-scale models. In image restoration works, the scale of training data is often limited, \eg Rain100L~\cite{Rain200H} contains only 200 pairs of training data. Previous works~\cite{wang2021realesrgan} has exploited the benefits brought by scaling up the training data, we thus plan to collect more high-quality training data to bring further improvement on our method.
On the other hand, we plan to expand and align our degradation space with real-world scenarios, aiming to well restore real-world image degradations.

\section{Broader Impacts}
\label{sec:impacts}
In this paper, we propose UniRestorer, a novel all-in-one image restoration framework that can estimate and leverage robust degradation representation to adaptively handle diverse image corruptions. Our research aims to contribute to the advancement of the relevant community without any negative social impact.

\section{More Results and Visual Comparisons}
\label{sec:suppl_more_visual_comparisons}
\begin{figure*}[t]
	\centering
	\setlength{\abovecaptionskip}{0.1cm}
	\setlength{\belowcaptionskip}{-0.3cm}
	\includegraphics[width=\textwidth]{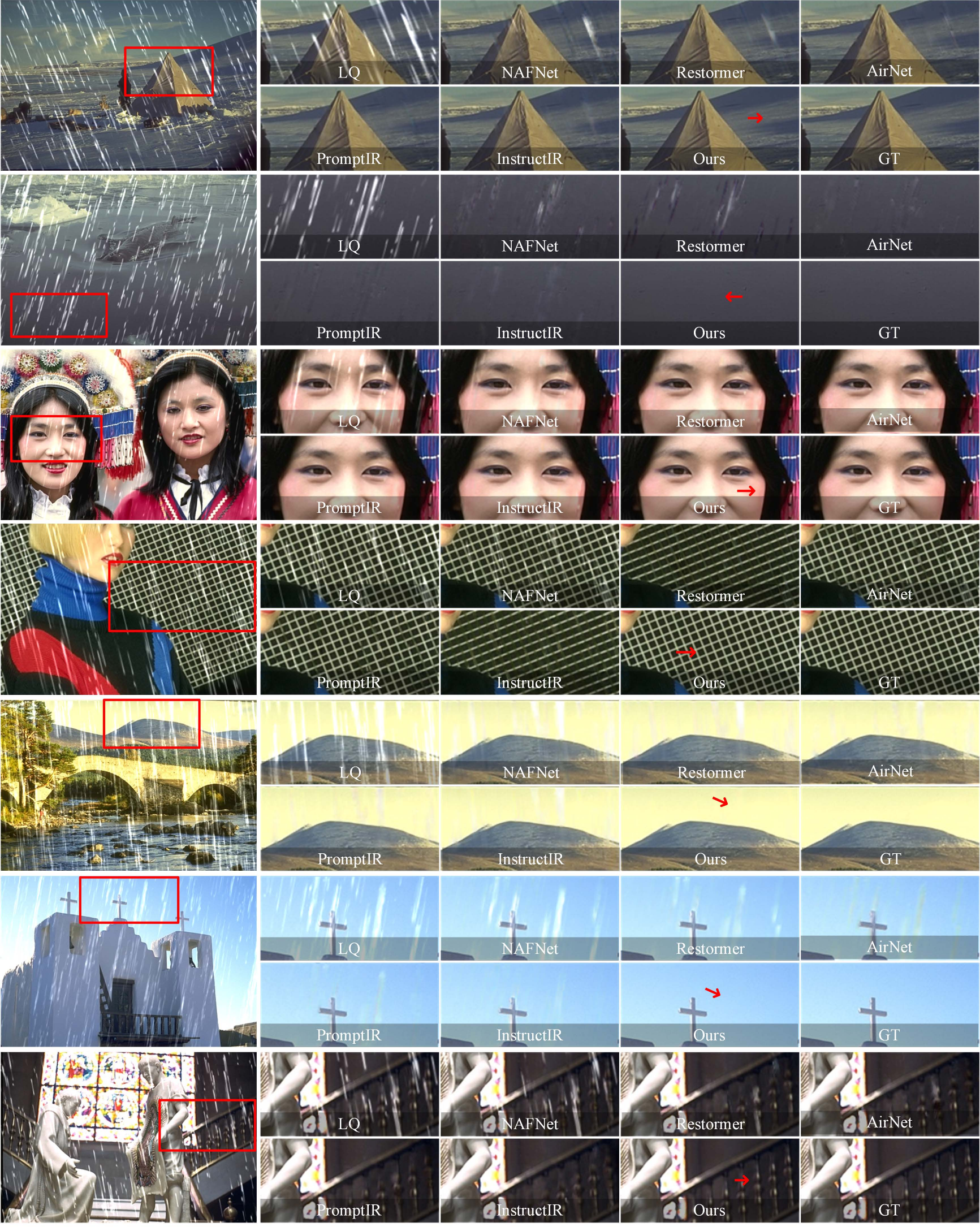}
	\caption{\small \small Visual comparison of results on All-in-One image restoration (deraining).}
	\label{fig:suppl_aio_derain}
   \vspace{1mm}
\end{figure*}

\begin{figure*}[t]
	\centering
	\setlength{\abovecaptionskip}{0.1cm}
	\setlength{\belowcaptionskip}{-0.3cm}
	\includegraphics[width=\textwidth]{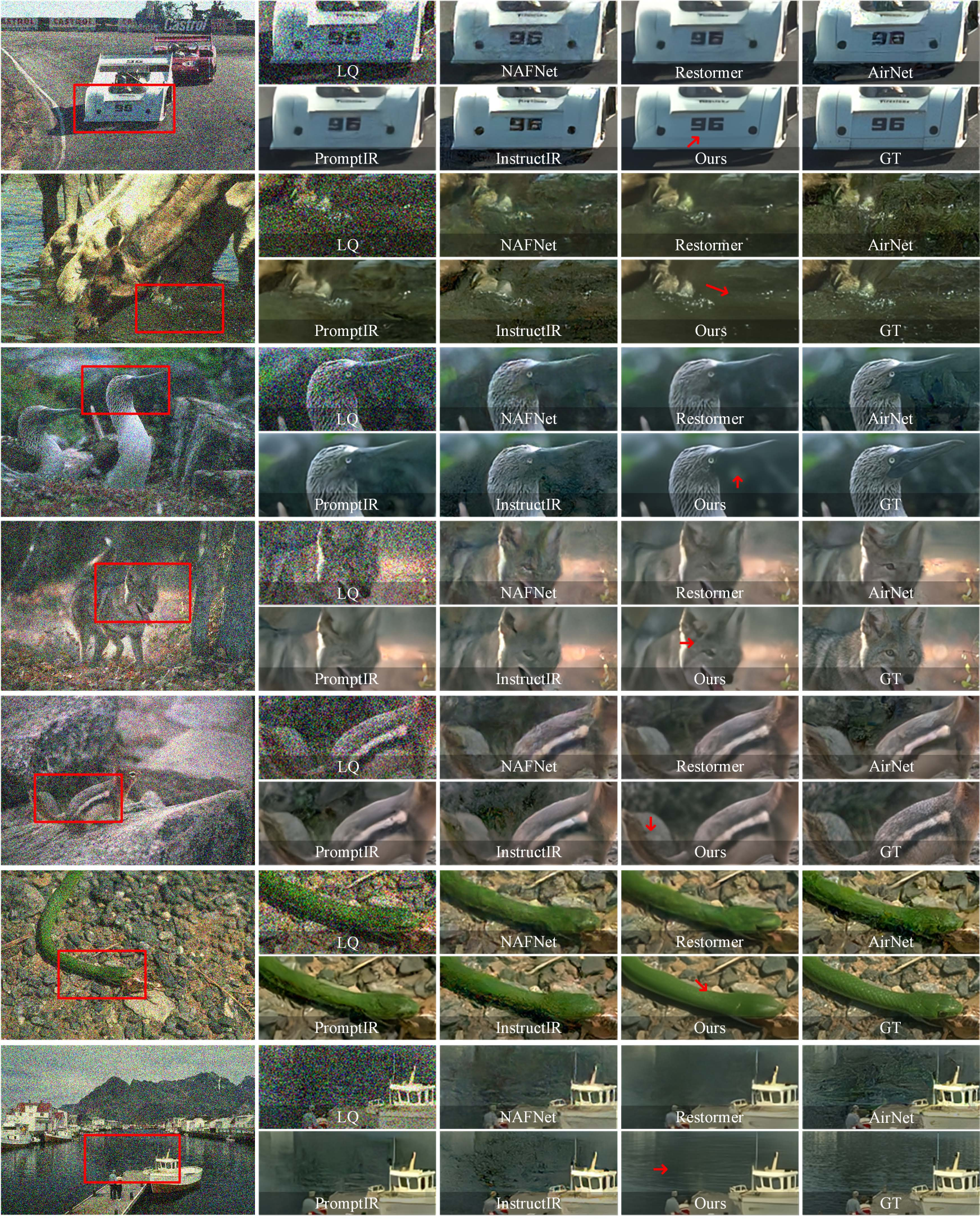}
	\caption{\small \small Visual comparison of results on All-in-One image restoration (Gaussian-denoising).}
	\label{fig:suppl_aio_denoise}
   \vspace{1mm}
\end{figure*}

\begin{figure*}[t]
	\centering
	\setlength{\abovecaptionskip}{0.1cm}
	\setlength{\belowcaptionskip}{-0.3cm}
	\includegraphics[width=\textwidth]{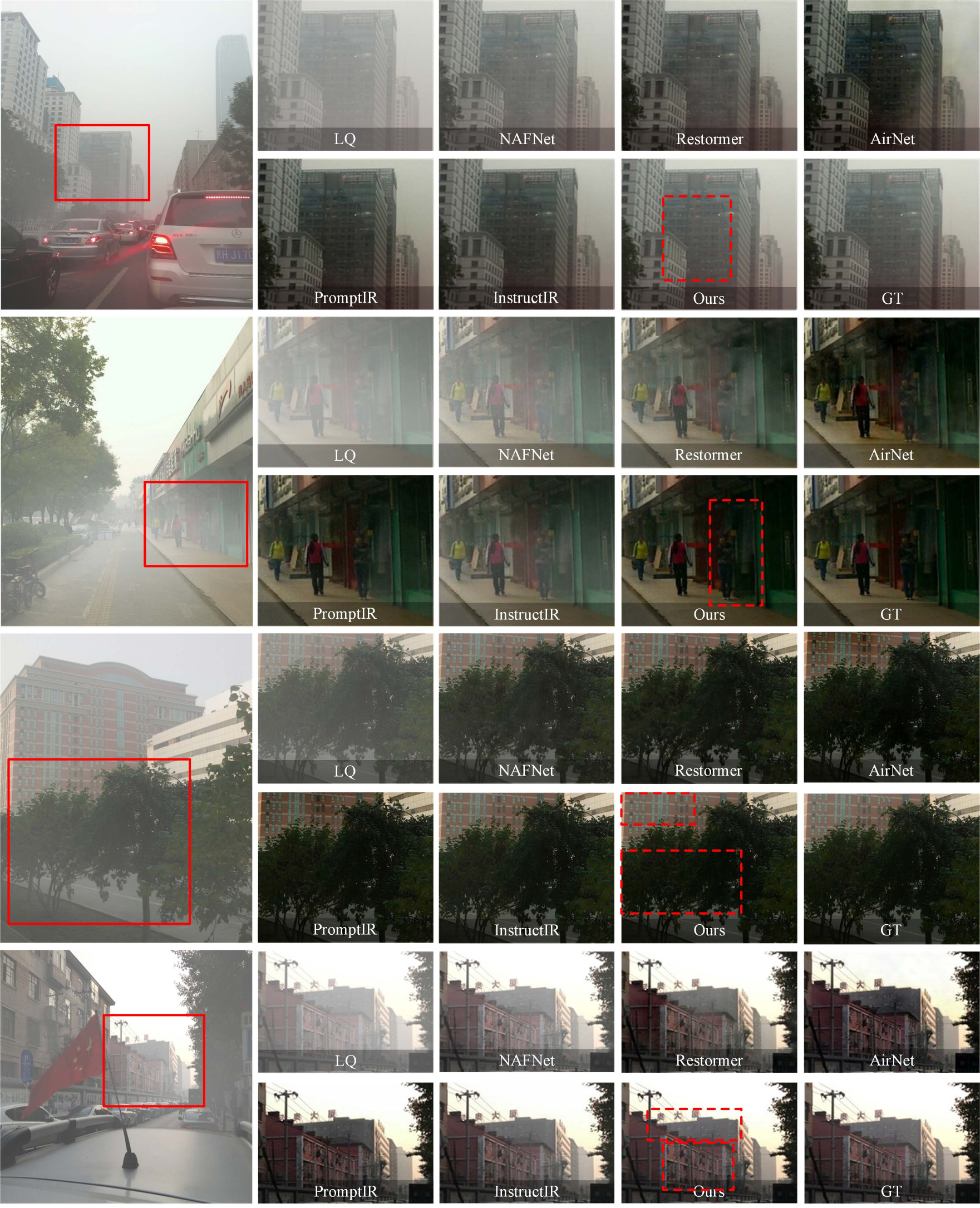}
	\caption{\small \small Visual comparison of results on All-in-One image restoration (dehazing).}
	\label{fig:suppl_aio_dehaze}
   \vspace{1mm}
\end{figure*}

\begin{figure*}[t]
	\centering
	\setlength{\abovecaptionskip}{0.1cm}
	\setlength{\belowcaptionskip}{-0.3cm}
	\includegraphics[width=\textwidth]{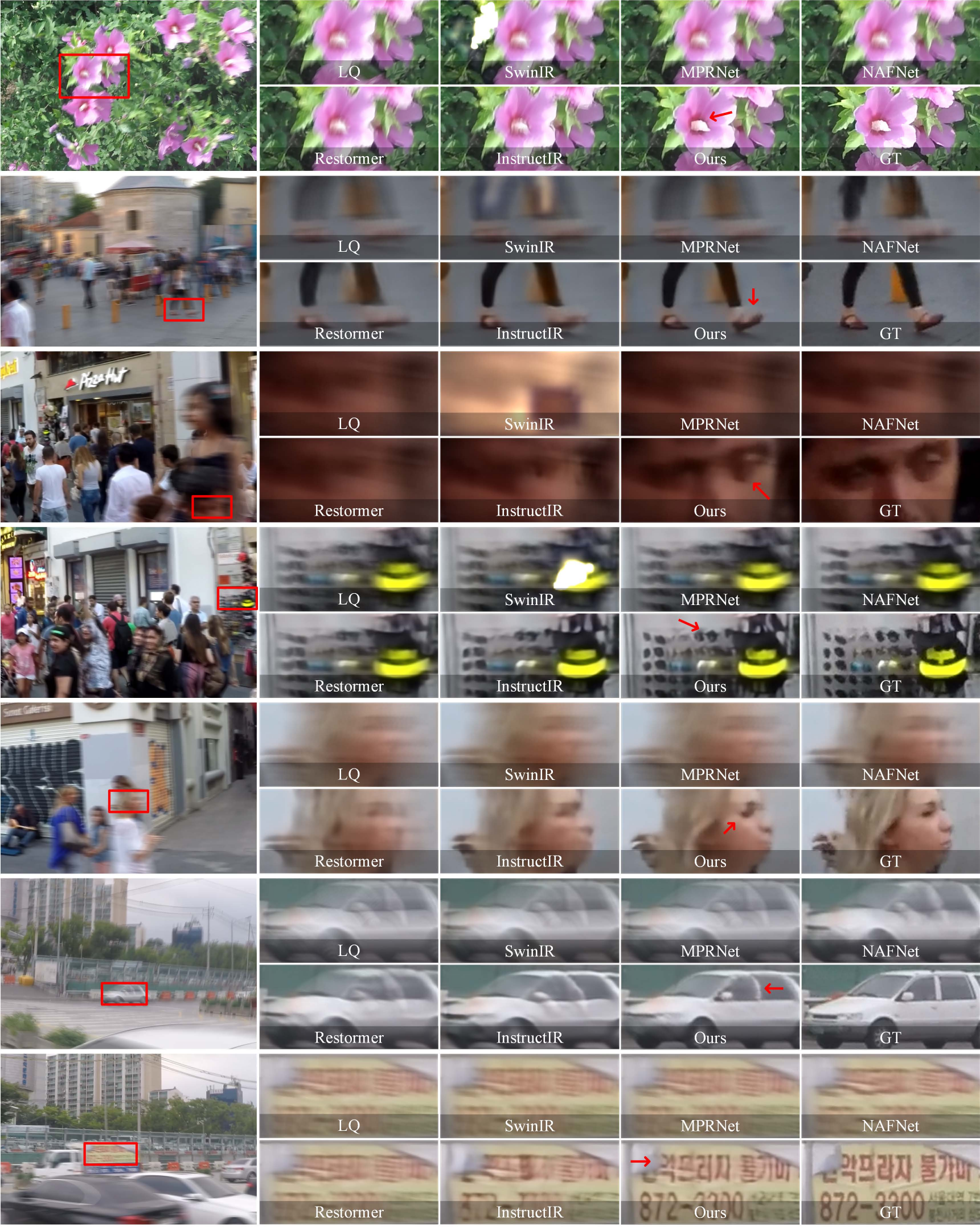}
	\caption{\small \small Visual comparison of results on All-in-One image restoration (motion-deblurring).}
	\label{fig:suppl_aio_deblur}
   \vspace{1mm}
\end{figure*}

\begin{figure*}[t]
	\centering
	\setlength{\abovecaptionskip}{0.1cm}
	\setlength{\belowcaptionskip}{-0.3cm}
	\includegraphics[width=\textwidth]{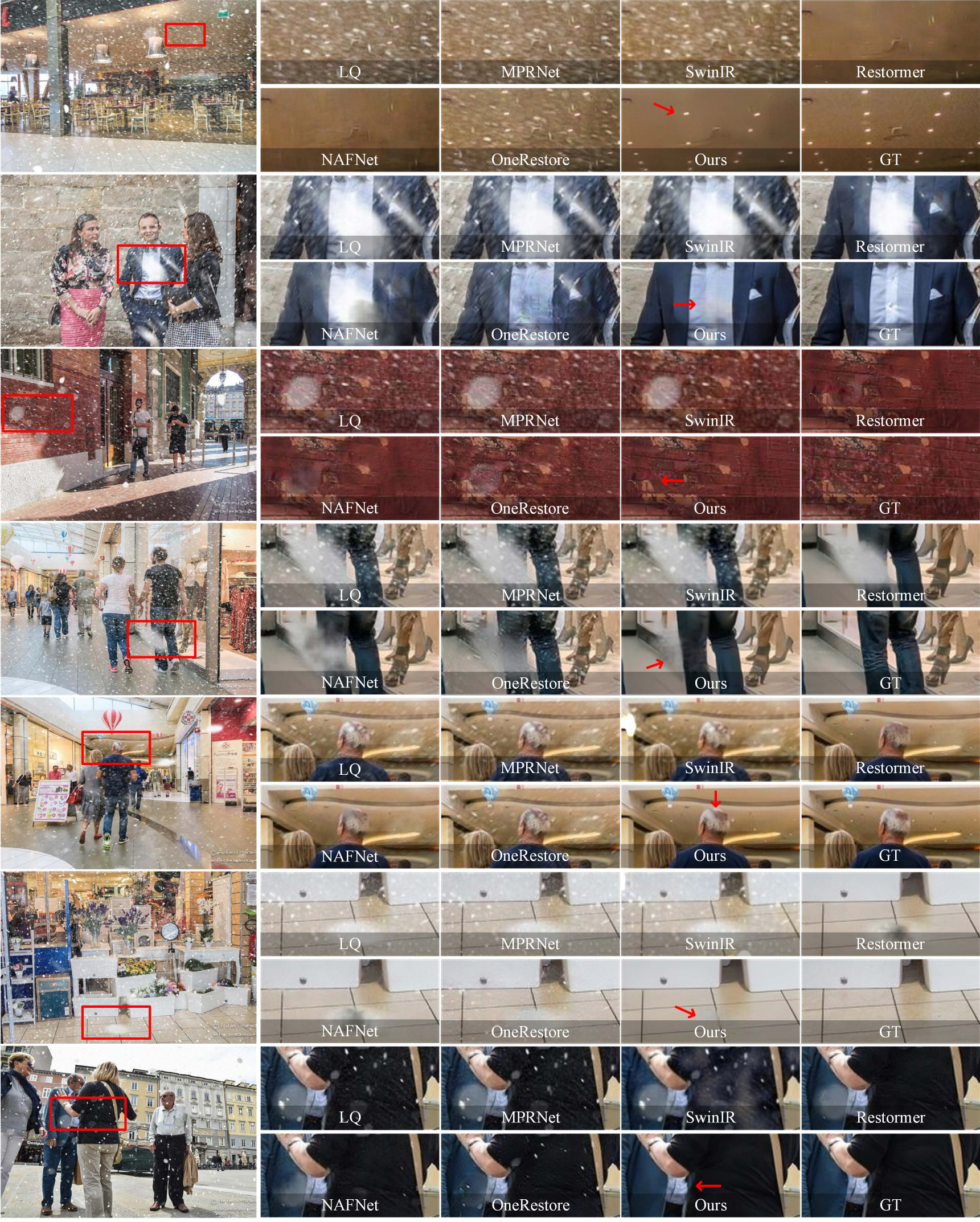}
	\caption{\small \small Visual comparison of results on All-in-One image restoration (desnowing).}
	\label{fig:suppl_aio_desnow}
   \vspace{1mm}
\end{figure*}

\begin{figure*}[t]
	\centering
	\setlength{\abovecaptionskip}{0.1cm}
	\setlength{\belowcaptionskip}{-0.3cm}
	\includegraphics[width=\textwidth]{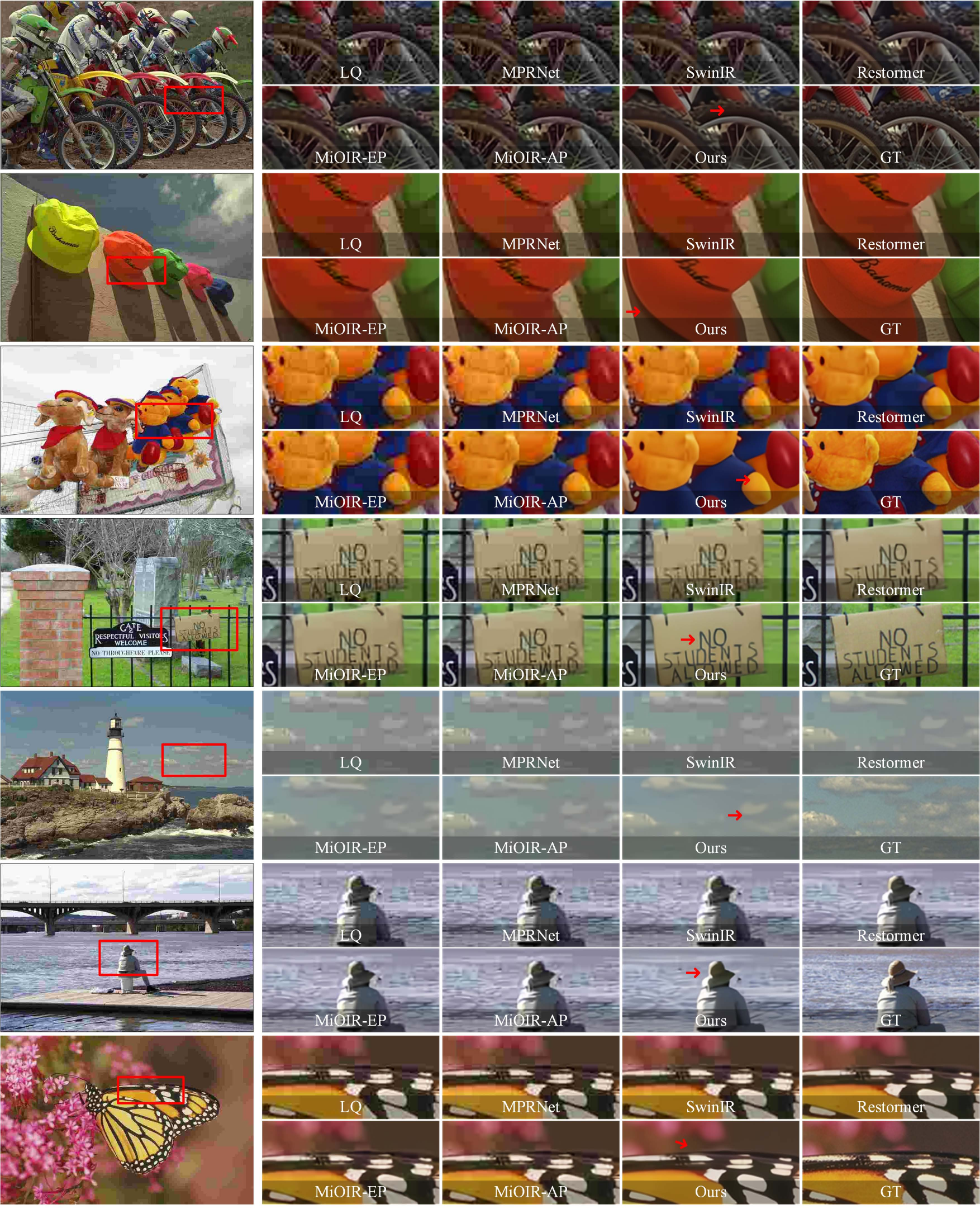}
	\caption{\small \small Visual comparison of results on All-in-One image restoration (compression artifacts removal).}
	\label{fig:suppl_aio_dejpeg}
   \vspace{1mm}
\end{figure*}

\begin{figure*}[t]
	\centering
	\setlength{\abovecaptionskip}{0.1cm}
	\setlength{\belowcaptionskip}{-0.3cm}
	\includegraphics[width=\textwidth]{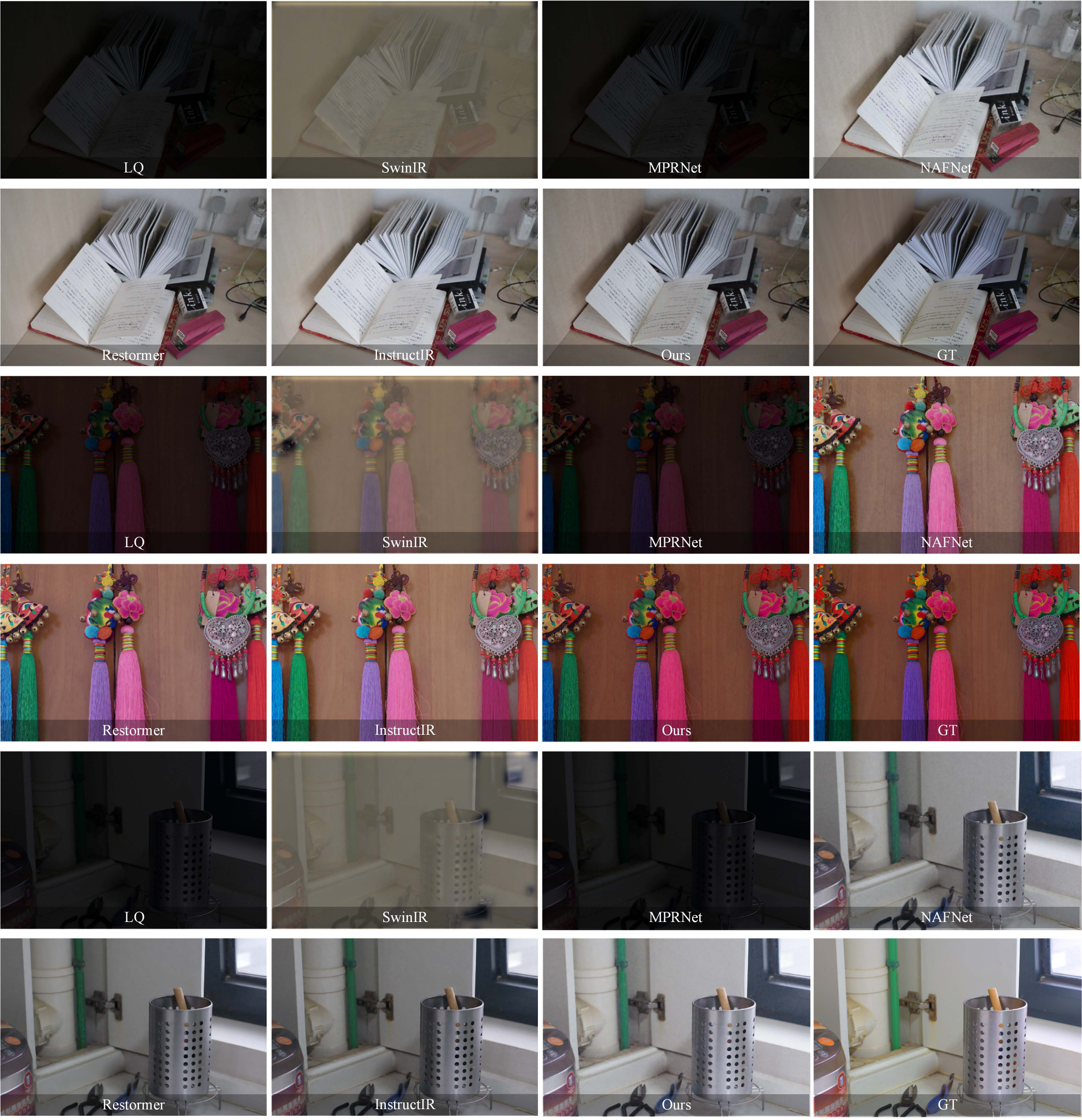}
	\caption{\small \small Visual comparison of results on All-in-One image restoration (lowlight enhancement).}
	\label{fig:suppl_aio_lowlight}
   \vspace{1mm}
\end{figure*}

\begin{figure*}[t]
	\centering
	\setlength{\abovecaptionskip}{0.1cm}
	\setlength{\belowcaptionskip}{-0.3cm}
	\includegraphics[width=\textwidth]{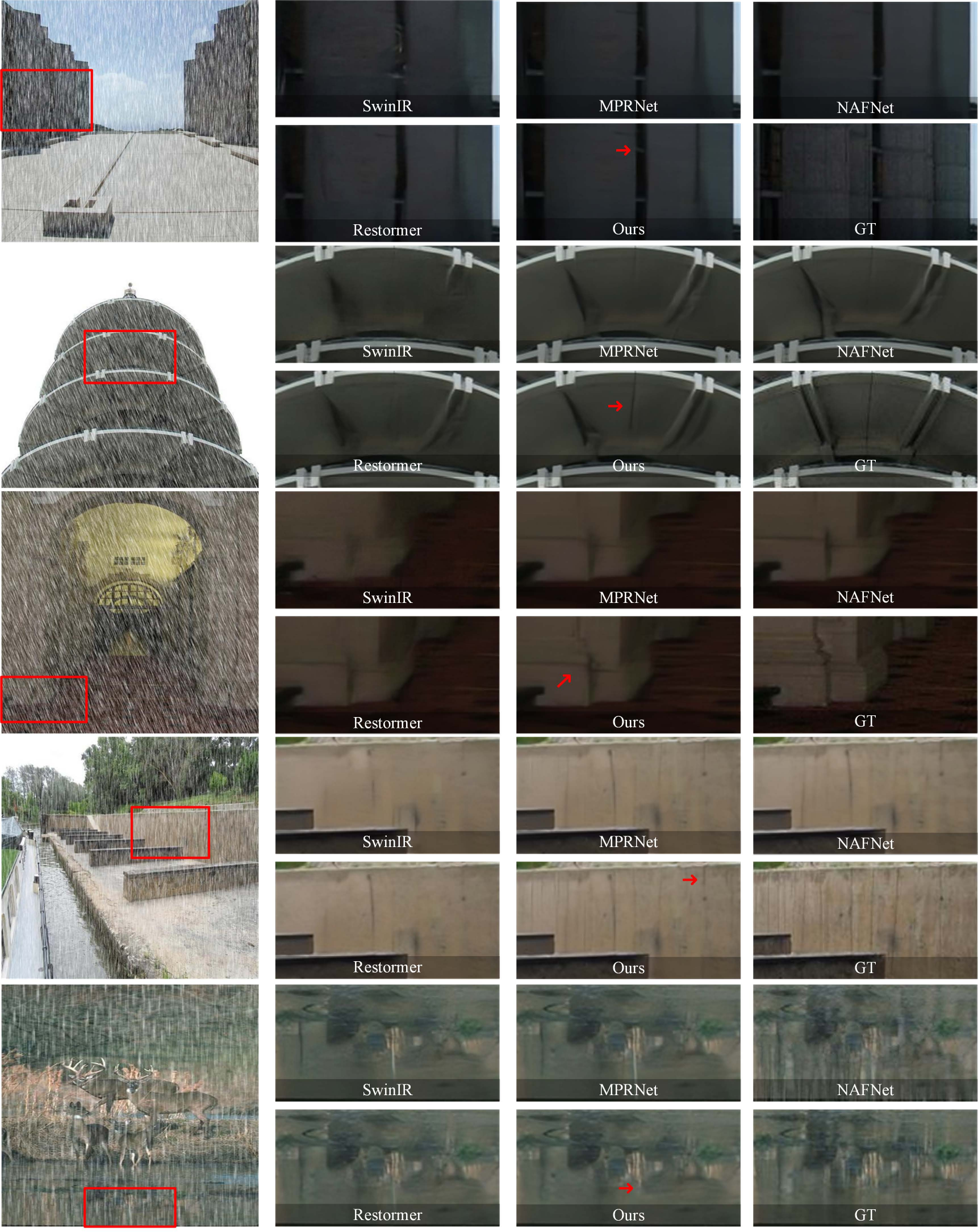}
	\caption{\small \small Visual comparison of results with specific models (deraining).}
	\label{fig:suppl_spe_derain}
   \vspace{1mm}
\end{figure*}

\begin{figure*}[t]
	\centering
	\setlength{\abovecaptionskip}{0.1cm}
	\setlength{\belowcaptionskip}{-0.3cm}
	\includegraphics[width=\textwidth]{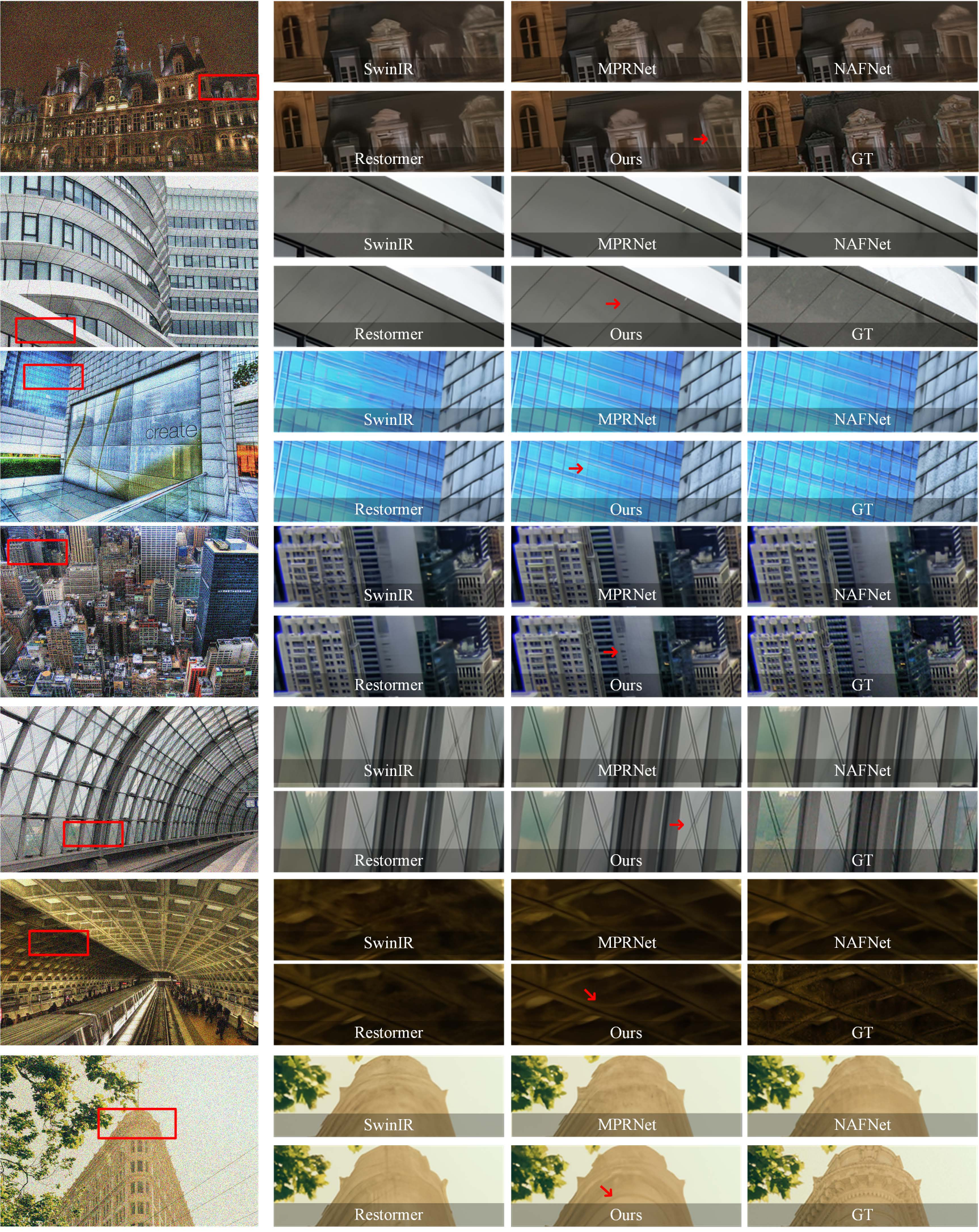}
	\caption{\small \small Visual comparison of results with specific models (Gaussian-denoising).}
	\label{fig:suppl_sep_denoise}
   \vspace{1mm}
\end{figure*}

\begin{figure*}[t]
	\centering
	\setlength{\abovecaptionskip}{0.1cm}
	\setlength{\belowcaptionskip}{-0.3cm}
	\includegraphics[width=\textwidth]{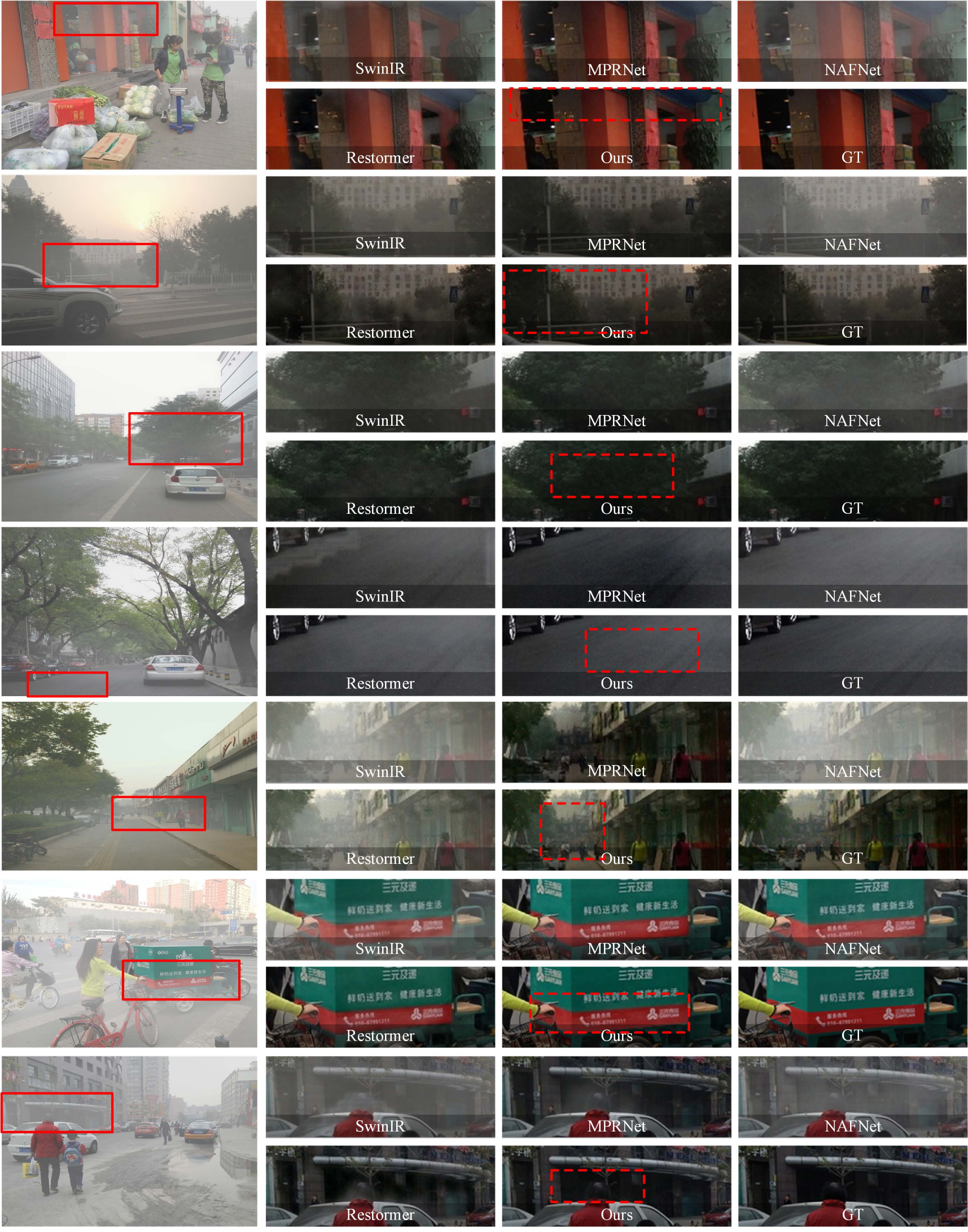}
	\caption{\small \small Visual comparison of results with specific models (dehaze).}
	\label{fig:suppl_spe_dehaze}
   \vspace{1mm}
\end{figure*}

\begin{figure*}[t]
	\centering
	\setlength{\abovecaptionskip}{0.1cm}
	\setlength{\belowcaptionskip}{-0.3cm}
	\includegraphics[width=\textwidth]{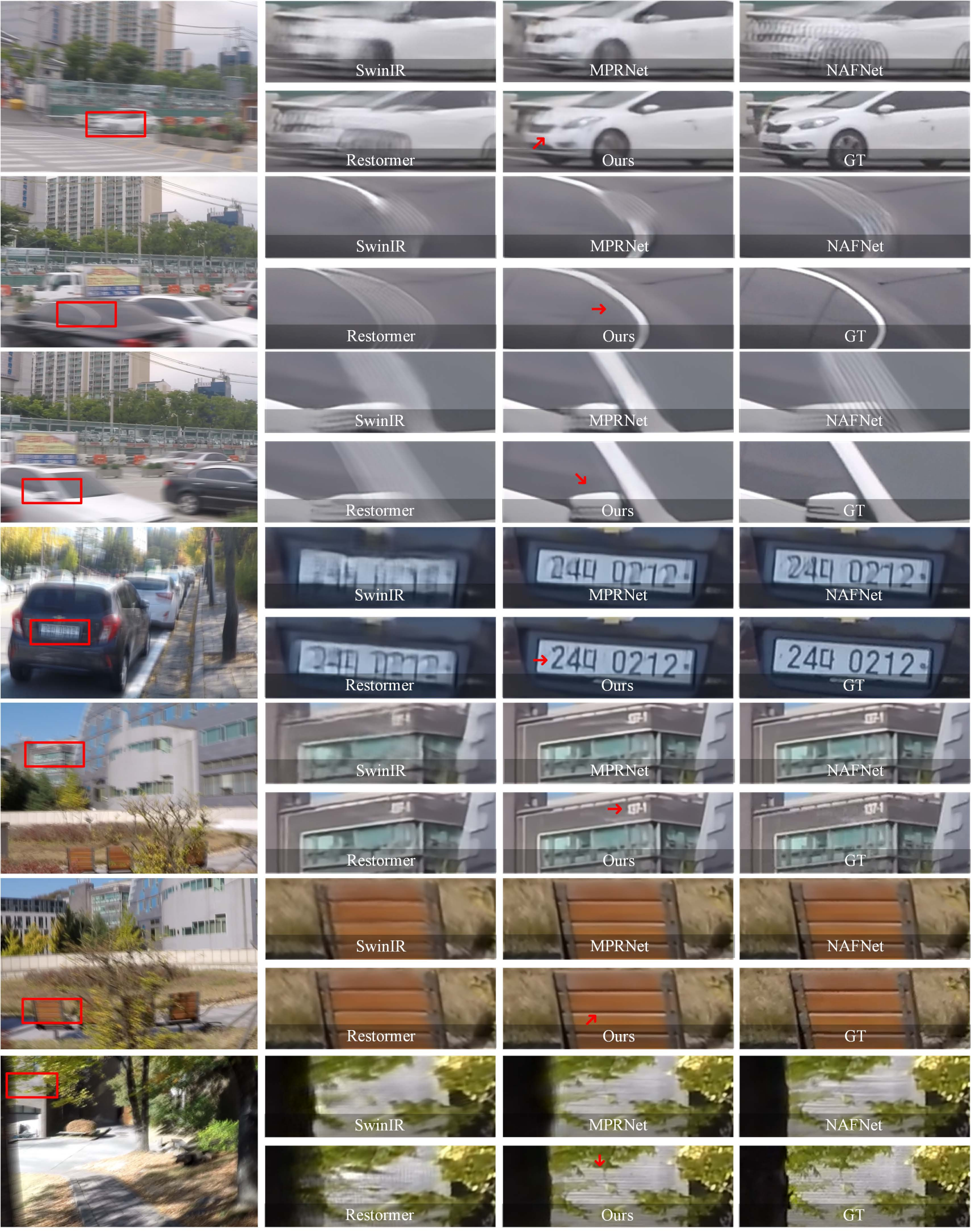}
	\caption{\small \small Visual comparison of results with specific models (motion-deblurring).}
	\label{fig:suppl_spe_deblur}
   \vspace{1mm}
\end{figure*}

\begin{figure*}[t]
	\centering
	\setlength{\abovecaptionskip}{0.1cm}
	\setlength{\belowcaptionskip}{-0.3cm}
	\includegraphics[width=\textwidth]{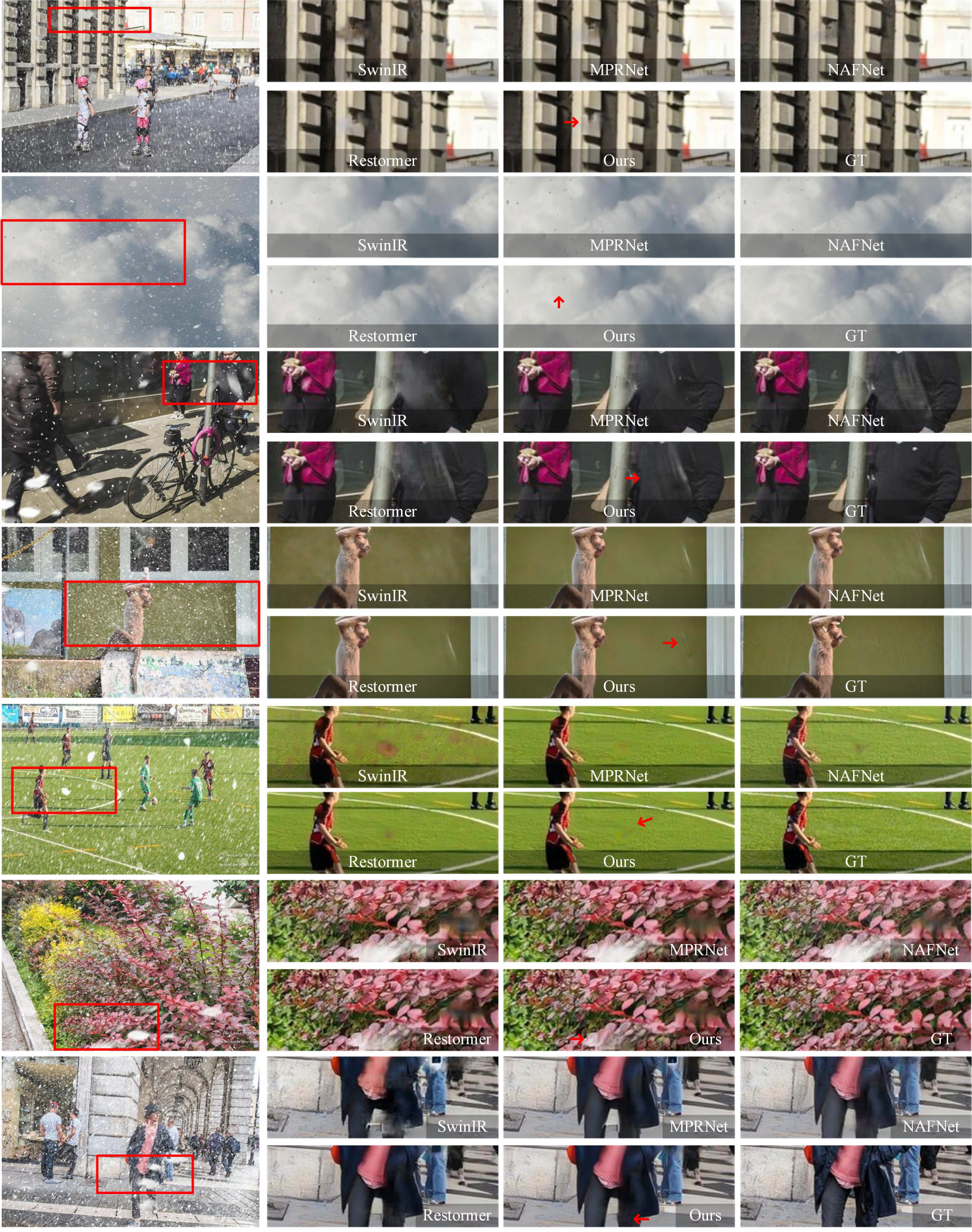}
	\caption{\small \small Visual comparison of results with specific models (desnowing).}
	\label{fig:suppl_spe_desnow}
   \vspace{1mm}
\end{figure*}

\begin{figure*}[t]
	\centering
	\setlength{\abovecaptionskip}{0.1cm}
	\setlength{\belowcaptionskip}{-0.3cm}
	\includegraphics[width=\textwidth]{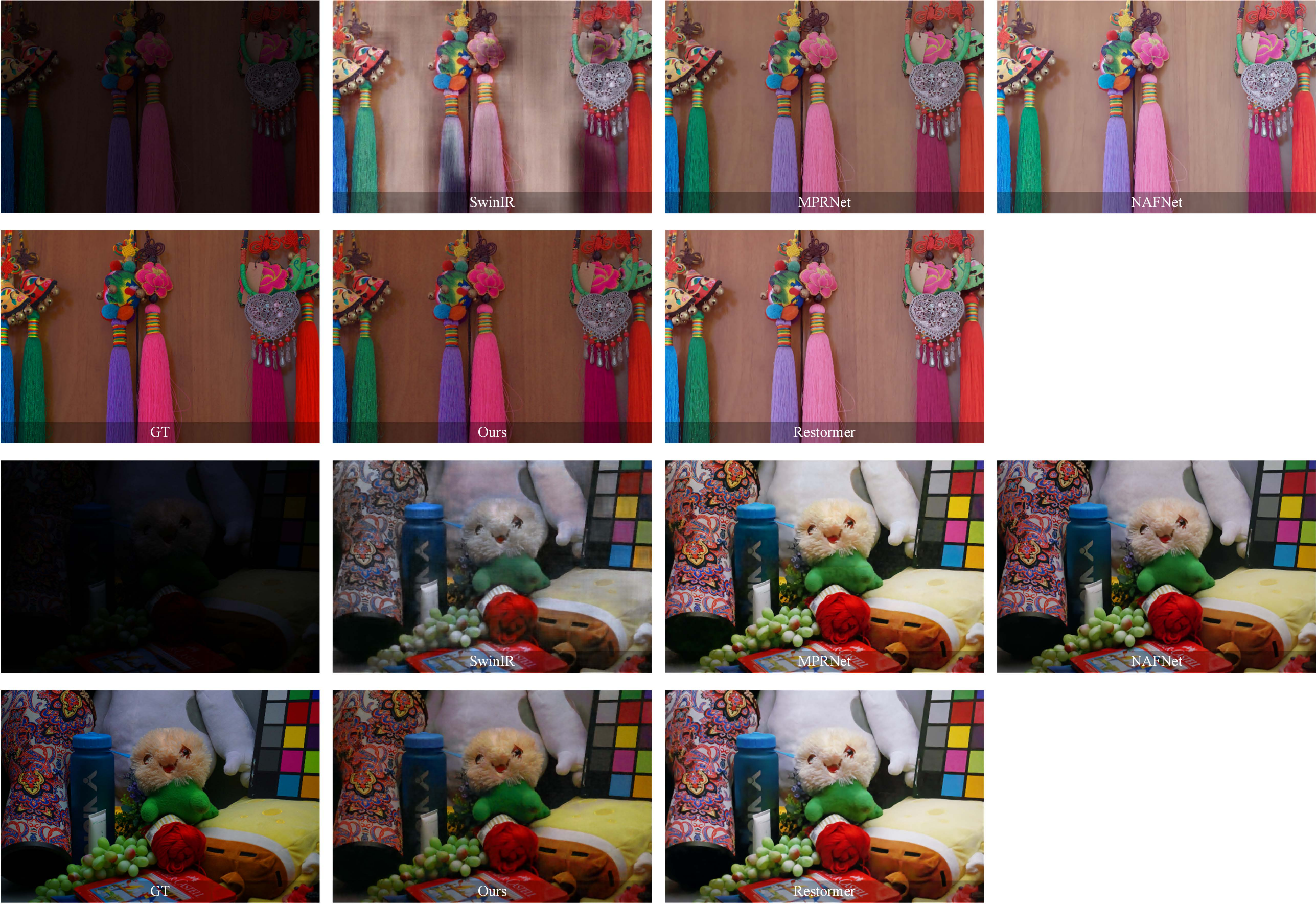}
	\caption{\small \small Visual comparison of results with specific models (lowlight enhancement).}
	\label{fig:suppl_spe_lowlight}
   \vspace{1mm}
\end{figure*}

\begin{figure*}[t]
	\centering
	\setlength{\abovecaptionskip}{0.1cm}
	\setlength{\belowcaptionskip}{-0.3cm}
	\includegraphics[width=\textwidth]{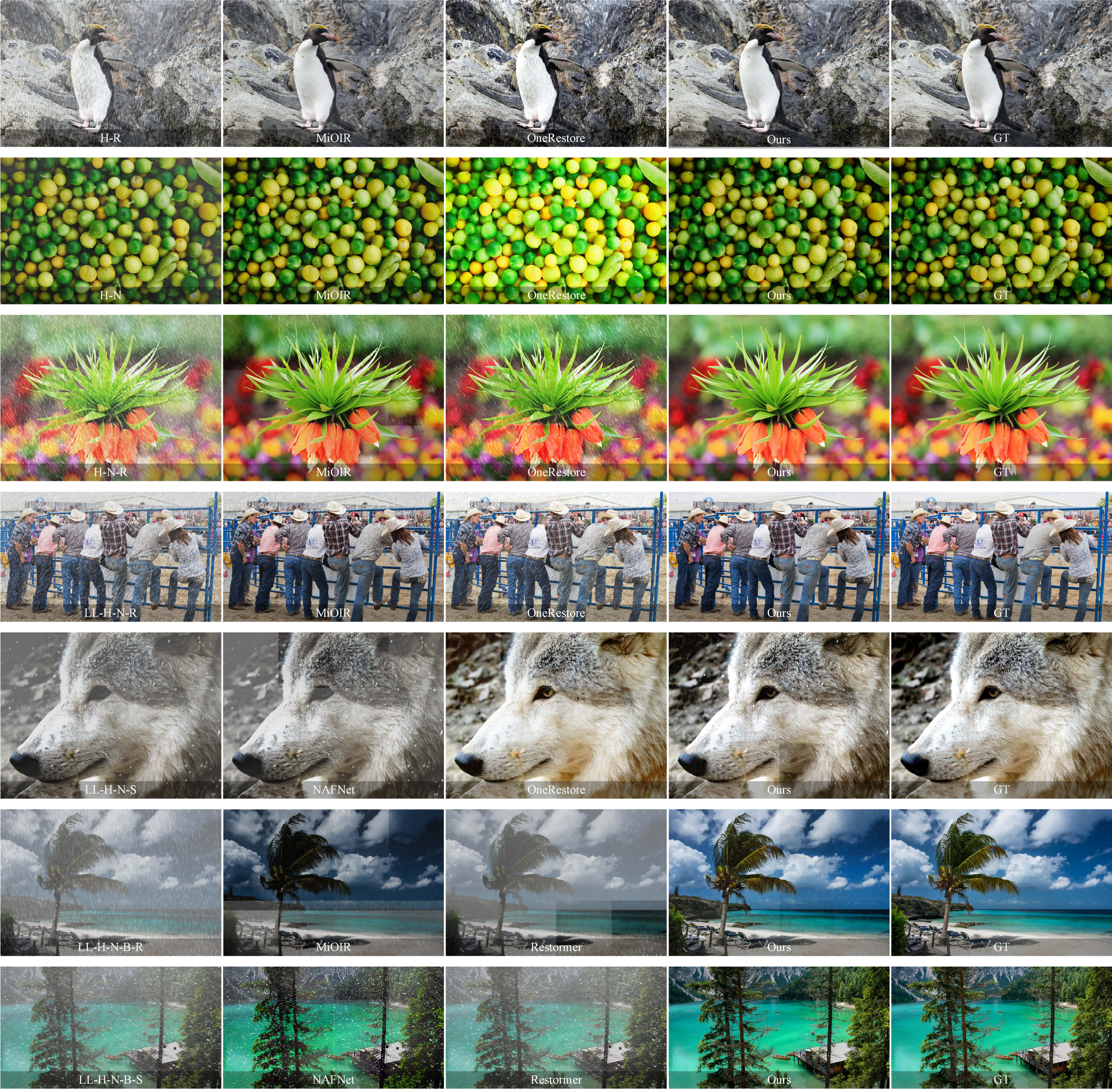}
	\caption{\small \small Visual comparison of results on All-in-One image restoration (mixed-degradation, \textit{in-dist}). Zoom in for more details.}
	\label{fig:suppl_indist}
   \vspace{1mm}
\end{figure*}

\begin{figure*}[t]
	\centering
	\setlength{\abovecaptionskip}{0.1cm}
	\setlength{\belowcaptionskip}{-0.3cm}
	\includegraphics[width=\textwidth]{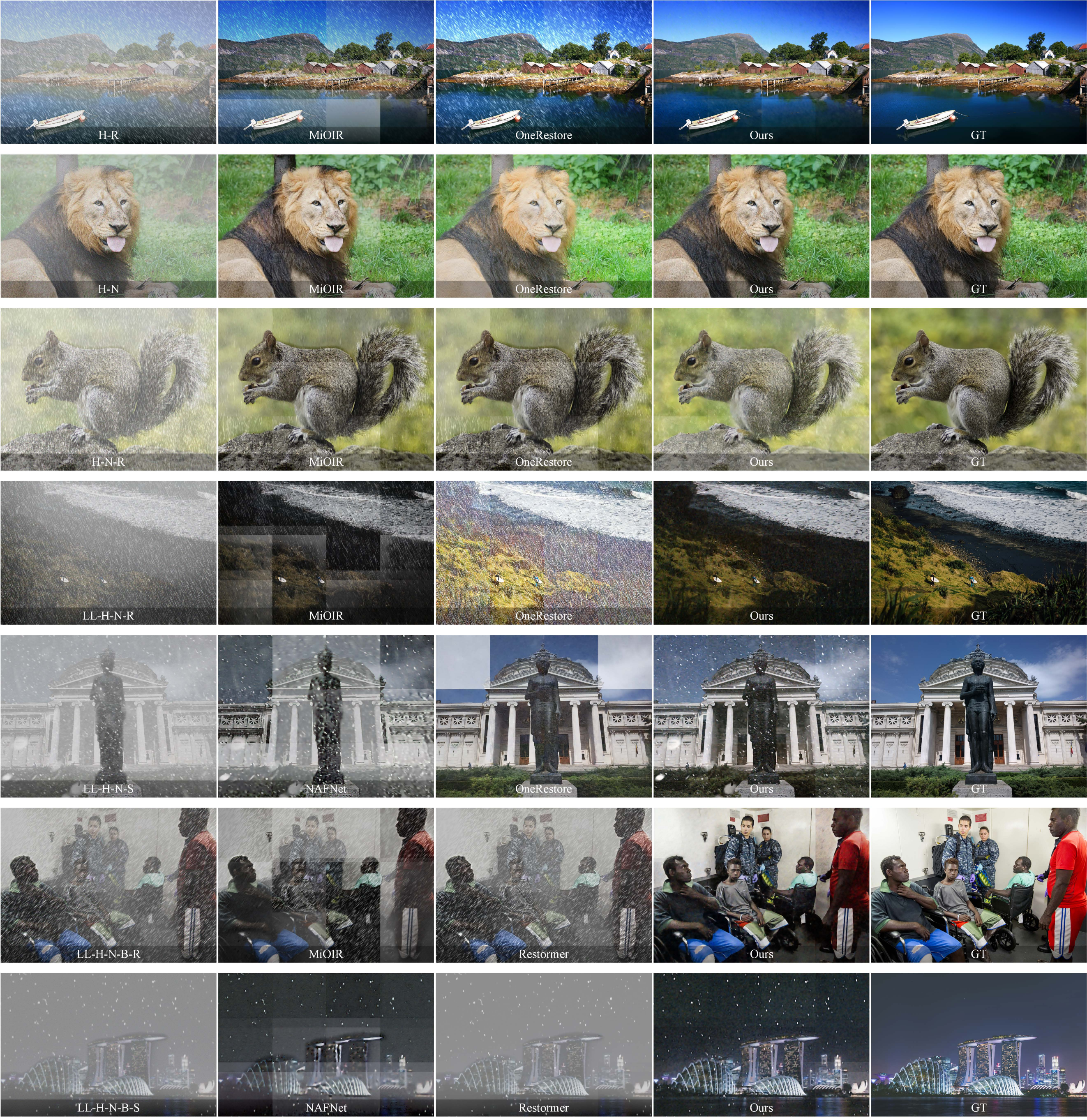}
	\caption{\small \small Visual comparison of results on All-in-One image restoration (mixed-degradation, \textit{out-dist}). Zoom in for more details.}
	\label{fig:suppl_outdist}
   \vspace{1mm}
\end{figure*}

\noindent\textbf{More Visual Comparisons}.
We show more visual comparison results in this section. Compared to task-agnostic methods, all-in-one methods, and specific single-task models, our method can restore more clear results with detailed textures.
The single-degradation results of all-in-one image restoration comparisons can be seen in Fig.~\ref{fig:suppl_aio_derain} to Fig.~\ref{fig:suppl_aio_lowlight}. The comparison results with specific single-task models can be seen in Fig.~\ref{fig:suppl_spe_derain} to Fig.~\ref{fig:suppl_spe_lowlight}.
In the evaluation of mixed degradation, since the resolution of images from DIV2K-valid is too high, limiting inference on the whole image for all methods. We first crop high-resolution images into overlapped $512\times 512$ crops, after inference, we inverse the cropping process and recompose crops back to one whole image, though averaging the overlapped area, the border artifacts cannot be avoided. Comparison result of \textit{in-dist.} and \textit{out-dist.} can be seen in Fig.~\ref{fig:suppl_indist} and Fig.~\ref{fig:suppl_outdist}, respectively.

\clearpage


\clearpage
{
\small

\bibliography{neurips_2025}

\begin{thebibliography}{118}
\providecommand{\natexlab}[1]{#1}
\providecommand{\url}[1]{\texttt{#1}}
\expandafter\ifx\csname urlstyle\endcsname\relax
  \providecommand{\doi}[1]{doi: #1}\else
  \providecommand{\doi}{doi: \begingroup \urlstyle{rm}\Url}\fi

\bibitem[Abdelhamed et~al.(2018)Abdelhamed, Lin, and Brown]{sidd}
Abdelrahman Abdelhamed, Stephen Lin, and Michael~S. Brown.
\newblock A high-quality denoising dataset for smartphone cameras.
\newblock In \emph{CVPR}, 2018.

\bibitem[Agustsson and Timofte(2017)]{div2k}
Eirikur Agustsson and Radu Timofte.
\newblock Ntire 2017 challenge on single image super-resolution: Dataset and study.
\newblock In \emph{CVPRW}, July 2017.

\bibitem[Ai et~al.(2024{\natexlab{a}})Ai, Huang, and He]{lorair}
Yuang Ai, Huaibo Huang, and Ran He.
\newblock Lora-ir: Taming low-rank experts for efficient all-in-one image restoration.
\newblock \emph{arXiv preprint arXiv:2410.15385}, 2024{\natexlab{a}}.

\bibitem[Ai et~al.(2024{\natexlab{b}})Ai, Huang, Zhou, Wang, and He]{MPerceiver}
Yuang Ai, Huaibo Huang, Xiaoqiang Zhou, Jiexiang Wang, and Ran He.
\newblock Multimodal prompt perceiver: Empower adaptiveness, generalizability and fidelity for all-in-one image restoration.
\newblock In \emph{CVPR}, 2024{\natexlab{b}}.

\bibitem[Alpher(2002)]{Alpher02}
FirstName Alpher.
\newblock Frobnication.
\newblock \emph{IEEE TPAMI}, 12\penalty0 (1):\penalty0 234--778, 2002.

\bibitem[Alpher and Fotheringham-Smythe(2003)]{Alpher03}
FirstName Alpher and FirstName Fotheringham-Smythe.
\newblock Frobnication revisited.
\newblock \emph{Journal of Foo}, 13\penalty0 (1):\penalty0 234--778, 2003.

\bibitem[Alpher and Gamow(2005)]{Alpher05}
FirstName Alpher and FirstName Gamow.
\newblock Can a computer frobnicate?
\newblock In \emph{CVPR}, pages 234--778, 2005.

\bibitem[Alpher et~al.(2004)Alpher, Fotheringham-Smythe, and Gamow]{Alpher04}
FirstName Alpher, FirstName Fotheringham-Smythe, and FirstName Gamow.
\newblock Can a machine frobnicate?
\newblock \emph{Journal of Foo}, 14\penalty0 (1):\penalty0 234--778, 2004.

\bibitem[Brian et~al.(2021)Brian, Rami, and Noah]{prompt}
Lester Brian, Al-Rfou Rami, and Constant Noah.
\newblock The power of scale for parameter-efficient prompt tuning.
\newblock In \emph{EMNLP}, November 2021.

\bibitem[Cai et~al.(2023)Cai, Bian, Lin, Wang, Timofte, and Zhang]{retinexformer}
Yuanhao Cai, Hao Bian, Jing Lin, Haoqian Wang, Radu Timofte, and Yulun Zhang.
\newblock Retinexformer: One-stage retinex-based transformer for low-light image enhancement.
\newblock In \emph{ICCV}, 2023.

\bibitem[Cao et~al.(2024{\natexlab{a}})Cao, Cao, Pang, Meng, and Cao]{HAIR}
Jin Cao, Yi~Cao, Li~Pang, Deyu Meng, and Xiaoyong Cao.
\newblock Hair: Hypernetworks-based all-in-one image restoration.
\newblock \emph{arXiv preprint arXiv:2408.08091}, 2024{\natexlab{a}}.

\bibitem[Cao et~al.(2024{\natexlab{b}})Cao, Liu, Zhang, Qiao, and Dong]{grids}
Shuo Cao, Yihao Liu, Wenlong Zhang, Yu~Qiao, and Chao Dong.
\newblock Grids: Grouped multiple-degradation restoration with image degradation similarity.
\newblock In \emph{ECCV}, 2024{\natexlab{b}}.

\bibitem[Chang et~al.(2020)Chang, Lan, Cheng, and Wei]{Chang_2020_CVPR}
Jie Chang, Zhonghao Lan, Changmao Cheng, and Yichen Wei.
\newblock Data uncertainty learning in face recognition.
\newblock In \emph{CVPR}, 2020.

\bibitem[Chen et~al.(2024)Chen, Li, Gu, Ren, Chen, Ye, Pei, Zhou, Song, and Zhu]{RestoreAgent}
Haoyu Chen, Wenbo Li, Jinjin Gu, Jingjing Ren, Sixiang Chen, Tian Ye, Renjing Pei, Kaiwen Zhou, Fenglong Song, and Lei Zhu.
\newblock Restoreagent: Autonomous image restoration agent via multimodal large language models.
\newblock In \emph{NeurIPS}, 2024.

\bibitem[Chen et~al.(2022)Chen, Chu, Zhang, and Sun]{NAFNet}
Liangyu Chen, Xiaojie Chu, Xiangyu Zhang, and Jian Sun.
\newblock Simple baselines for image restoration.
\newblock In \emph{ECCV}, 2022.

\bibitem[Chen et~al.(2023{\natexlab{a}})Chen, Ding, Shen, Zhan, Tomizuka, Learned-Miller, and Gan]{mthl}
Zitian Chen, Mingyu Ding, Yikang Shen, Wei Zhan, Masayoshi Tomizuka, Erik Learned-Miller, and Chuang Gan.
\newblock An efficient general-purpose modular vision model via multi-task heterogeneous training.
\newblock \emph{arXiv preprint arXiv:2306.17165}, 2023{\natexlab{a}}.

\bibitem[Chen et~al.(2023{\natexlab{b}})Chen, Shen, Ding, Chen, Zhao, Learned-Miller, and Gan]{mod_squad}
Zitian Chen, Yikang Shen, Mingyu Ding, Zhenfang Chen, Hengshuang Zhao, Erik Learned-Miller, and Chuang Gan.
\newblock Mod-squad: Designing mixtures of experts as modular multi-task learners.
\newblock \emph{CVPR}, 2023{\natexlab{b}}.

\bibitem[Clark et~al.(2022)Clark, de~las Casas, Guy, Mensch, Paganini, Hoffmann, Damoc, Hechtman, Cai, Borgeaud, van~den Driessche, Rutherford, Hennigan, Johnson, Millican, Cassirer, Jones, Buchatskaya, Budden, Sifre, Osindero, Vinyals, Rae, Elsen, Kavukcuoglu, and Simonyan]{s_base}
Aidan Clark, Diego de~las Casas, Aurelia Guy, Arthur Mensch, Michela Paganini, Jordan Hoffmann, Bogdan Damoc, Blake Hechtman, Trevor Cai, Sebastian Borgeaud, George van~den Driessche, Eliza Rutherford, Tom Hennigan, Matthew Johnson, Katie Millican, Albin Cassirer, Chris Jones, Elena Buchatskaya, David Budden, Laurent Sifre, Simon Osindero, Oriol Vinyals, Jack Rae, Erich Elsen, Koray Kavukcuoglu, and Karen Simonyan.
\newblock Unified scaling laws for routed language models.
\newblock In \emph{NeurIPS}, 2022.

\bibitem[Conde et~al.(2024)Conde, Geigle, and Timofte]{instructir}
Marcos~V. Conde, Gregor Geigle, and Radu Timofte.
\newblock Instructir: High-quality image restoration following human instructions.
\newblock In \emph{ECCV}, 2024.

\bibitem[Dong et~al.(2014)Dong, Loy, He, and Tang]{SRCNN}
Chao Dong, Chen~Change Loy, Kaiming He, and Xiaoou Tang.
\newblock Learning a deep convolutional network for image super-resolution.
\newblock In \emph{ECCV}, 2014.

\bibitem[Fan et~al.(2019)Fan, Chen, Yuan, Hua, Yu, and Chen]{DL}
Qingnan Fan, Dongdong Chen, Lu~Yuan, Gang Hua, Nenghai Yu, and Baoquan Chen.
\newblock A general decoupled learning framework for parameterized image operators.
\newblock \emph{IEEE TPAMI}, 2019.

\bibitem[Fedus et~al.(2022)Fedus, Zoph, and Shazeer]{switch_transformer}
William Fedus, Barret Zoph, and Noam Shazeer.
\newblock Switch transformers: Scaling to trillion parameter models with simple and efficient sparsity.
\newblock \emph{Journal of Machine Learning Research}, 2022.

\bibitem[Fu et~al.(2017)Fu, Huang, Zeng, Huang, Ding, and Paisley]{test2800}
Xueyang Fu, Jiabin Huang, Delu Zeng, Yue Huang, Xinghao Ding, and John Paisley.
\newblock Removing rain from single images via a deep detail network.
\newblock In \emph{CVPR}, 2017.

\bibitem[Guo et~al.(2024)Guo, Gao, Lu, Liu, and He]{OneRestore}
Yu~Guo, Yuan Gao, Yuxu Lu, Ryan~Wen Liu, and Shengfeng He.
\newblock Onerestore: A universal restoration framework for composite degradation.
\newblock In \emph{ECCV}, 2024.

\bibitem[Guo et~al.(2023)Guo, Xiao, Chang, Deng, and Yan]{lhp}
Yun Guo, Xueyao Xiao, Yi~Chang, Shumin Deng, and Luxin Yan.
\newblock From sky to the ground: A large-scale benchmark and simple baseline towards real rain removal.
\newblock In \emph{ICCV}, 2023.

\bibitem[He et~al.(2019)He, Dong, and Qiao]{AdaFM}
Jingwen He, Chao Dong, and Yu~Qiao.
\newblock Modulating image restoration with continual levels via adaptive feature modification layers.
\newblock In \emph{The IEEE Conference on Computer Vision and Pattern Recognition (CVPR)}, June 2019.

\bibitem[He et~al.(2016)He, Zhang, Ren, and Sun]{ResNet}
Kaiming He, Xiangyu Zhang, Shaoqing Ren, and Jian Sun.
\newblock Deep residual learning for image recognition.
\newblock In \emph{CVPR}, 2016.

\bibitem[Houlsby et~al.(2019)Houlsby, Giurgiu, Jastrzebski, Morrone, de~Laroussilhe, Gesmundo, Attariyan, and Gelly]{adapter}
Neil Houlsby, Andrei Giurgiu, Stanislaw Jastrzebski, Bruna Morrone, Quentin de~Laroussilhe, Andrea Gesmundo, Mona Attariyan, and Sylvain Gelly.
\newblock Parameter-efficient transfer learning for nlp.
\newblock In \emph{ICML}, 2019.

\bibitem[Hu et~al.(2022)Hu, Shen, Wallis, Allen-Zhu, Li, Wang, Wang, and Chen]{lora}
Edward~J Hu, Yelong Shen, Phillip Wallis, Zeyuan Allen-Zhu, Yuanzhi Li, Shean Wang, Lu~Wang, and Weizhu Chen.
\newblock Lo{RA}: Low-rank adaptation of large language models.
\newblock In \emph{ICLR}, 2022.

\bibitem[Jacobs et~al.(1991)Jacobs, Jordan, Nowlan, and Hinton]{moe_jacobs}
Robert~A. Jacobs, Michael~I. Jordan, Steven~J. Nowlan, and Geoffrey~E. Hinton.
\newblock Adaptive mixtures of local experts.
\newblock \emph{Neural Computation}, 1991.

\bibitem[JiaKui et~al.(2025)JiaKui, Yao, Lujia, and Yanye]{dcpt}
Hu~JiaKui, Zhengjian Yao, Jin Lujia, and Lu~Yanye.
\newblock Universal image restoration pre-training via degradation classification.
\newblock In \emph{ICLR}, 2025.

\bibitem[Jiang et~al.(2024)Jiang, Zuo, Wu, Jiang, and Liu]{jiang2024survey}
Junjun Jiang, Zengyuan Zuo, Gang Wu, Kui Jiang, and Xianming Liu.
\newblock A survey on all-in-one image restoration: Taxonomy, evaluation and future trends.
\newblock \emph{arXiv preprint arXiv:2410.15067}, 2024.

\bibitem[Jordan and Jacobs(1993)]{h_moe_em}
M.I. Jordan and R.A. Jacobs.
\newblock Hierarchical mixtures of experts and the em algorithm.
\newblock In \emph{Proceedings of 1993 International Conference on Neural Networks (IJCNN-93-Nagoya, Japan)}, 1993.

\bibitem[Kupyn et~al.(2019)Kupyn, Martyniuk, Wu, and Wang]{DeblurGANv2}
Orest Kupyn, Tetiana Martyniuk, Junru Wu, and Zhangyang Wang.
\newblock Deblurgan-v2: Deblurring (orders-of-magnitude) faster and better.
\newblock In \emph{ICCV}, Oct 2019.

\bibitem[LastName(2014{\natexlab{a}})]{Authors14}
FirstName LastName.
\newblock The frobnicatable foo filter, 2014{\natexlab{a}}.
\newblock Face and Gesture submission ID 324. Supplied as supplemental material {\tt fg324.pdf}.

\bibitem[LastName(2014{\natexlab{b}})]{Authors14b}
FirstName LastName.
\newblock Frobnication tutorial, 2014{\natexlab{b}}.
\newblock Supplied as supplemental material {\tt tr.pdf}.

\bibitem[Lepikhin et~al.(2021)Lepikhin, Lee, Xu, Chen, Firat, Huang, Krikun, Shazeer, and Chen]{gshard}
Dmitry Lepikhin, HyoukJoong Lee, Yuanzhong Xu, Dehao Chen, Orhan Firat, Yanping Huang, Maxim Krikun, Noam Shazeer, and Zhifeng Chen.
\newblock {\{}GS{\}}hard: Scaling giant models with conditional computation and automatic sharding.
\newblock In \emph{ICLR}, 2021.

\bibitem[Lewis et~al.(2021)Lewis, Bhosale, Dettmers, Goyal, and Zettlemoyer]{base_layer}
Mike Lewis, Shruti Bhosale, Tim Dettmers, Naman Goyal, and Luke Zettlemoyer.
\newblock Base layers: Simplifying training of large, sparse models.
\newblock In \emph{NeurIPS}, 2021.

\bibitem[Li et~al.(2019)Li, Ren, Fu, Tao, Feng, Zeng, and Wang]{reside}
Boyi Li, Wenqi Ren, Dengpan Fu, Dacheng Tao, Dan Feng, Wenjun Zeng, and Zhangyang Wang.
\newblock Benchmarking single-image dehazing and beyond.
\newblock \emph{IEEE TIP}, 28\penalty0 (1):\penalty0 492--505, 2019.

\bibitem[Li et~al.(2022{\natexlab{a}})Li, Liu, Hu, Wu, Lv, and Peng]{AirNet}
Boyun Li, Xiao Liu, Peng Hu, Zhongqin Wu, Jiancheng Lv, and Xi~Peng.
\newblock {All-In-One Image Restoration for Unknown Corruption}.
\newblock In \emph{CVPR}, 2022{\natexlab{a}}.

\bibitem[Li et~al.(2020)Li, Guo, Ren, Cong, Hou, Kwong, and Tao]{uieb}
Chongyi Li, Chunle Guo, Wenqi Ren, Runmin Cong, Junhui Hou, Sam Kwong, and Dacheng Tao.
\newblock An underwater image enhancement benchmark dataset and beyond.
\newblock \emph{IEEE TIP}, 29:\penalty0 4376--4389, 2020.

\bibitem[Li et~al.(2024{\natexlab{a}})Li, Chen, Dong, Tang, and Pan]{foundir}
Hao Li, Xiang Chen, Jiangxin Dong, Jinhui Tang, and Jinshan Pan.
\newblock Foundir: Unleashing million-scale training data to advance foundation models for image restoration.
\newblock \emph{arXiv preprint arXiv:2412.01427}, 2024{\natexlab{a}}.

\bibitem[Li et~al.(2022{\natexlab{b}})Li, Li, Xiong, and Hoi]{BLIP}
Junnan Li, Dongxu Li, Caiming Xiong, and Steven C.~H. Hoi.
\newblock {BLIP:} bootstrapping language-image pre-training for unified vision-language understanding and generation.
\newblock In \emph{ICML}, 2022{\natexlab{b}}.

\bibitem[Li et~al.(2023{\natexlab{a}})Li, Li, Savarese, and Hoi]{BLIPv2}
Junnan Li, Dongxu Li, Silvio Savarese, and Steven C.~H. Hoi.
\newblock {BLIP-2:} bootstrapping language-image pre-training with frozen image encoders and large language models.
\newblock In \emph{ICML}, 2023{\natexlab{a}}.

\bibitem[Li et~al.(2023{\natexlab{b}})Li, Zhang, Liu, Feng, Wang, Lei, and Zuo]{li2023spatially}
Junyi Li, Zhilu Zhang, Xiaoyu Liu, Chaoyu Feng, Xiaotao Wang, Lei Lei, and Wangmeng Zuo.
\newblock Spatially adaptive self-supervised learning for real-world image denoising.
\newblock In \emph{CVPR}, 2023{\natexlab{b}}.

\bibitem[Li et~al.(2024{\natexlab{b}})Li, Zhang, and Zuo]{li2024tbsn}
Junyi Li, Zhilu Zhang, and Wangmeng Zuo.
\newblock Tbsn: Transformer-based blind-spot network for self-supervised image denoising.
\newblock \emph{arXiv preprint arXiv:2404.07846}, 2024{\natexlab{b}}.

\bibitem[Li et~al.(2022{\natexlab{c}})Li, Huang, Jia, Fan, and Liu]{D2CSR}
Youwei Li, Haibin Huang, Lanpeng Jia, Haoqiang Fan, and Shuaicheng Liu.
\newblock D2c-sr: A divergence to convergence approach for real-world image super-resolution.
\newblock In \emph{ECCV}, 2022{\natexlab{c}}.

\bibitem[Liang et~al.(2022)Liang, Zeng, and Zhang]{DASR}
Jie Liang, Hui Zeng, and Lei Zhang.
\newblock Efficient and degradation-adaptive network for real-world image super-resolution.
\newblock In \emph{ECCV}, 2022.

\bibitem[Liang et~al.(2021)Liang, Cao, Sun, Zhang, Van~Gool, and Timofte]{SwinIR}
Jingyun Liang, Jiezhang Cao, Guolei Sun, Kai Zhang, Luc Van~Gool, and Radu Timofte.
\newblock Swinir: Image restoration using swin transformer.
\newblock In \emph{ICCVW}, 2021.

\bibitem[Lim et~al.(2017)Lim, Son, Kim, Nah, and Lee]{flickr2k}
Bee Lim, Sanghyun Son, Heewon Kim, Seungjun Nah, and Kyoung~Mu Lee.
\newblock Enhanced deep residual networks for single image super-resolution.
\newblock In \emph{CVPRW}, July 2017.

\bibitem[Lin et~al.(2024)Lin, Zhang, Wei, Ren, Jiang, Tian, and Zuo]{lin2024improving}
Jingbo Lin, Zhilu Zhang, Yuxiang Wei, Dongwei Ren, Dongsheng Jiang, Qi~Tian, and Wangmeng Zuo.
\newblock Improving image restoration through removing degradations in textual representations.
\newblock In \emph{CVPR}, 2024.

\bibitem[Liu et~al.(2022{\natexlab{a}})Liu, Xie, Zhang, Yuan, Chen, Zhou, Li, and Tian]{TAPE}
Lin Liu, Lingxi Xie, Xiaopeng Zhang, Shanxin Yuan, Xiangyu Chen, Wengang Zhou, Houqiang Li, and Qi~Tian.
\newblock Tape: Task-agnostic prior embedding for image restoration.
\newblock In \emph{ECCV}, 2022{\natexlab{a}}.

\bibitem[Liu et~al.(2024)Liu, Zhang, Wu, Feng, Wang, Lei, and Zuo]{liu2024learning}
Xiaohui Liu, Zhilu Zhang, Xiaohe Wu, Chaoyu Feng, Xiaotao Wang, Lei Lei, and Wangmeng Zuo.
\newblock Learning real-world image de-weathering with imperfect supervision.
\newblock In \emph{AAAI}, 2024.

\bibitem[Liu et~al.(2022{\natexlab{b}})Liu, Liu, Gu, Zhang, Wu, Qiao, and Dong]{ddr}
Yihao Liu, Anran Liu, Jinjin Gu, Zhipeng Zhang, Wenhao Wu, Yu~Qiao, and Chao Dong.
\newblock Discovering distinctive "semantics" in super-resolution networks.
\newblock \emph{arXiv preprint arXiv:2108.00406}, 2022{\natexlab{b}}.

\bibitem[Liu et~al.(2023)Liu, He, Gu, Kong, Qiao, and Dong]{DegAE}
Yihao Liu, Jingwen He, Jinjin Gu, Xiangtao Kong, Yu~Qiao, and Chao Dong.
\newblock Degae: A new pretraining paradigm for low-level vision.
\newblock In \emph{CVPR}, June 2023.

\bibitem[Liu et~al.(2018{\natexlab{a}})Liu, Jaw, Huang, and Hwang]{DesnowNet}
Yun-Fu Liu, Da-Wei Jaw, Shih-Chia Huang, and Jenq-Neng Hwang.
\newblock Desnownet: Context-aware deep network for snow removal.
\newblock \emph{IEEE TIP}, 27\penalty0 (6):\penalty0 3064--3073, 2018{\natexlab{a}}.

\bibitem[Liu et~al.(2018{\natexlab{b}})Liu, Jaw, Huang, and Hwang]{Snow100K}
Yun-Fu Liu, Da-Wei Jaw, Shih-Chia Huang, and Jenq-Neng Hwang.
\newblock Desnownet: Context-aware deep network for snow removal.
\newblock \emph{IEEE TIP}, 27\penalty0 (6):\penalty0 3064--3073, 2018{\natexlab{b}}.

\bibitem[Liu et~al.(2021)Liu, Lin, Cao, Hu, Wei, Zhang, Lin, and Guo]{liu2021Swin}
Ze~Liu, Yutong Lin, Yue Cao, Han Hu, Yixuan Wei, Zheng Zhang, Stephen Lin, and Baining Guo.
\newblock Swin transformer: Hierarchical vision transformer using shifted windows.
\newblock In \emph{ICCV}, 2021.

\bibitem[Luo et~al.(2024)Luo, Gustafsson, Zhao, Sj{\"o}lund, and Sch{\"o}n]{daclip}
Ziwei Luo, Fredrik~K Gustafsson, Zheng Zhao, Jens Sj{\"o}lund, and Thomas~B Sch{\"o}n.
\newblock Controlling vision-language models for universal image restoration.
\newblock In \emph{ICLR}, 2024.

\bibitem[Ma et~al.(2017)Ma, Duanmu, Wu, Wang, Yong, Li, and Zhang]{wed}
Kede Ma, Zhengfang Duanmu, Qingbo Wu, Zhou Wang, Hongwei Yong, Hongliang Li, and Lei Zhang.
\newblock Waterloo exploration database: New challenges for image quality assessment models.
\newblock \emph{IEEE Transactions on Image Processing}, 2017.

\bibitem[Martin et~al.(2001)Martin, Fowlkes, Tal, and Malik]{bsd68}
David~R. Martin, Charless~C. Fowlkes, Doron Tal, and Jitendra Malik.
\newblock A database of human segmented natural images and its application to evaluating segmentation algorithms and measuring ecological statistics.
\newblock In \emph{ICCV}, 2001.

\bibitem[Mitchell(1997)]{mitchell1997machine}
Tom~M Mitchell.
\newblock \emph{Machine learning}, volume~1.
\newblock 1997.

\bibitem[Mustafa et~al.(2022)Mustafa, Ruiz, Puigcerver, Jenatton, and Houlsby]{multimodal_moe}
Basil Mustafa, Carlos~Riquelme Ruiz, Joan Puigcerver, Rodolphe Jenatton, and Neil Houlsby.
\newblock Multimodal contrastive learning with {LIM}oe: the language-image mixture of experts.
\newblock In \emph{NeurIPS}, 2022.

\bibitem[Nah et~al.(2017{\natexlab{a}})Nah, Kim, and Lee]{GoPro}
Seungjun Nah, Tae~Hyun Kim, and Kyoung~Mu Lee.
\newblock Deep multi-scale convolutional neural network for dynamic scene deblurring.
\newblock In \emph{CVPR}, July 2017{\natexlab{a}}.

\bibitem[Nah et~al.(2017{\natexlab{b}})Nah, Kim, and Lee]{urban100}
Seungjun Nah, Tae~Hyun Kim, and Kyoung~Mu Lee.
\newblock Deep multi-scale convolutional neural network for dynamic scene deblurring.
\newblock In \emph{CVPR}, July 2017{\natexlab{b}}.

\bibitem[Park et~al.(2023)Park, Lee, and Chun]{ADMS}
Dongwon Park, Byung~Hyun Lee, and Se~Young Chun.
\newblock All-in-one image restoration for unknown degradations using adaptive discriminative filters for specific degradations.
\newblock In \emph{CVPR}, 2023.

\bibitem[Potlapalli et~al.(2023)Potlapalli, Zamir, Khan, and Khan]{PromptIR}
Vaishnav Potlapalli, Syed~Waqas Zamir, Salman Khan, and Fahad Khan.
\newblock Promptir: Prompting for all-in-one image restoration.
\newblock In \emph{NeurIPS}, 2023.

\bibitem[Puigcerver et~al.(2024)Puigcerver, Ruiz, Mustafa, and Houlsby]{soft_moe}
Joan Puigcerver, Carlos~Riquelme Ruiz, Basil Mustafa, and Neil Houlsby.
\newblock From sparse to soft mixtures of experts.
\newblock In \emph{ICLR}, 2024.

\bibitem[Qian et~al.(2018)Qian, Tan, Yang, Su, and Liu]{raindrop}
Rui Qian, Robby~T. Tan, Wenhan Yang, Jiajun Su, and Jiaying Liu.
\newblock Attentive generative adversarial network for raindrop removal from a single image.
\newblock In \emph{CVPR}, June 2018.

\bibitem[Radford et~al.(2021)Radford, Kim, Hallacy, Ramesh, Goh, Agarwal, Sastry, Askell, Mishkin, Clark, Krueger, and Sutskever]{CLIP}
Alec Radford, Jong~Wook Kim, Chris Hallacy, Aditya Ramesh, Gabriel Goh, Sandhini Agarwal, Girish Sastry, Amanda Askell, Pamela Mishkin, Jack Clark, Gretchen Krueger, and Ilya Sutskever.
\newblock Learning transferable visual models from natural language supervision.
\newblock In \emph{ICML}, 2021.

\bibitem[Rajbhandari et~al.(2022)Rajbhandari, Li, Yao, Zhang, Aminabadi, Awan, Rasley, and He]{deepspeed_moe}
Samyam Rajbhandari, Conglong Li, Zhewei Yao, Minjia Zhang, {Reza Yazdani} Aminabadi, {Ammar Ahmad} Awan, Jeff Rasley, and Yuxiong He.
\newblock Deepspeed-moe: Advancing mixture-of-experts inference and training to power next-generation ai scale.
\newblock In \emph{ICML}, 2022.

\bibitem[Ren et~al.(2019)Ren, Zuo, Hu, Zhu, and Meng]{PReNet}
Dongwei Ren, Wangmeng Zuo, Qinghua Hu, Pengfei Zhu, and Deyu Meng.
\newblock Progressive image deraining networks: {A} better and simpler baseline.
\newblock In \emph{CVPR}, 2019.

\bibitem[Rim et~al.(2020)Rim, Lee, Won, and Cho]{realblur}
Jaesung Rim, Haeyun Lee, Jucheol Won, and Sunghyun Cho.
\newblock Real-world blur dataset for learning and benchmarking deblurring algorithms.
\newblock In \emph{ECCV}, 2020.

\bibitem[Ruiz et~al.(2021)Ruiz, Puigcerver, Mustafa, Neumann, Jenatton, Pinto, Keysers, and Houlsby]{v_moe}
Carlos~Riquelme Ruiz, Joan Puigcerver, Basil Mustafa, Maxim Neumann, Rodolphe Jenatton, Andr{\'e}~Susano Pinto, Daniel Keysers, and Neil Houlsby.
\newblock Scaling vision with sparse mixture of experts.
\newblock In \emph{Advances in Neural Information Processing Systems}, 2021.

\bibitem[Shazeer et~al.(2017)Shazeer, Mirhoseini, Maziarz, Davis, Le, Hinton, and Dean]{sparse_gated_moe}
Noam Shazeer, *Azalia Mirhoseini, *Krzysztof Maziarz, Andy Davis, Quoc Le, Geoffrey Hinton, and Jeff Dean.
\newblock Outrageously large neural networks: The sparsely-gated mixture-of-experts layer.
\newblock In \emph{ICLR}, 2017.

\bibitem[Sheikh et~al.()Sheikh, Z.Wang, Cormack, and Bovik]{live1}
H.R. Sheikh, Z.Wang, L.~Cormack, and A.C. Bovik.
\newblock Live image quality assessment database release 2.
\newblock \url{http://live.ece.utexas.edu/research/quality}.

\bibitem[Shen et~al.(2023)Shen, Yao, Li, Darrell, Keutzer, and He]{scaling_vlm}
Sheng Shen, Zhewei Yao, Chunyuan Li, Trevor Darrell, Kurt Keutzer, and Yuxiong He.
\newblock Scaling vision-language models with sparse mixture of experts.
\newblock In \emph{EMNLP}, 2023.

\bibitem[Simonyan and Zisserman(2015)]{vgg}
Karen Simonyan and Andrew Zisserman.
\newblock Very deep convolutional networks for large-scale image recognition.
\newblock In \emph{ICLR}, 2015.

\bibitem[Song et~al.(2023)Song, He, Qian, and Du]{Dehazeformer}
Yuda Song, Zhuqing He, Hui Qian, and Xin Du.
\newblock Vision transformers for single image dehazing.
\newblock \emph{IEEE TIP}, 32:\penalty0 1927--1941, 2023.

\bibitem[Sun et~al.(2018)Sun, Yu, and Wang]{tip2018}
Yujing Sun, Yizhou Yu, and Wenping Wang.
\newblock Moiré photo restoration using multiresolution convolutional neural networks.
\newblock \emph{IEEE TIP}, 27\penalty0 (8):\penalty0 4160--4172, 2018.

\bibitem[Tian et~al.(2024)Tian, Han, Chen, Xi, Zhang, Hu, Xu, and Wang]{InstructIPT}
Yuchuan Tian, Jianhong Han, Hanting Chen, Yuanyuan Xi, Guoyang Zhang, Jie Hu, Chao Xu, and Yunhe Wang.
\newblock Instruct-ipt: All-in-one image processing transformer via weight modulation.
\newblock \emph{arXiv preprint arXiv:2407.00676}, 2024.

\bibitem[Valanarasu et~al.(2022)Valanarasu, Yasarla, and Patel]{Transweather}
Jeya Maria~Jose Valanarasu, Rajeev Yasarla, and Vishal~M. Patel.
\newblock Transweather: Transformer-based restoration of images degraded by adverse weather conditions.
\newblock In \emph{CVPR}, 2022.

\bibitem[Vaswani et~al.(2017)Vaswani, Shazeer, Parmar, Uszkoreit, Jones, Gomez, Kaiser, and Polosukhin]{Transformer}
Ashish Vaswani, Noam Shazeer, Niki Parmar, Jakob Uszkoreit, Llion Jones, Aidan~N Gomez, \L~ukasz Kaiser, and Illia Polosukhin.
\newblock Attention is all you need.
\newblock In \emph{NeurIPS}, 2017.

\bibitem[Wang et~al.(2018)Wang, Cheng, Liu, and Liu]{ams_softmax}
Feng Wang, Jian Cheng, Weiyang Liu, and Haijun Liu.
\newblock Additive margin softmax for face verification.
\newblock \emph{IEEE Signal Processing Letters}, 2018.

\bibitem[Wang et~al.(2021)Wang, Xie, Dong, and Shan]{wang2021realesrgan}
Xintao Wang, Liangbin Xie, Chao Dong, and Ying Shan.
\newblock Real-esrgan: Training real-world blind super-resolution with pure synthetic data.
\newblock In \emph{ICCVW}, 2021.

\bibitem[Wei et~al.(2018)Wei, Wang, Yang, and Liu]{LOL}
Chen Wei, Wenjing Wang, Wenhan Yang, and Jiaying Liu.
\newblock Deep retinex decomposition for low-light enhancement.
\newblock In \emph{BMVC}, 2018.

\bibitem[Wenlong et~al.(2023)Wenlong, Xiaohui, SHI, Xiangyu, Yu, Xiaoyun, Xiao-Ming, and Chao]{tgsr}
Zhang Wenlong, Li~Xiaohui, Guangyuan SHI, Chen Xiangyu, Qiao Yu, Zhang Xiaoyun, Wu~Xiao-Ming, and Dong Chao.
\newblock Real-world image super-resolution as multi-task learning.
\newblock In \emph{NeurIPS}, 2023.

\bibitem[Wu et~al.(2025)Wu, Jiang, Wang, Jiang, and Liu]{tur}
Gang Wu, Junjun Jiang, Yijun Wang, Kui Jiang, and Xianming Liu.
\newblock Debiased all-in-one image restoration with task uncertainty regularization.
\newblock In \emph{AAAI}, 2025.

\bibitem[Wu et~al.(2022)Wu, Liu, Chen, Chen, Dai, and Yuan]{residue_moe}
Lemeng Wu, Mengchen Liu, Yinpeng Chen, Dongdong Chen, Xiyang Dai, and Lu~Yuan.
\newblock Residual mixture of experts.
\newblock \emph{arXiv preprint arXiv:2204.09636}, 2022.

\bibitem[Wu et~al.(2024)Wu, Huang, and Wei]{moe_lora}
Xun Wu, Shaohan Huang, and Furu Wei.
\newblock Mixture of lo{RA} experts.
\newblock In \emph{ICLR}, 2024.

\bibitem[Xiangtao et~al.(2024)Xiangtao, Chao, and Lei]{mioir}
Kong Xiangtao, Dong Chao, and Zhang Lei.
\newblock Towards effective multiple-in-one image restoration: A sequential and prompt learning strategy.
\newblock \emph{arXiv preprint arXiv:2401.03379}, 2024.

\bibitem[Xie et~al.(2021)Xie, Wang, Dong, Qi, and Shan]{FAIG}
Liangbin Xie, Xintao Wang, Chao Dong, Zhongang Qi, and Ying Shan.
\newblock Finding discriminative filters for specific degradations in blind super-resolution.
\newblock In \emph{NeurIPS}, 2021.

\bibitem[Xu et~al.(2022)Xu, Wang, Fu, , and Jia]{lolv2}
Xiaogang Xu, Ruixing Wang, Chi-Wing Fu, , and Jiaya Jia.
\newblock Snr-aware low-light image enhancement.
\newblock In \emph{CVPR}, 2022.

\bibitem[Xu et~al.(2024)Xu, Gao, Zhong, Chao, and Ji]{U_WADN}
Yimin Xu, Nanxi Gao, Yunshan Zhong, Fei Chao, and Rongrong Ji.
\newblock Unified-width adaptive dynamic network for all-in-one image restoration.
\newblock \emph{arXiv preprint arXiv:2405.15475}, 2024.

\bibitem[Yang et~al.(2023)Yang, Chen, Tan, Liu, Chu, Bao, Yuan, Hua, and Yu]{HQ_50K}
Qinhong Yang, Dongdong Chen, Zhentao Tan, Qiankun Liu, Qi~Chu, Jianmin Bao, Lu~Yuan, Gang Hua, and Nenghai Yu.
\newblock Hq-50k: A large-scale, high-quality dataset for image restoration.
\newblock \emph{arXiv preprint arXiv:2306.05390}, 2023.

\bibitem[Yang et~al.(2017)Yang, Tan, Feng, Liu, Guo, and Yan]{Rain200H}
Wenhan Yang, Robby~T. Tan, Jiashi Feng, Jiaying Liu, Zongming Guo, and Shuicheng Yan.
\newblock Deep joint rain detection and removal from a single image.
\newblock In \emph{CVPR}, 2017.

\bibitem[Zamfir et~al.(2024)Zamfir, Wu, Mehta, Paudel, Zhang, and Timofte]{DaAIR}
Eduard Zamfir, Zongwei Wu, Nancy Mehta, Danda~Dani Paudel, Yulun Zhang, and Radu Timofte.
\newblock Efficient degradation-aware any image restoration.
\newblock \emph{arXiv preprint arXiv:2405.15475}, 2024.

\bibitem[Zamfir et~al.(2025)Zamfir, Wu, Mehta, Tan, Paudel, Zhang, and Timofte]{moce}
Eduard Zamfir, Zongwei Wu, Nancy Mehta, Yuedong Tan, Danda~Pani Paudel, Yulun Zhang, and Radu Timofte.
\newblock Complexity experts are task-discriminative learners for any image restoration.
\newblock In \emph{CVPR}, 2025.

\bibitem[Zamir et~al.(2021)Zamir, Arora, Khan, Hayat, Khan, Yang, and Shao]{MPRNet}
Syed~Waqas Zamir, Aditya Arora, Salman Khan, Munawar Hayat, Fahad~Shahbaz Khan, Ming-Hsuan Yang, and Ling Shao.
\newblock Multi-stage progressive image restoration.
\newblock In \emph{CVPR}, 2021.

\bibitem[Zamir et~al.(2022)Zamir, Arora, Khan, Hayat, Khan, and Yang]{Restormer}
Syed~Waqas Zamir, Aditya Arora, Salman Khan, Munawar Hayat, Fahad~Shahbaz Khan, and Ming-Hsuan Yang.
\newblock Restormer: Efficient transformer for high-resolution image restoration.
\newblock In \emph{CVPR}, 2022.

\bibitem[Zhang and Patel(2018)]{test1200}
He~Zhang and Vishal~M. Patel.
\newblock Density-aware single image de-raining using a multi-stream dense network.
\newblock In \emph{CVPR}, 2018.

\bibitem[Zhang et~al.(2023{\natexlab{a}})Zhang, Huang, Yao, Yang, Yu, Zhou, and Zhao]{IDR}
Jinghao Zhang, Jie Huang, Mingde Yao, Zizheng Yang, Hu~Yu, Man Zhou, and Feng Zhao.
\newblock Ingredient-oriented multi-degradation learning for image restoration.
\newblock In \emph{CVPR}, 2023{\natexlab{a}}.

\bibitem[Zhang et~al.(2017)Zhang, Zuo, Chen, Meng, and Zhang]{DnCNN}
Kai Zhang, Wangmeng Zuo, Yunjin Chen, Deyu Meng, and Lei Zhang.
\newblock Beyond a gaussian denoiser: Residual learning of deep {CNN} for image denoising.
\newblock \emph{IEEE TIP}, 26\penalty0 (7):\penalty0 3142--3155, 2017.

\bibitem[Zhang et~al.(2018)Zhang, Zuo, and Zhang]{SRMD}
Kai Zhang, Wangmeng Zuo, and Lei Zhang.
\newblock Learning a single convolutional super-resolution network for multiple degradations.
\newblock In \emph{CVPR}, 2018.

\bibitem[Zhang et~al.(2021{\natexlab{a}})Zhang, Liang, Van~Gool, and Timofte]{BSRGAN}
Kai Zhang, Jingyun Liang, Luc Van~Gool, and Radu Timofte.
\newblock Designing a practical degradation model for deep blind image super-resolution.
\newblock In \emph{ICCV}, 2021{\natexlab{a}}.

\bibitem[Zhang et~al.(2011)Zhang, Wu, Buades, and Li]{mcmaster}
Lei Zhang, Xiaolin Wu, Antoni Buades, and Xin Li.
\newblock Color demosaicking by local directional interpolation and nonlocal adaptive thresholding.
\newblock \emph{J. Electronic Imaging}, 20\penalty0 (2):\penalty0 023016, 2011.

\bibitem[Zhang et~al.(2023{\natexlab{b}})Zhang, Gu, Chen, Dong, Zhang, and Yang]{crafting}
Ruofan Zhang, Jinjin Gu, Haoyu Chen, Chao Dong, Yulun Zhang, and Wenming Yang.
\newblock Crafting training degradation distribution for the accuracy-generalization trade-off in real-world super-resolution.
\newblock In \emph{ICML}, 2023{\natexlab{b}}.

\bibitem[Zhang et~al.(2022{\natexlab{a}})Zhang, Shi, Liu, Dong, and Wu]{Zhang_2022_CVPR}
Wenlong Zhang, Guangyuan Shi, Yihao Liu, Chao Dong, and Xiao-Ming Wu.
\newblock A closer look at blind super-resolution: Degradation models, baselines, and performance upper bounds.
\newblock In \emph{CVPRW}, June 2022{\natexlab{a}}.

\bibitem[Zhang et~al.(2021{\natexlab{b}})Zhang, Dong, Pan, Zhu, Tai, Wang, Li, Huang, and Wang]{revide}
Xinyi Zhang, Hang Dong, Jinshan Pan, Chao Zhu, Ying Tai, Chengjie Wang, Jilin Li, Feiyue Huang, and Fei Wang.
\newblock Learning to restore hazy video: A new real-world dataset and a new method.
\newblock In \emph{CVPR}, pages 9239--9248, 2021{\natexlab{b}}.

\bibitem[Zhang et~al.(2023{\natexlab{c}})Zhang, Wei, Jiang, Zhang, Zuo, and Tian]{zhang2023controlvideo}
Yabo Zhang, Yuxiang Wei, Dongsheng Jiang, Xiaopeng Zhang, Wangmeng Zuo, and Qi~Tian.
\newblock Controlvideo: Training-free controllable text-to-video generation.
\newblock \emph{arXiv preprint arXiv:2305.13077}, 2023{\natexlab{c}}.

\bibitem[Zhang et~al.(2021{\natexlab{c}})Zhang, Wang, Liu, Wang, Zhang, and Zuo]{zhang2021learning}
Zhilu Zhang, Haolin Wang, Ming Liu, Ruohao Wang, Jiawei Zhang, and Wangmeng Zuo.
\newblock Learning raw-to-srgb mappings with inaccurately aligned supervision.
\newblock In \emph{ICCV}, 2021{\natexlab{c}}.

\bibitem[Zhang et~al.(2022{\natexlab{b}})Zhang, Xu, Liu, Yan, and Zuo]{zhang2022self}
Zhilu Zhang, RongJian Xu, Ming Liu, Zifei Yan, and Wangmeng Zuo.
\newblock Self-supervised image restoration with blurry and noisy pairs.
\newblock In \emph{NeurIPS}, 2022{\natexlab{b}}.

\bibitem[Zhang et~al.(2024)Zhang, Zhang, Wu, Yan, and Zuo]{zhang2024bracketing}
Zhilu Zhang, Shuohao Zhang, Renlong Wu, Zifei Yan, and Wangmeng Zuo.
\newblock Bracketing is all you need: Unifying image restoration and enhancement tasks with multi-exposure images.
\newblock \emph{arXiv preprint arXiv:2401.00766}, 2024.

\bibitem[Zhong et~al.(2024)Zhong, Tang, He, Fang, and Yuan]{convolution_meets_lora}
Zihan Zhong, Zhiqiang Tang, Tong He, Haoyang Fang, and Chun Yuan.
\newblock Convolution meets lo{RA}: Parameter efficient finetuning for segment anything model.
\newblock In \emph{ICLR}, 2024.

\bibitem[Zhou et~al.(2021)Zhou, Ren, Emerton, Lim, and Large]{toled}
Yuqian Zhou, David Ren, Neil Emerton, Sehoon Lim, and Timothy Large.
\newblock Image restoration for under-display camera.
\newblock In \emph{CVPR}, 2021.

\bibitem[Zhu et~al.(2022)Zhu, Zhu, Wang, Wang, Li, Wang, and Dai]{uniperceivermoe}
Jinguo Zhu, Xizhou Zhu, Wenhai Wang, Xiaohua Wang, Hongsheng Li, Xiaogang Wang, and Jifeng Dai.
\newblock Uni-perceiver-moe: Learning sparse generalist models with conditional moes.
\newblock In \emph{NeurIPS}, 2022.

\bibitem[Zhu et~al.(2023)Zhu, Wang, Fu, Yang, Guo, Dai, Qiao, and Hu]{WGWSNet}
Yurui Zhu, Tianyu Wang, Xueyang Fu, Xuanyu Yang, Xin Guo, Jifeng Dai, Yu~Qiao, and Xiaowei Hu.
\newblock Learning weather-general and weather-specific features for image restoration under multiple ad verse weather conditions.
\newblock In \emph{CVPR}, 2023.

\bibitem[Zoph et~al.(2022)Zoph, Bello, Kumar, Du, Huang, Dean, Shazeer, and Fedus]{st_moe}
Barret Zoph, Irwan Bello, Sameer Kumar, Nan Du, Yanping Huang, Jeff Dean, Noam Shazeer, and William Fedus.
\newblock St-moe: Design stable and transferable sparse expert models.
\newblock \emph{arXiv preprint arXiv:2202.08906}, 2022.

\end{thebibliography}
\bibliographystyle{plainnat}

}

\end{document}